\documentclass{article}

\usepackage[square,numbers]{natbib}
\usepackage[nonatbib, final]{neurips_2022}
\usepackage[utf8]{inputenc} 
\usepackage[T1]{fontenc}    
\usepackage{hyperref}       
\usepackage{url}            
\usepackage{booktabs}       
\usepackage{amsfonts}       
\usepackage{nicefrac}       
\usepackage{microtype}      
\usepackage{graphicx}
\usepackage{url}            
\usepackage{booktabs}       
\usepackage{amsfonts}       
\usepackage{nicefrac}       
\usepackage{rotating}
\usepackage[table,xcdraw]{xcolor}
\usepackage{multirow}
\usepackage{xspace}
\usepackage{amsmath}
\usepackage{mathtools}
\usepackage{listings}
\usepackage{amsfonts}
\usepackage{amssymb,algorithm}
\usepackage{algpseudocode}
\usepackage{textcomp}
\usepackage{enumitem}
\usepackage{cleveref}
\usepackage{subcaption}
\usepackage{float}
\usepackage{wrapfig}
\usepackage{enumitem}
\usepackage{diagbox}
\usepackage{listings}
\usepackage[textsize=tiny]{todonotes}

\newcommand{\mdoll}{\raisebox{-2pt}{\includegraphics[height=12pt]{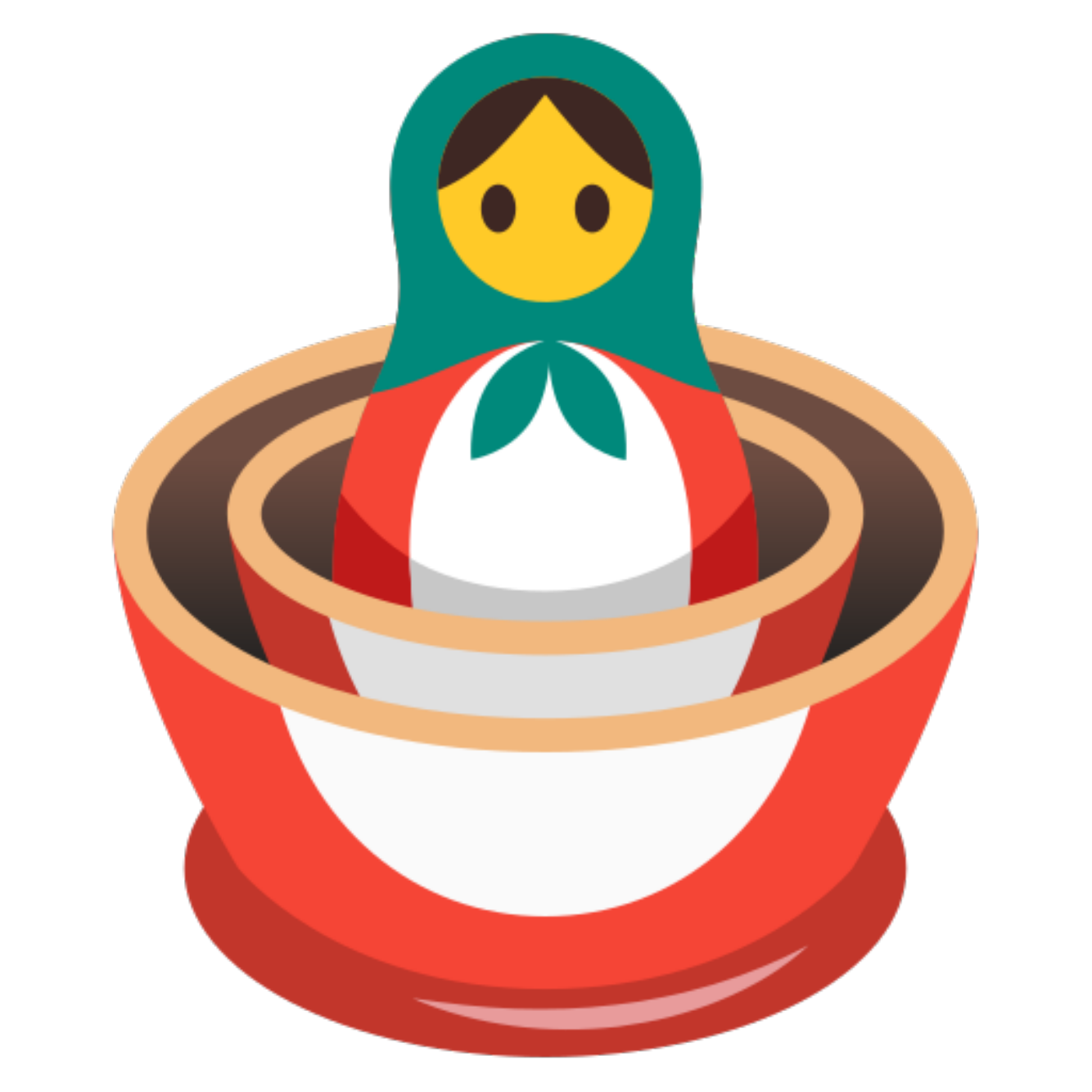}}}

\crefformat{section}{\S#2#1#3}
\crefformat{subsection}{\S#2#1#3}

\newcommand{\floor}[1]{\left\lfloor#1\right\rfloor}

\newcommand{\Mcal}{\mathcal{M}}

\newcommand{\Xcal}{\mathcal{X}}

\hypersetup{
    colorlinks=true,
    linkcolor=blue,
    filecolor=magenta,      
    urlcolor=cyan,
}
\newcommand{\nrl}{\ensuremath{{\rm MRL}}\xspace}
\newcommand{\mrl}{\ensuremath{{\rm MRL}}\xspace}
\newcommand{\mrle}{\ensuremath{{\rm MRL\text{--}E}}\xspace}
\newcommand{\alg}{\ensuremath{{\rm Matryoshka~Representation~Learning}}\xspace}
\newcommand{\MRL}{\alg}
\newcommand{\mr}{\ensuremath{{\rm Matryoshka~Representation}}\xspace}
\newcommand{\mrs}{\ensuremath{{\rm Matryoshka~Representations}}\xspace}
\newcommand{\ma}{\ensuremath{{\rm Matryoshka}}\xspace}

\newcommand{\FF}{FF}
\newcommand{\MH}{\mrl}
\newcommand{\SH}{\mrle}
\newcommand{\INVTwo}{ImageNetV2\xspace}
\newcommand{\Dims}{Rep. Size\xspace}
\newcommand{\InIk}{ImageNet-1K\xspace}
\newcommand{\InIVk}{ImageNet-4K\xspace}
\newcommand{\retdim}{$D_s$}
\newcommand{\rerankdim}{$D_r$}

\begin{document}
\title{Matryoshka Representation Learning}
\author{
Aditya Kusupati\thanks{Equal contribution -- AK led the project with extensive support from GB and AR for experimentation.}~~$^{\dagger\diamond}$, Gantavya Bhatt$^{*\dagger}$, Aniket Rege$^{*\dagger}$,\\
\textbf{Matthew Wallingford$^\dagger$, Aditya Sinha$^\diamond$, Vivek Ramanujan$^\dagger$, William Howard-Snyder$^\dagger$,}\\
\textbf{Kaifeng Chen$^\diamond$, Sham Kakade$^\ddagger$, Prateek Jain$^\diamond$ and Ali Farhadi$^\dagger$}\\
$^\dagger$University of Washington, 
$^\diamond$Google Research, $^\ddagger$Harvard University\\
\texttt{\{kusupati,ali\}@cs.washington.edu}, \texttt{prajain@google.com}\vspace{-4mm}
}

\maketitle
\begin{abstract}
Learned representations are a central component in modern ML systems, serving a multitude of downstream tasks. When training such representations, it is often the case that computational and statistical constraints for each downstream task are unknown. In this context, rigid fixed-capacity representations can be either over or under-accommodating to the task at hand. This leads us to ask: \emph{can we design a flexible representation that can adapt to multiple downstream tasks with varying computational resources?} Our main contribution is \mdoll~\alg~(\mrl) which encodes information at different granularities and allows a single embedding to adapt to the computational constraints of downstream tasks. \mrl~minimally modifies existing representation learning pipelines and imposes no additional cost during inference and deployment. \mrl learns coarse-to-fine representations that are at least as accurate and rich as independently trained low-dimensional representations. The flexibility within the learned \mrs~offer: (a) up to $\mathbf{14}\times$ smaller embedding size for ImageNet-1K classification at the same level of accuracy; (b) up to \textbf{$\mathbf{14}\times$} real-world speed-ups for large-scale retrieval on ImageNet-1K and 4K; and (c) up to $\mathbf{2}$\% accuracy improvements for long-tail few-shot classification, all while being as robust as the original representations. Finally, we show that \mrl~extends seamlessly to web-scale datasets (ImageNet, JFT) across various modalities -- vision (ViT, ResNet), vision + language (ALIGN) and language (BERT). \mrl code and pretrained models are open-sourced at \url{https://github.com/RAIVNLab/MRL}.
\vspace{-2mm}
\end{abstract}

\section{Introduction}
\label{sec:intro}

Learned representations~\citep{lecun2015deep} are fundamental building blocks of real-world ML systems~\citep{NayakUnderstanding,Waldburger2019Search}. Trained once and frozen, $d$-dimensional representations encode rich information and can be used to perform multiple downstream tasks~\citep{bengio2012deep}. The deployment of deep representations has two steps: (1)~an expensive yet constant-cost forward pass to compute the representation~\citep{he2016deep} and (2) utilization of the representation for downstream applications~\citep{VertexAIMatchingEngine,varma2019extreme}. Compute costs for the latter part of the pipeline scale with the embedding dimensionality as well as the data size ($N$) and label space ($L$). At web-scale~\citep{dean2009challenges,sun2017revisiting} this utilization cost overshadows the feature computation cost. The rigidity in these representations forces the use of high-dimensional embedding vectors across multiple tasks despite the varying resource and accuracy constraints that require flexibility. 

Human perception of the natural world has a naturally coarse-to-fine granularity~\citep{harris2000coarse,hegde2008time}. However, perhaps due to the inductive bias of gradient-based training~\citep{soudry2018implicit}, deep learning models tend to diffuse ``information'' across the entire representation vector. The desired elasticity is usually enabled in the existing flat and fixed representations either through training multiple low-dimensional models~\citep{he2016deep}, jointly optimizing sub-networks of varying capacity~\citep{cai2019once,yu2018slimmable} or post-hoc compression~\citep{hotelling1933analysis,linde1980algorithm}. Each of these techniques struggle to meet the requirements for adaptive large-scale deployment either due to training/maintenance overhead, numerous expensive forward passes through all of the data, storage and memory cost for multiple copies of encoded data, expensive on-the-fly feature selection or a significant drop in accuracy. By encoding coarse-to-fine-grained representations, which are as accurate as the independently trained counterparts, we learn with minimal overhead a representation that can be deployed \emph{adaptively} at no additional cost during inference.

\begin{wrapfigure}{r}{0.6\columnwidth}\vspace{-4mm}
 \centering
 \vspace{-2pt}
 \includegraphics[width=0.57\columnwidth,]{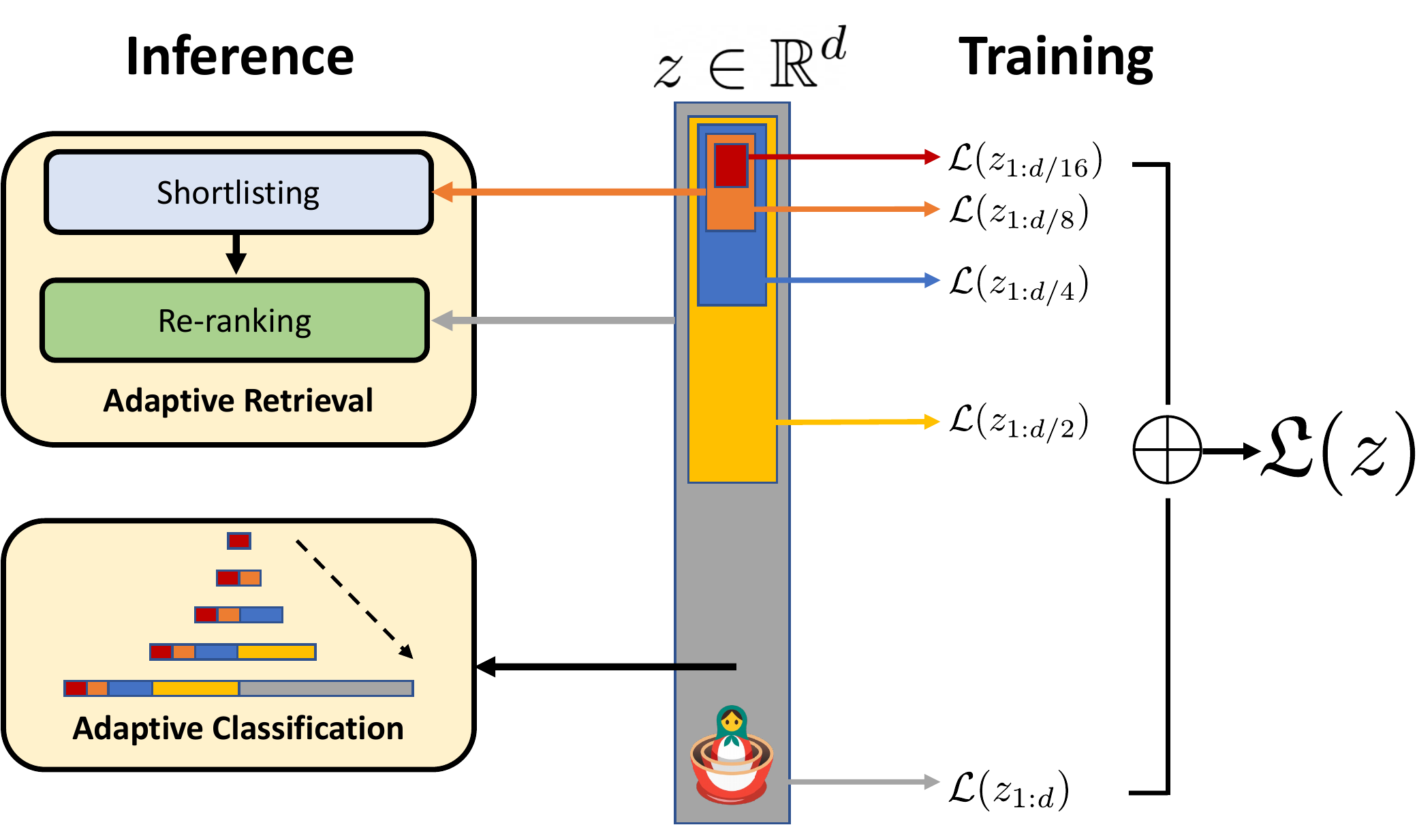}
 \vspace{-2pt}
 \caption{\small \mdoll~\alg is adaptable to any representation learning setup and begets a \mr $z$ by optimizing the original loss $\mathcal{L}(.)$ at $O(\log(d))$ chosen representation sizes. \mr can be utilized effectively for adaptive deployment across environments and downstream tasks.}
\label{fig:teaser}	
\vspace{-6pt}
\end{wrapfigure}

We introduce \mdoll~\alg~(\mrl) to induce flexibility in the learned representation. \mrl learns representations of varying capacities within the same high-dimensional vector through explicit optimization of $O(\log(d))$ lower-dimensional vectors in a nested fashion, hence the name \ma. \mrl can be adapted to any existing representation pipeline and is easily extended to many standard tasks in computer vision and natural language processing. Figure~\ref{fig:teaser} illustrates the core idea of \alg~(\mrl) and the adaptive deployment settings of the learned \mrs.

\setlength{\textfloatsep}{5pt}
The first $m$-dimensions, $m\in[d]$, of the \mr is an information-rich low-dimensional vector, at no additional training cost, that is as accurate as an independently trained $m$-dimensional representation. The information within the \mr increases with the dimensionality creating a coarse-to-fine grained representation, all without significant training or additional deployment overhead. \mrl equips the representation vector with the desired flexibility and multifidelity that can ensure a near-optimal accuracy-vs-compute trade-off. With these advantages, \mrl enables adaptive deployment based on accuracy and compute constraints.

The \mrs improve efficiency for large-scale classification and retrieval without any significant loss of accuracy. While there are potentially several applications of coarse-to-fine \mrs, in this work we focus on two key building blocks of real-world ML systems: large-scale classification and retrieval. For classification, we use adaptive cascades with the variable-size representations from a model trained with \mrl, significantly reducing the average dimension of embeddings needed to achieve a particular accuracy. For example, on ImageNet-1K, \mrl + adaptive classification results in up to a $14\times$ smaller representation size at the same accuracy as baselines (Section~\ref{sec:adaptive_classification_main}). Similarly, we use \mrl in an adaptive retrieval system. Given a query, we shortlist retrieval candidates using the first few dimensions of the query embedding, and then successively use more dimensions to re-rank the retrieved set. A simple implementation of this approach leads to  $128\times$ theoretical (in terms of FLOPS) and $14\times$  wall-clock time speedups compared to a single-shot retrieval system that uses a standard embedding vector; note that \mrl's retrieval accuracy is comparable to that of single-shot retrieval (Section~\ref{sec:adaptive_retrieval_main}). Finally, as \mrl explicitly learns coarse-to-fine representation vectors, intuitively it should share more semantic information among its various dimensions (Figure~\ref{fig:int-acc}). This is reflected in up to $2\%$  accuracy gains in  long-tail continual learning settings while being as robust as the original embeddings. Furthermore, due to its coarse-to-fine grained nature, \mrl can also be used as method to analyze hardness of classification among instances and information bottlenecks.

\textbf{We make the following key contributions:}
\begin{enumerate}[leftmargin=*]\vspace{-4mm}
    \itemsep 0pt
    \topsep 0pt
    \parskip 2pt
    \item We introduce \mdoll~\alg~(\mrl) to obtain flexible representations (\mrs) for adaptive deployment (Section \ref{sec:method}).
    \item Up to $14\times$ faster yet accurate large-scale classification and retrieval using \mrl (Section \ref{sec:apps}).
    \item Seamless adaptation of \mrl across modalities (vision - ResNet \& ViT, vision + language - ALIGN, language - BERT) and to web-scale data (ImageNet-1K/4K, JFT-300M and ALIGN data).
    \item Further analysis of \mrl's representations in the context of other downstream tasks (Section \ref{sec:analysis}).
\end{enumerate}

\section{Related Work}
\label{sec:rw}

\paragraph{Representation Learning.} Large-scale datasets like ImageNet~\citep{deng2009imagenet,russakovsky2015imagenet} and JFT~\citep{sun2017revisiting} enabled the learning of general purpose representations for computer vision~\citep{bengio2012deep,yosinski2014transferable}. These representations are typically learned through supervised and un/self-supervised learning paradigms. Supervised pretraining~\citep{he2016deep,krizhevsky2012imagenet,simonyan2014very} casts representation learning as a multi-class/label classification problem, while un/self-supervised learning learns representation via proxy tasks like instance classification~\citep{wu2018unsupervised} and reconstruction~\citep{he2021masked,masci2011stacked}. Recent advances~\citep{chen2020simple,he2020momentum} in contrastive learning~\citep{gutmann2010noise} enabled learning from web-scale data~\citep{divvala2014learning} that powers large-capacity cross-modal models~\citep{desai2021virtex,jia2021scaling,radford2021learning,zellers2022merlot}. Similarly, natural language applications are built~\citep{howard2018universal} on large language models~\citep{brown2020language} that are pretrained~\citep{peters-etal-2018-deep,ruder2019transfer} in a un/self-supervised fashion with masked language modelling~\citep{devlin2018bert} or autoregressive training~\citep{radford2018improving}.

\mdoll~\alg~(\mrl) is complementary to all these setups and can be adapted with minimal overhead (Section~\ref{sec:method}). \mrl equips representations with multifidelity at no additional cost which enables adaptive deployment based on the data and task (Section~\ref{sec:apps}). 

\paragraph{Efficient Classification and Retrieval.} Efficiency in classification and retrieval during inference can be studied with respect to the high yet constant deep featurization costs or the search cost which scales with the size of the label space and data. Efficient neural networks address the first issue through a variety of algorithms~\citep{gholami2021survey,kusupati2020soft} and design choices~\citep{howard2017mobilenets,kusupati2018fastgrnn,tan2019efficientnet}. However, with a strong featurizer, most of the issues with scale are due to the linear dependence on number of labels ($L$), size of the data ($N$) and representation size ($d$), stressing RAM, disk and processor all at the same time. 

The sub-linear complexity dependence on number of labels has been well studied in context of compute~\citep{bengio2010label,jain2019slice,prabhu2020extreme} and memory~\citep{dietterich1994solving} using Approximate Nearest Neighbor Search (ANNS)~\citep{malkov2018efficient} or leveraging the underlying hierarchy~\citep{deng2011hierarchical,kusupati2021llc}. In case of the representation size, often dimensionality reduction~\citep{salakhutdinov2007learning,van2009dimensionality}, hashing techniques~\citep{datar2004locality,kulis2009fast,salakhutdinov2009semantic} and feature selection~\citep{mitra2002unsupervised} help in alleviating selective aspects of the $O(d)$ scaling at a cost of significant drops in accuracy. Lastly, most real-world search systems~\citep{chang2021extreme,dean2009challenges} are often powered by large-scale embedding based retrieval~\citep{chang2020pre,NayakUnderstanding} that scales in cost with the ever increasing web-data. While categorization~\citep{varma2019extreme,yu2022pecos} clusters similar things together, it is imperative to be equipped with retrieval capabilities that can bring forward every instance~\citep{brin1998anatomy}. Approximate Nearest Neighbor Search (ANNS)~\citep{indyk1998approximate} makes it feasible with efficient indexing~\citep{datar2004locality} and traversal~\citep{bentley1990k,beygelzimer2006cover} to present the users with the most similar documents/images from the database for a requested query. Widely adopted HNSW~\citep{malkov2018efficient} ($O(d\log(N))$) is as accurate as exact retrieval ($O(dN)$) at the cost of a graph-based index overhead for RAM and disk~\citep{jayaram2019diskann}.

\mrl tackles the linear dependence on embedding size, $d$, by learning multifidelity \mrs. Lower-dimensional \mrs are as accurate as independently trained counterparts without the multiple expensive forward passes. \mrs provide an \textit{intermediate abstraction} between high-dimensional vectors and their efficient ANNS indices through the adaptive embeddings nested within the original representation vector (Section~\ref{sec:apps}). All other aforementioned efficiency techniques are complementary and can be readily applied to the learned \mrs obtained from \mrl.

Several works in efficient neural network literature~\citep{cai2019once,wallingford2022task,yu2018slimmable} aim at packing neural networks of varying capacity within the same larger network. However, the weights for each progressively smaller network can be different and often require distinct forward passes to isolate the final representations. This is detrimental for adaptive inference due to the need for re-encoding the entire retrieval database with expensive sub-net forward passes of varying capacities. Several works~\citep{engelsma2022hers,gong2019intrinsic,nanda2023diffused, li2018measuring} investigate the notions of intrinsic dimensionality and redundancy of representations and objective spaces pointing to minimum description length~\citep{rissanen1978modeling}. Finally, ordered representations proposed by~\citet{rippel2014learning} use nested dropout in the context of autoencoders to learn nested representations. \mrl differentiates itself in formulation by optimizing only for $O(\log(d))$ nesting dimensions instead of $O(d)$. Despite this, \mrl diffuses information to intermediate dimensions interpolating between the optimized \mr sizes accurately (Figure~\ref{fig:int-acc}); making web-scale feasible.
\section{\mdoll~\alg}
\label{sec:method}

For $d\in\mathbb{N}$, consider a set $\mathcal{M}\subset [d]$ of representation sizes. For a datapoint $x$ in the input domain $\Xcal$, our goal is to learn a $d$-dimensional representation vector $z \in \mathbb{R}^d$. For every $m\in\mathcal{M}$, \alg~(\mrl) enables each of the first $m$ dimensions of the embedding vector, $z_{1:m}\in\mathbb{R}^m$ to be independently capable of being a transferable and general purpose representation of the datapoint $x$. We
obtain $z$ using a deep neural network $F(\, \cdot\,  ;\theta_F)\colon \mathcal{X} \rightarrow \mathbb{R}^d$ parameterized by learnable weights $\theta_F$, i.e., $z \coloneqq F(x; \theta_F)$. The multi-granularity is captured through the set of the chosen dimensions $\mathcal{M}$, that contains less than $\log(d)$ elements, i.e., $\lvert \Mcal\rvert \leq \floor{\log(d)}$. The usual set $\mathcal{M}$ consists of consistent halving until the representation size hits a low information bottleneck. We discuss the design choices in Section~\ref{sec:apps} for each of the representation learning settings.  

For the ease of exposition, we present the formulation for fully supervised representation learning via multi-class classification. \alg modifies the typical setting to become a multi-scale representation learning problem on the same task. For example, we train ResNet50~\citep{he2016deep} on ImageNet-1K~\citep{russakovsky2015imagenet} which embeds a $224 \times 224$ pixel image into a $d=2048$ representation vector and then passed through a linear classifier to make a prediction, $\hat{y}$ among the $L=1000$ labels. For \mrl, we choose $\mathcal{M} = \{8,16,\ldots,1024,2048\}$ as the nesting dimensions.

Suppose we are given a labelled dataset $\mathcal{D}=\{(x_1, y_1), \ldots, (x_N,y_N)\}$ where $x_i\in \mathcal{X}$ is an input point and $y_i \in [L]$ is the label of $x_i$ for all $i\in[N]$. \mrl optimizes the multi-class classification loss for each of the nested dimension $m\in \mathcal{M}$ using standard empirical risk minimization using a separate linear classifier, parameterized by $\mathbf{W}^{(m)} \in \mathbb{R}^{L\times m}$. All the losses are aggregated after scaling with their relative importance $\left(c_m \geq 0\right)_{m\in\mathcal{M}}$ respectively. That is, we solve 
\begin{equation}
    \label{eq:phase1}
    \min_{\left\lbrace\mathbf{W}^{(m)}\right\rbrace_{m\in\mathcal{M}},\ \theta_F}
    \frac{1}{N}\sum_{i\in [N]} \sum_{m\in \mathcal{M}} c_m\cdot{\cal L}\left(\mathbf{W}^{(m)} \cdot F(x_i; \theta_F)_{1:m}\ ;\ y_i\right) \ ,
\end{equation}
where ${\cal L}\colon \mathbb{R}^L\times [L] \to \mathbb{R}_+$ is the multi-class softmax cross-entropy loss function. This is a standard optimization problem that can be solved using sub-gradient descent methods. We set all the importance scales, $c_m=1$ for all $m\in\mathcal{M}$; see Section~\ref{sec:analysis} for ablations. Lastly, despite only optimizing for $O(\log(d))$ nested dimensions, \mrl results in accurate representations, that interpolate, for dimensions that fall between the chosen granularity of the representations (Section~\ref{sec:classification}).

We call this formulation as \alg~(\mrl). A natural way to make this efficient is through weight-tying across all the linear classifiers, i.e., by defining $\mathbf{W}^{(m)} = \mathbf{W}_{1:m}$ for a set of common weights $\mathbf{W}\in\mathbb{R}^{L\times d}$. This would reduce the memory cost due to the linear classifiers by almost half, which would be crucial in cases of extremely large output spaces~\citep{varma2019extreme,yu2022pecos}. This variant is called \textit{Efficient} \alg~(\SH). Refer to  Alg~\ref{code:NCE-Loss} and Alg~\ref{code:MRL} in Appendix~\ref{sec:code} for the building blocks of \alg~(\mrl). 

\paragraph{Adaptation to Learning Frameworks.} \mrl can be adapted seamlessly to most representation learning frameworks at web-scale with minimal modifications (Section~\ref{sec:rep_learning}). For example, \mrl's adaptation to masked language modelling reduces to \SH~due to the weight-tying between the input embedding matrix and the linear classifier. For contrastive learning, both in context of vision \& vision + language, \mrl is applied to both the embeddings that are being contrasted with each other. The presence of normalization on the representation needs to be handled independently for each of the nesting dimension for best results (see Appendix~\ref{sec:appendix-mrl_model_training} for more details).

\section{Applications}
\label{sec:apps}

In this section, we discuss \alg (\mrl) for a diverse set of applications along with an extensive evaluation of the learned multifidelity representations. Further, we showcase the downstream applications of the learned \mrs for flexible large-scale deployment through (a) Adaptive Classification (AC) and (b) Adaptive Retrieval (AR).

\begin{figure}[t!]
\centering
  \hspace{-3mm}
\begin{minipage}{.48\columnwidth}
  \centering
        \includegraphics[width=\columnwidth]{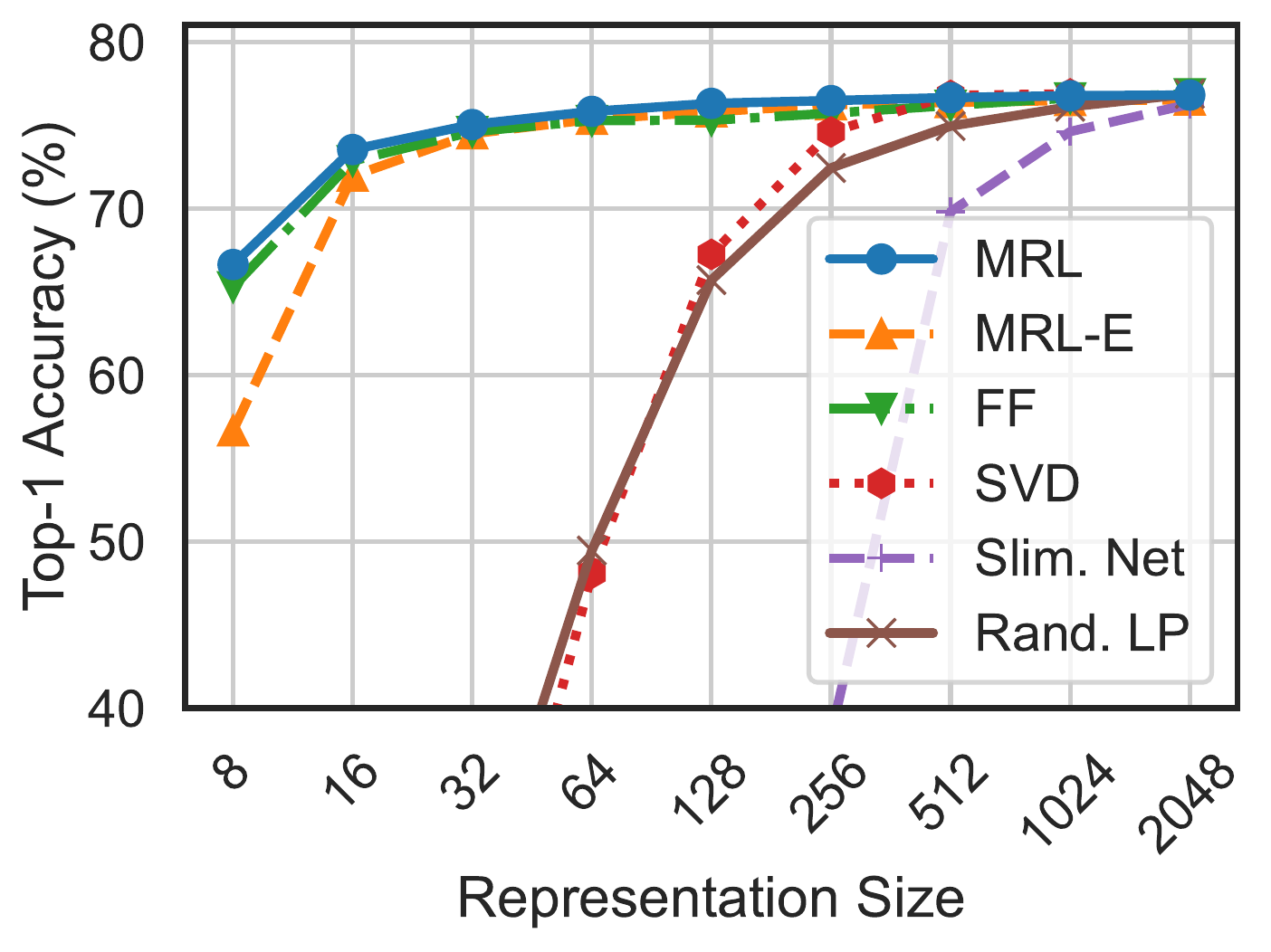}
        \caption{\InIk~linear classification accuracy of ResNet50 models. \mrl is as accurate as the independently trained FF models for every representation size.}
  \label{fig:r50-acc}
\end{minipage}%
\hspace{3mm}
\begin{minipage}{.48\columnwidth}
    \centering
  \includegraphics[width=\columnwidth]{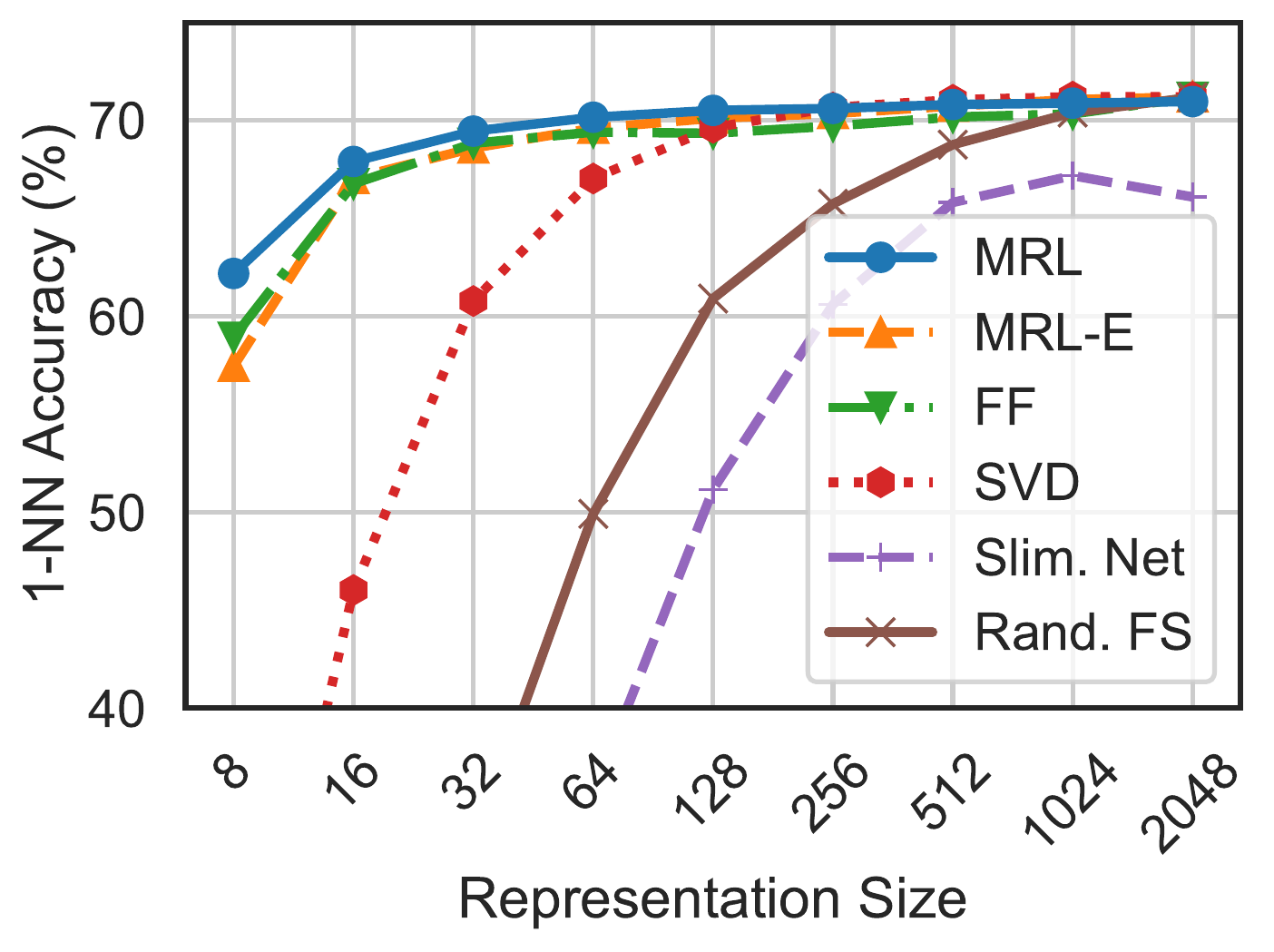}
    \caption{\InIk~1-NN accuracy of ResNet50 models measuring the representation quality for downstream task. \mrl outperforms all the baselines across all representation sizes.}
  \label{fig:r50-knn-acc}
\end{minipage}
\end{figure}
\subsection{Representation Learning} 
\label{sec:rep_learning}
We adapt \alg~(\mrl) to various representation learning setups (a) Supervised learning for vision: ResNet50~\citep{he2016deep} on ImageNet-1K~\citep{russakovsky2015imagenet} and ViT-B/16~\citep{dosovitskiy2020image} on JFT-300M~\citep{sun2017revisiting}, (b) Contrastive learning for vision + language: ALIGN model with ViT-B/16 vision encoder and BERT language encoder on ALIGN data~\citep{jia2021scaling} and (c) Masked language modelling: BERT~\citep{devlin2018bert} on English Wikipedia and BooksCorpus~\citep{zhu2015aligning}. Please refer to Appendices~\ref{sec:datasets} and~\ref{sec:appendix-mrl_model_training}  for details regarding the model architectures, datasets and training specifics.

We do not search for best hyper-parameters for all \mrl experiments but use the same hyper-parameters as the independently trained baselines. ResNet50 outputs a $2048$-dimensional representation while ViT-B/16 and BERT-Base output $768$-dimensional embeddings for each data point. We use $\mathcal{M} = \{8,16,32,64,128,256,512,1024,2048\}$ and $\mathcal{M} = \{12,24,48,96,192,384,768\}$ as the explicitly optimized nested dimensions respectively. Lastly, we extensively compare the \mrl and \SH~models to independently trained low-dimensional (fixed feature) representations (FF), dimensionality reduction (SVD), sub-net method (slimmable networks~\citep{yu2018slimmable}) and randomly selected features of the highest capacity FF model.

In section~\ref{sec:classification}, we evaluate the quality and capacity of the learned representations through linear classification/probe (LP) and 1-nearest neighbour (1-NN) accuracy. Experiments show that \mrl models remove the dependence on $|\mathcal{M}|$ resource-intensive independently trained models for the coarse-to-fine representations while being as accurate. Lastly, we show that despite optimizing only for $|\mathcal{M}|$ dimensions, \mrl models diffuse the information, in an interpolative fashion, across all the $d$ dimensions providing the finest granularity required for adaptive deployment.

\subsection{Classification}
\label{sec:classification}

Figure~\ref{fig:r50-acc} compares the linear classification accuracy of ResNet50 models trained and evaluated on ImageNet-1K. ResNet50--\mrl model is at least as accurate as each FF model at every representation size in $\mathcal{M}$ while \SH~is within $1\%$ starting from $16$-dim. Similarly, Figure~\ref{fig:r50-knn-acc} showcases the comparison of learned representation quality through 1-NN accuracy on ImageNet-1K (trainset with 1.3M samples as the database and validation set with 50K samples as the queries). \mrs are up to $2\%$ more accurate than their fixed-feature counterparts for the lower-dimensions while being as accurate elsewhere. 1-NN accuracy is an excellent proxy, at no additional training cost, to gauge the utility of learned representations in the downstream tasks. 

\begin{figure}[t!]
\centering
\begin{minipage}{.46\columnwidth}
  \centering
  \hspace{-5mm}
        \includegraphics[height=2in]{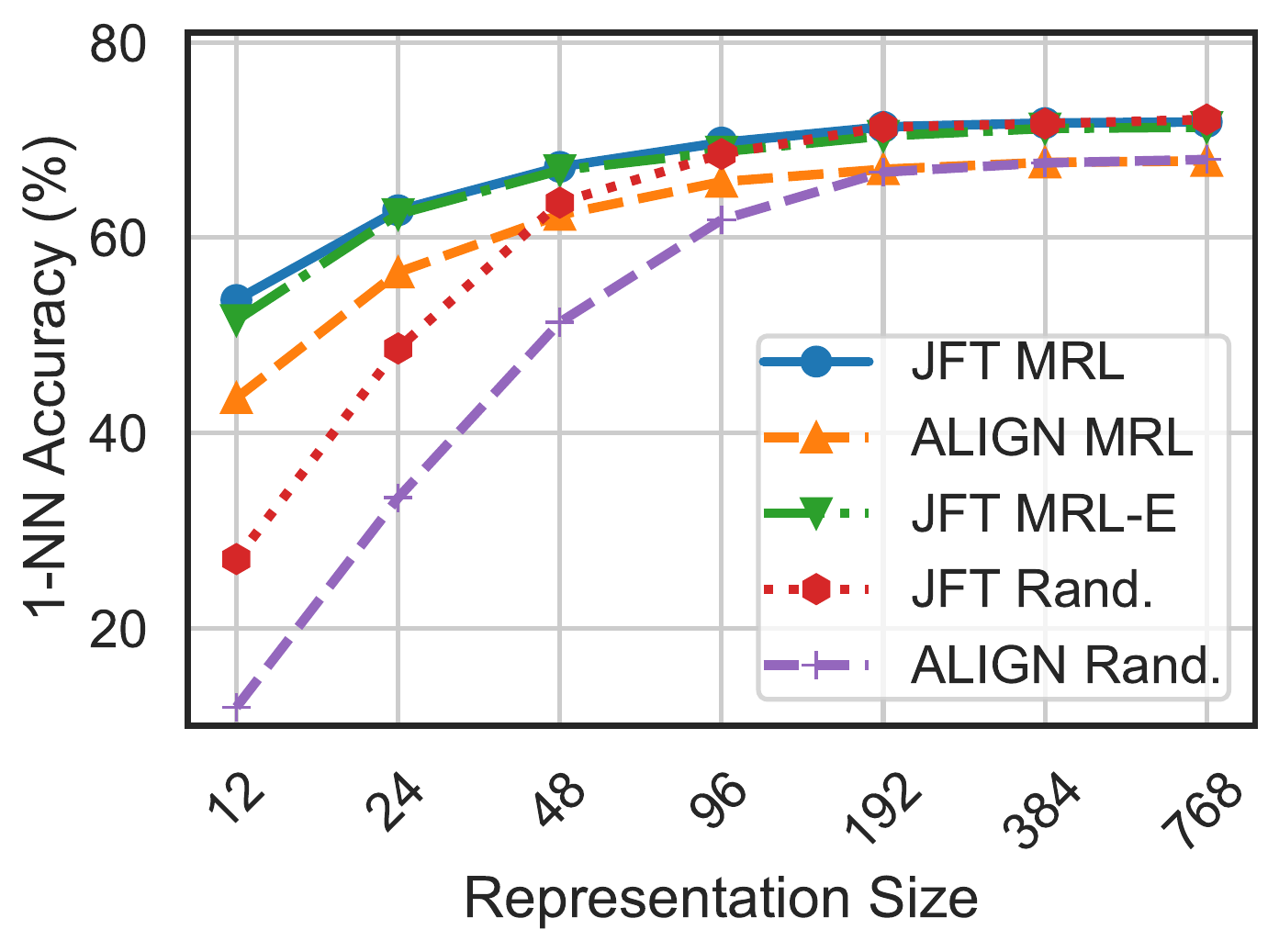}
        \caption{ImageNet-1K 1-NN accuracy for ViT-B/16 models trained on JFT-300M \& as part of ALIGN. \mrl scales seamlessly to web-scale with minimal training overhead.}
  \label{fig:ViT-acc}
\end{minipage}%
\hspace{5mm}
\begin{minipage}{.48\columnwidth}
  \centering
        \includegraphics[height=2in]{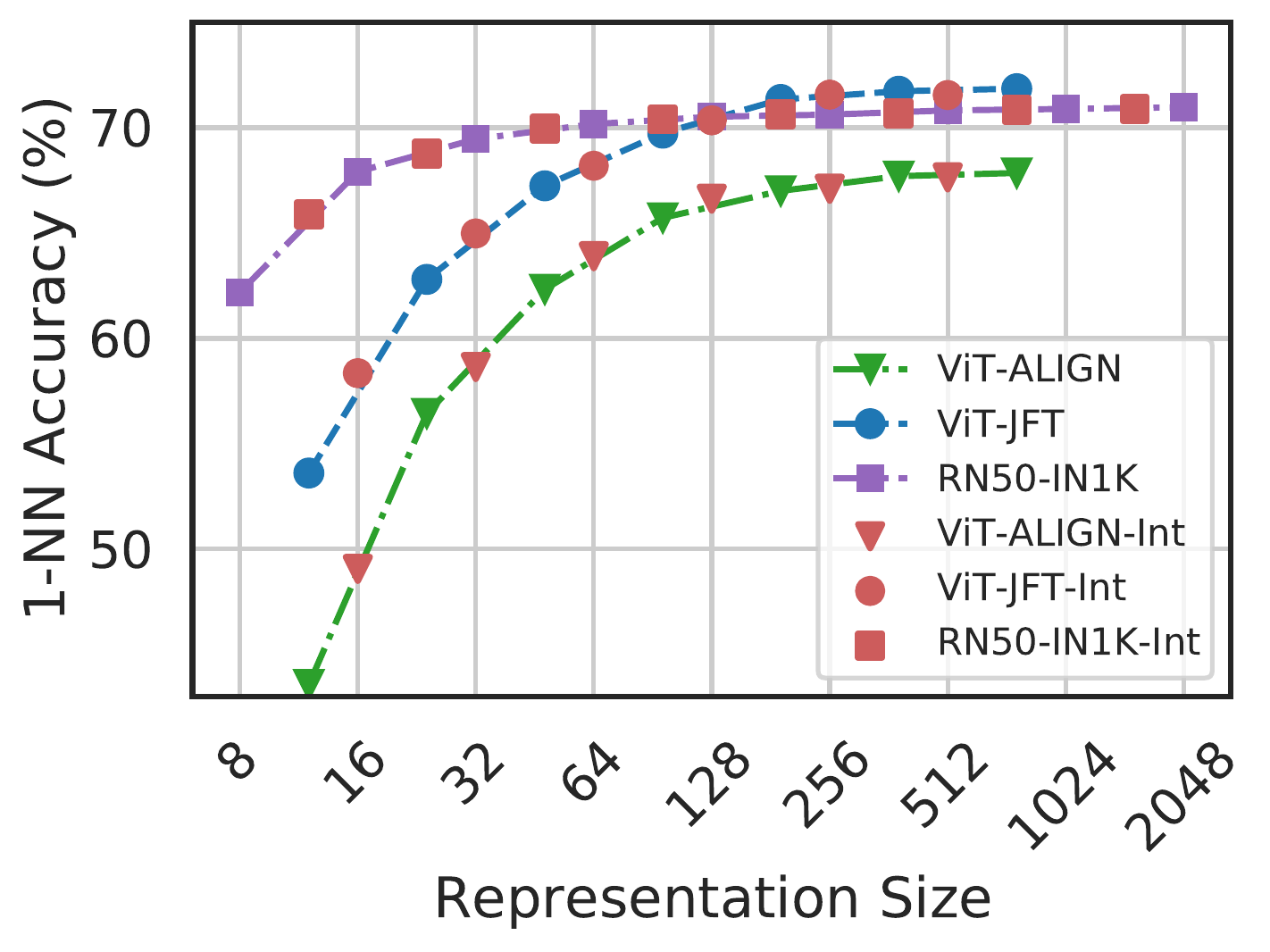}
        \caption{Despite optimizing \mrl only for $O(\log(d))$ dimensions for ResNet50 and ViT-B/16 models; the accuracy in the intermediate dimensions shows interpolating behaviour.}
  \label{fig:int-acc}
\end{minipage}%
\end{figure}

We also evaluate the quality of the representations from training ViT-B/16 on JFT-300M alongside the ViT-B/16 vision encoder of the ALIGN model -- two web-scale setups. Due to the expensive nature of these experiments, we only train the highest capacity fixed feature model and choose random features for evaluation in lower-dimensions. Web-scale is a compelling setting for \mrl due to its relatively inexpensive training overhead while providing multifidelity representations for downstream tasks. Figure~\ref{fig:ViT-acc}, evaluated with 1-NN on ImageNet-1K, shows that all the \mrl models for JFT and ALIGN are highly accurate while providing an excellent cost-vs-accuracy trade-off at lower-dimensions. These experiments show that \mrl seamlessly scales to large-scale models and web-scale datasets while providing the otherwise prohibitively expensive multi-granularity in the process. We also have similar observations when pretraining BERT; please see Appendix~\ref{sec:appendix_jft_align_bert} for more details. 
Our experiments also show that post-hoc compression (SVD), linear probe on random features, and sub-net style slimmable networks drastically lose accuracy compared to \mrl as the representation size decreases. Finally, Figure~\ref{fig:int-acc} shows that, while \mrl explicitly optimizes $O(\log(d))$ nested representations -- removing the $O(d)$ dependence~\citep{rippel2014learning} --, the coarse-to-fine grained information is interpolated across all $d$ dimensions providing highest flexibility for adaptive deployment.

\subsubsection{Adaptive Classification}
\label{sec:adaptive_classification_main}
The flexibility and coarse-to-fine granularity within \mrs allows model cascades~\citep{viola2001rapid} for Adaptive Classification (AC)~\citep{harris2000coarse}. Unlike standard model cascades~\citep{wang2020multiple}, \mrl does not require multiple expensive neural network forward passes. To perform AC with an \mrl trained model, we learn thresholds on the maximum softmax probability~\citep{hendrycks2016baseline} for each nested classifier on a holdout validation set. We then use these thresholds to decide when to transition to the higher dimensional representation (e.g $8\to16\to32$) of the \mrl model. Appendix~\ref{sec:appendix_adaptive_classification} discusses the implementation and learning of thresholds for cascades used for adaptive classification in detail. 

Figure~\ref{fig:r50-mrl-cascade-acc} shows the comparison between cascaded \mrl representations (\mrl--AC) and independently trained fixed feature (FF) models on ImageNet-1K with ResNet50. We computed the expected representation size for \mrl--AC based on the final dimensionality used in the cascade. We observed that \mrl--AC was as accurate, $76.30\%$, as a 512-dimensional FF model but required an expected dimensionality of $\sim37$ while being only $0.8\%$ lower than the 2048-dimensional FF baseline. Note that all \mrl--AC models are significantly more accurate than the FF baselines at comparable representation sizes. \mrl--AC uses up to $\sim14\times$ smaller representation size for the same accuracy which affords computational efficiency as the label space grows~\citep{varma2019extreme}. Lastly, our results with \mrl--AC indicate that instances and classes vary in difficulty which we analyze in Section~\ref{sec:analysis} and Appendix~\ref{app:Model_Disagree}.

\subsection{Retrieval}
\label{sec:retrieval}
Nearest neighbour search with learned representations powers a plethora of retrieval and search applications~\citep{dean2009challenges,Waldburger2019Search,chang2021extreme,NayakUnderstanding}. In this section, we discuss the image retrieval performance of the pretrained ResNet50 models (Section~\ref{sec:rep_learning}) on two large-scale datasets ImageNet-1K~\citep{russakovsky2015imagenet} and ImageNet-4K. ImageNet-1K has a database size of $\sim$1.3M and a query set of 50K samples uniformly spanning 1000 classes. We also introduce ImageNet-4K which has a database size of $\sim$4.2M and query set of $\sim$200K samples uniformly spanning 4202 classes (see Appendix~\ref{sec:datasets} for details). A single forward pass on ResNet50 costs 4 GFLOPs while exact retrieval costs 2.6 GFLOPs per query for ImageNet-1K. Although retrieval overhead is $40\%$ of the total cost, retrieval cost grows linearly with the size of the database. ImageNet-4K presents a retrieval benchmark where the exact search cost becomes the computational bottleneck ($8.6$ GFLOPs per query). In both these settings, the memory and disk usage are also often bottlenecked by the large databases. However, in most real-world applications exact search, $O(dN)$, is replaced with an approximate nearest neighbor search (ANNS) method like HNSW~\citep{malkov2018efficient}, $O(d\log(N))$, with minimal accuracy drop at the cost of additional memory overhead.

The goal of image retrieval is to find images that belong to the same class as the query using representations obtained from a pretrained model. In this section, we compare retrieval performance using mean Average Precision @ 10 (mAP@$10$) which comprehensively captures the setup of relevant image retrieval at scale. We measure the cost per query using exact search in MFLOPs. All embeddings are unit normalized and retrieved using the L2 distance metric. Lastly, we report an extensive set of metrics spanning mAP@$k$ and P@$k$ for $k=\{10,25,50,100\}$ and real-world wall-clock times for exact search and HNSW. See Appendices~\ref{sec:appendix-retrieval} and~\ref{sec:adaptive_retrieval} for more details.

\begin{figure}[t!]
\centering
\begin{minipage}{.48\columnwidth}
    \centering
      \hspace{-7mm}
      \vspace{3mm}
  \includegraphics[width=\columnwidth]{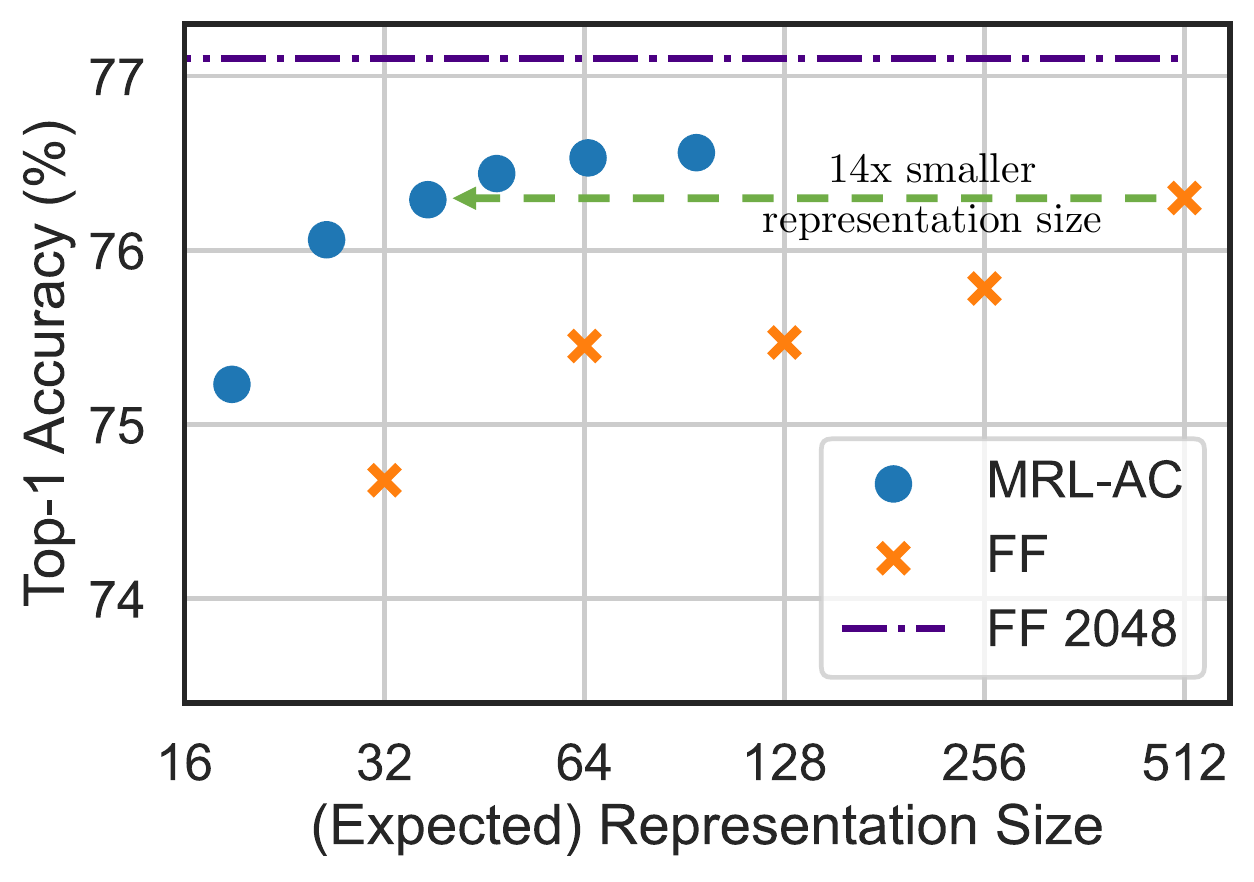}
    \caption{Adaptive classification on \mrl ResNet50 using cascades results in $14\times$ smaller representation size for the same level of accuracy on ImageNet-1K ($\sim37$ vs $512$ dims for $76.3\%$).}
  \label{fig:r50-mrl-cascade-acc}
\end{minipage}
\hspace{3mm}
\begin{minipage}{.48\columnwidth}
    \centering
  \includegraphics[width=\columnwidth]{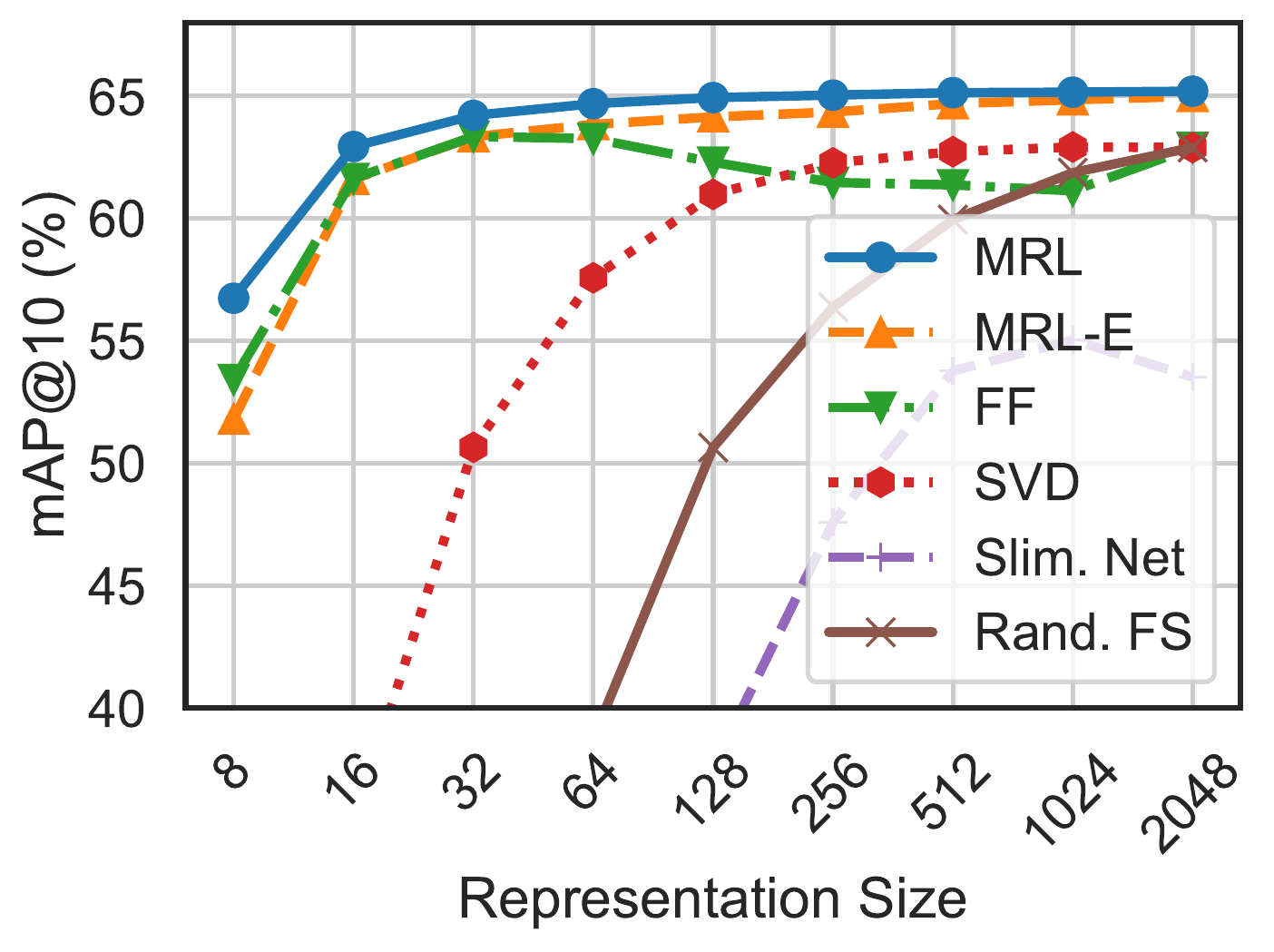}
    \caption{mAP@$10$ for Image Retrieval on ImageNet-1K with ResNet50. \mrl consistently produces better retrieval performance over the baselines across all the representation sizes.}
  \label{fig:r50-check}
\end{minipage}
\end{figure}
Figure~\ref{fig:r50-check} compares the mAP@$10$ performance of ResNet50 representations on ImageNet-1K across dimensionalities for \mrl, \SH, FF, slimmable networks along with post-hoc compression of vectors using SVD and random feature selection. \mrs are often the most accurate while being up to $3\%$ better than the FF baselines. Similar to classification, post-hoc compression and slimmable network baselines suffer from significant drop-off in retrieval mAP@$10$ with $\le256$ dimensions. Appendix~\ref{sec:appendix-retrieval} discusses the mAP@$10$ of the same models on ImageNet-4K. 

\mrl models are capable of performing accurate retrieval at various granularities without the additional expense of multiple model forward passes for the web-scale databases. FF models also generate independent databases which become prohibitively expense to store and switch in between. \mrs enable adaptive retrieval (AR) which alleviates the need to use full-capacity representations, $d=2048$, for all data and downstream tasks. Lastly, all the vector compression techniques~\citep{linde1980algorithm,jegou2010product} used as part of the ANNS pipelines are complimentary to \mrs and can further improve the  efficiency-vs-accuracy trade-off.

\subsubsection{Adaptive Retrieval}
\label{sec:adaptive_retrieval_main}

We benchmark \mrl in the adaptive retrieval setting (AR)~\cite{VertexAIMatchingEngine}. For a given query image, we obtained a shortlist, $K=200$, of images from the database using a lower-dimensional representation, e.g. $D_s=16$ followed by reranking with a higher capacity representation, e.g. $D_r=2048$. In real-world scenarios where top ranking performance is the key objective, measured with mAP@$k$ where k covers a limited yet crucial real-estate, AR provides significant compute and memory gains over single-shot retrieval with representations of fixed dimensionality. Finally, the most expensive part of AR, as with any retrieval pipeline, is the nearest neighbour search for shortlisting. For example, even naive re-ranking of 200 images with 2048 dimensions only costs 400 KFLOPs. While we report exact search cost per query for all AR experiments, the shortlisting component of the pipeline can be sped-up using ANNS (HNSW). Appendix~\ref{sec:appendix_real-world-perf} has a detailed discussion on compute cost for exact search, memory overhead of HNSW indices and wall-clock times for both implementations. We note that using HNSW with 32 neighbours for shortlisting does not decrease accuracy during retrieval. 

\begin{figure}[t!]
\centering
\resizebox{1\columnwidth}{!}{%
\begin{tabular}{ccc}
\includegraphics[width=0.48\columnwidth]{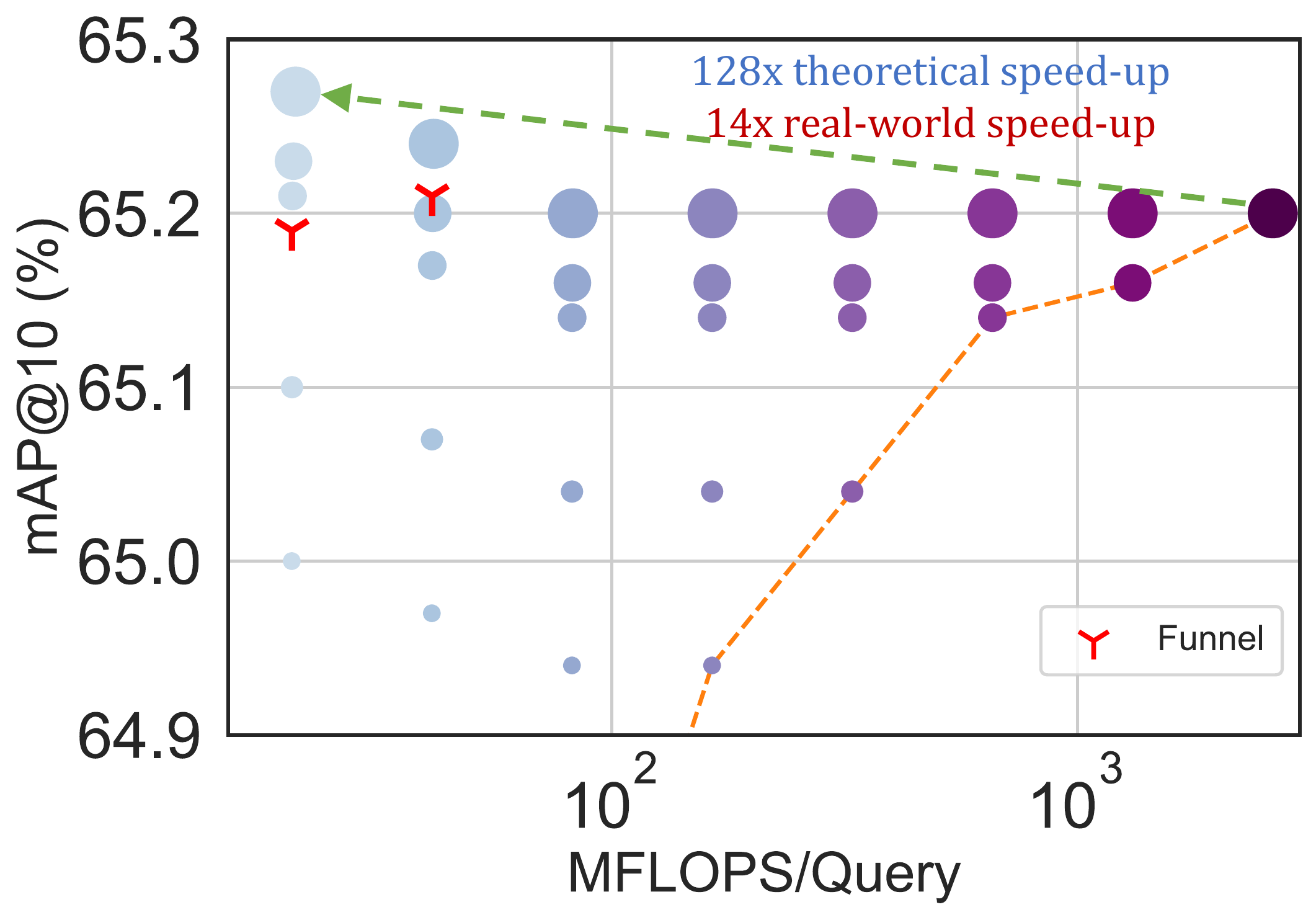}&
\raisebox{7mm}{
\includegraphics[width=0.075\columnwidth]{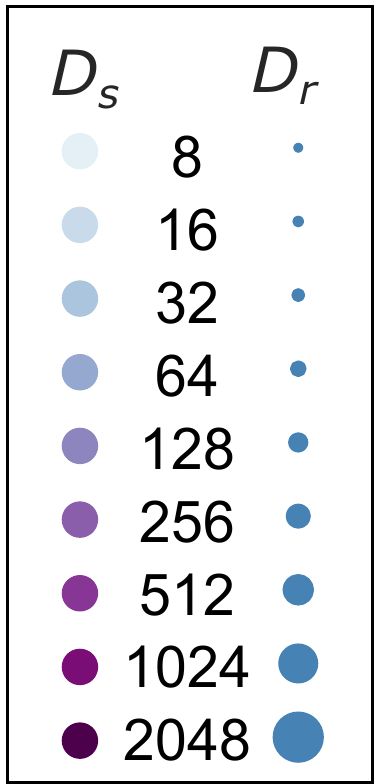}}&
\includegraphics[width=0.49\columnwidth]{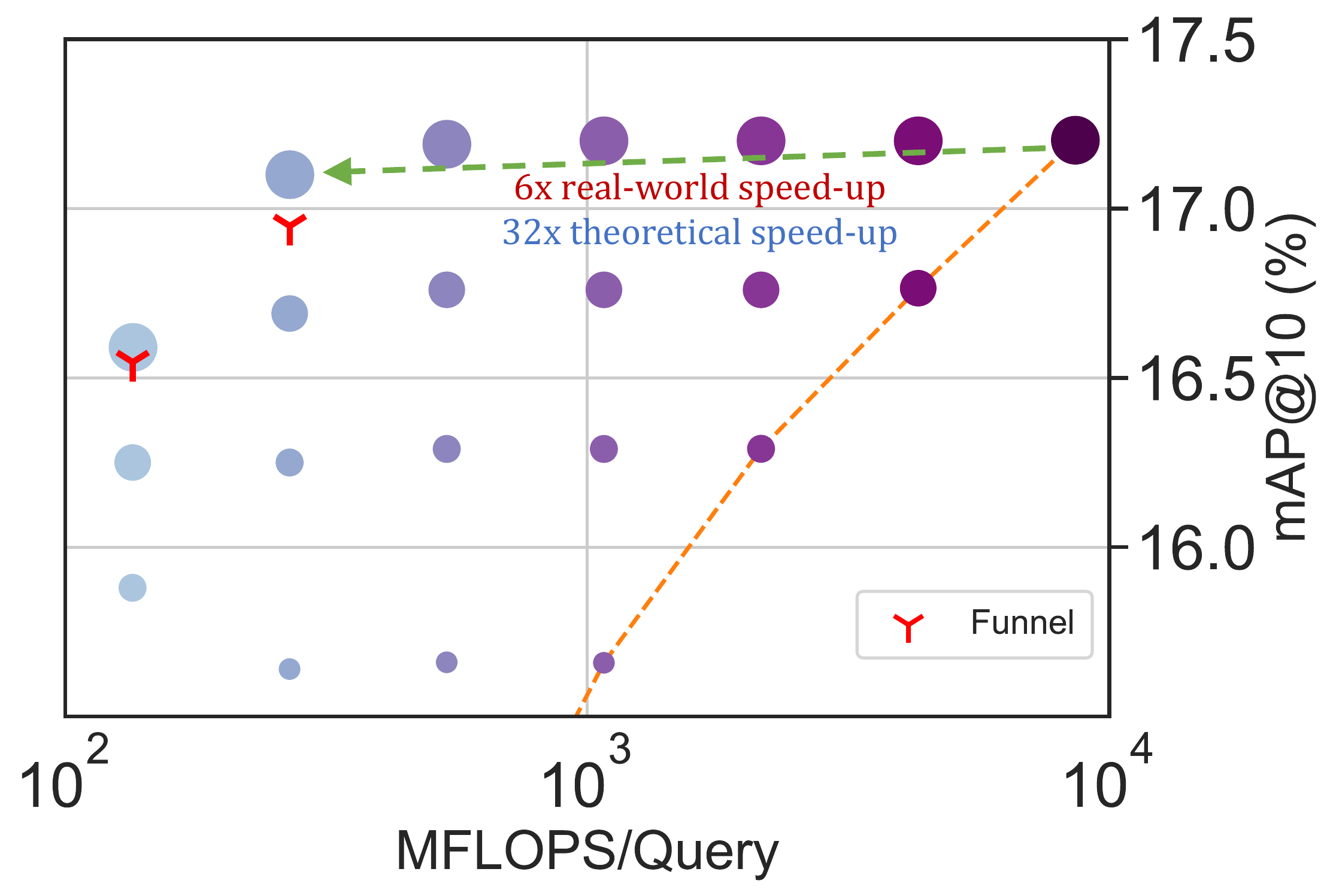}\\
     (a) ImageNet-1K&&(b) ImageNet-4K 
\end{tabular}
}
\caption{The trade-off between mAP@$10$ vs MFLOPs/Query for Adaptive Retrieval (AR) on ImageNet-1K (left) and ImageNet-4K (right). Every combination of $D_s$ \& $D_r$ falls above the Pareto line (orange dots) of single-shot retrieval with a fixed representation size while having configurations that are as accurate while being up to $14\times$ faster in real-world deployment. Funnel retrieval is almost as accurate as the baseline while alleviating some of the parameter choices of Adaptive Retrieval.} 
\label{fig:r50-retrieval}
\end{figure}
Figure~\ref{fig:r50-retrieval} showcases the compute-vs-accuracy trade-off for adaptive retrieval using \mrs compared to single-shot using fixed features with ResNet50 on ImageNet-1K. We observed that all AR settings lied above the Pareto frontier of single-shot retrieval with varying representation sizes. In particular for ImageNet-1K, we show that the AR model with $D_s=16$ \& $D_r=2048$ is as accurate as single-shot retrieval with $d=2048$ while being $\mathbf{\sim128\times}$ more efficient in theory and $\mathbf{\sim14\times}$ faster in practice (compared using HNSW on the same hardware). We show similar trends with \InIVk, but note that we require $D_s=64$ given the increased difficulty of the dataset. This results in $\sim32\times$ and $\sim6\times$ theoretical and in-practice speedups respectively. Lastly, while $K=200$ works well for our adaptive retrieval experiments, we ablated over the shortlist size $k$ in Appendix~\ref{sec:appendix_retrieval_ablation} and found that the accuracy gains stopped after a point, further strengthening the use-case for \alg and adaptive retrieval.

Even with adaptive retrieval, it is hard to determine the choice of $D_s$ \& $D_r$. In order to alleviate this issue to an extent, we propose \textbf{Funnel Retrieval}, a consistent cascade for adaptive retrieval. Funnel thins out the initial shortlist by a repeated re-ranking and shortlisting with a series of increasing capacity representations. Funnel halves the shortlist size and doubles the representation size at every step of re-ranking. For example on ImageNet-1K, a funnel with the shortlist progression of $200\to100\to50\to25\to10$ with the cascade of $16\to32\to64\to128\to256\to2048$ representation sizes within \mr is as accurate as the single-shot 2048-dim retrieval while being $\sim128\times$ more efficient theoretically (see Appendix~\ref{sec:adaptive_retrieval} for more results). All these results showcase the potential of \mrl and AR for large-scale multi-stage search systems~\citep{dean2009challenges}.

\section{Further Analysis and Ablations}
\label{sec:analysis}
\paragraph{Robustness.} We evaluate the robustness of the \mrl models trained on ImageNet-1K on out-of-domain datasets, ImageNetV2/R/A/Sketch~\citep{recht2019imagenet,hendrycks2021many,hendrycks2021natural,wang2019learning}, and compare them to the FF baselines. Table~\ref{tab:robustness} in Appendix~\ref{sec:appendix_robustness} demonstrates that \mrs for classification are at least as robust as the original representation while improving the performance on ImageNet-A by $0.6\%$ -- a $20\%$ relative improvement. We also study the robustness in the context of retrieval by using ImageNetV2 as the query set for ImageNet-1K database. Table~\ref{tab:retrieval_INV2} in Appendix~\ref{sec:appendix-retrieval} shows that \mrl models have more robust retrieval compared to the FF baselines by having up to $3\%$ higher mAP@$10$ performance. This observation also suggests the need for further investigation into robustness using nearest neighbour based classification and retrieval instead of the standard linear probing setup. We also find that the zero-shot robustness of ALIGN-\mrl (Table~\ref{tab:r50-align_zeroshot} in Appendix~\ref{sec:appendix_robustness})  agrees with the observations made by~\citet{wortsman2021robust}. Lastly, Table~\ref{tab:align-cosine-sim} in Appendix~\ref{sec:appendix_jft_align_bert} shows that \mrl also improves the cosine similarity span between positive and random image-text pairs.

\paragraph{Few-shot and Long-tail Learning.} We exhaustively evaluated few-shot learning on \mrl models using nearest class mean~\citep{sanchez1997use}. Table~\ref{Tab:FSL-INV2} in Appendix~\ref{sec:appendix_few_shot} shows that that representations learned through \mrl perform comparably to FF representations across varying shots and number of classes.

\mrs realize a unique pattern while evaluating on  FLUID~\citep{wallingford2020overfitting}, a long-tail sequential learning framework. We observed that \mrl provides up to $2\%$ accuracy higher on novel classes in the tail of the distribution, without sacrificing accuracy on other classes (Table \ref{tab:fluid} in Appendix~\ref{sec:appendix_few_shot}). Additionally we find the accuracy between low-dimensional and high-dimensional representations is marginal for pretrain classes. We hypothesize that the higher-dimensional representations are required to differentiate the classes when few training examples of each are known. This results provides further evidence that different tasks require varying capacity based on their difficulty.

\begin{figure}[t!]
\centering
    \begin{tabular}{ccc}
{\begin{sideways}
\hspace{13mm}(a) \end{sideways}}& \includegraphics[width=0.9\columnwidth]{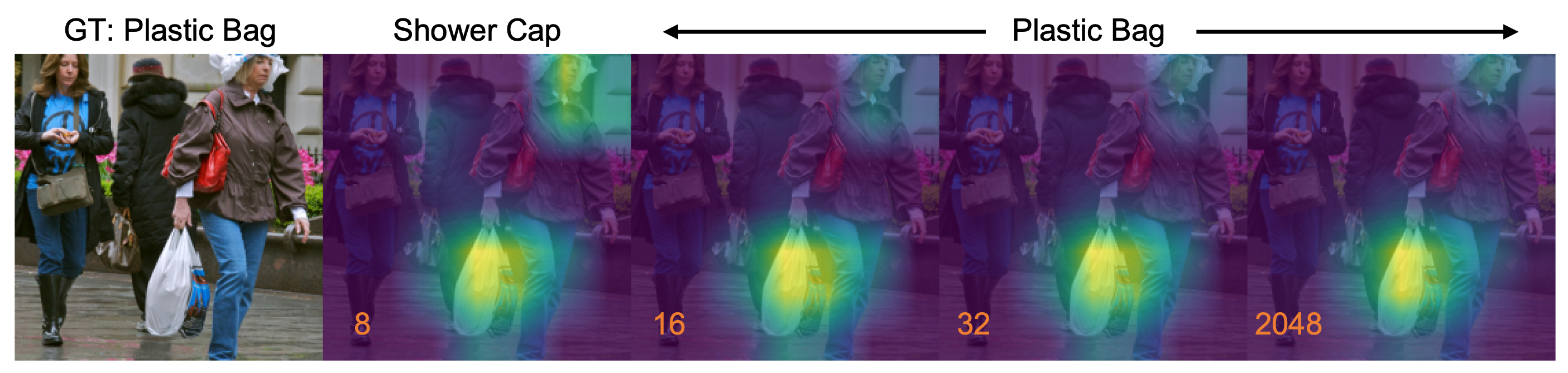}\\

{\begin{sideways}
\hspace{13mm}(b) \end{sideways}} &\includegraphics[width=0.9\columnwidth]{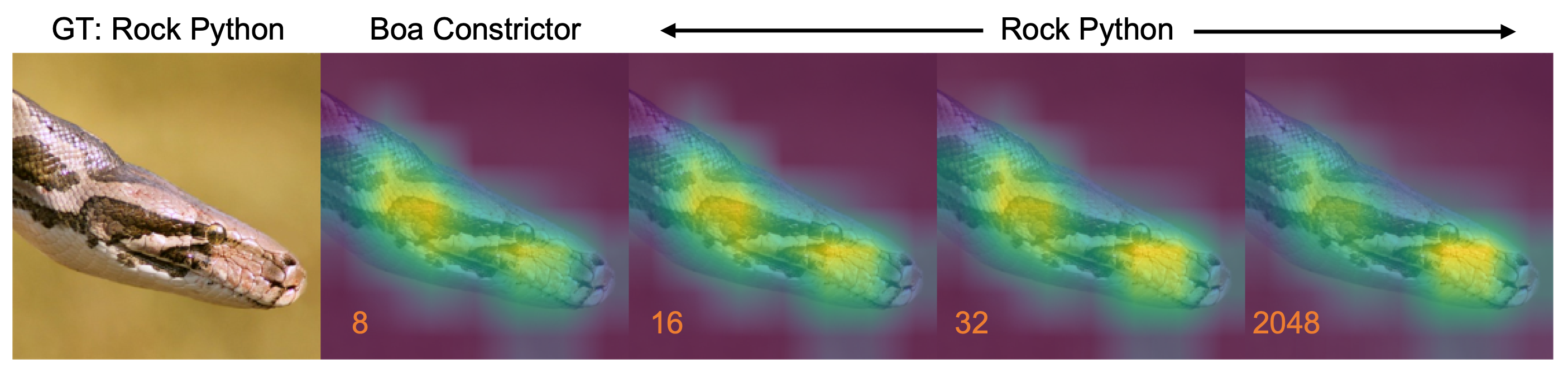}\\

{\begin{sideways}
\hspace{13mm}(c) \end{sideways}} &\includegraphics[width=0.9\columnwidth]{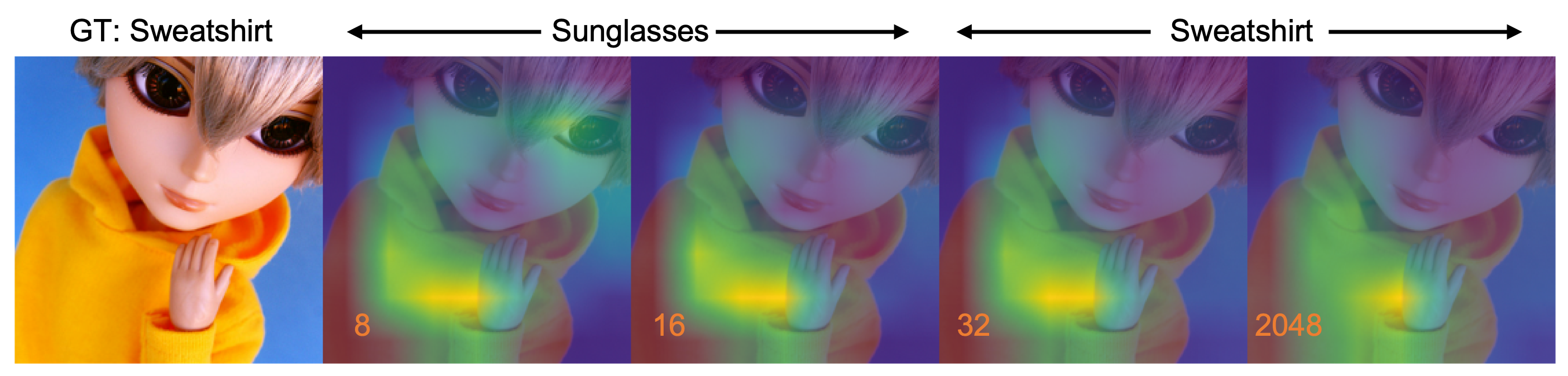}\\
    \end{tabular}

    \caption{Grad-CAM~\citep{selvaraju2017grad} progression of predictions in \nrl~model across $8, 16, 32 \text{ and } 2048$ dimensions. (a) $8$-dimensional representation confuses due to presence of other relevant objects (with a larger field of view) in the scene and predicts ``shower cap'' ; (b) $8$-dim model confuses within the same super-class of ``boa'' ; (c) $8$ and $16$-dim models incorrectly focus on the eyes of the doll ("sunglasses") and not the "sweatshirt" which is correctly in focus at higher dimensions; \mrl fails gracefully in these scenarios and shows potential use cases of disagreement across dimensions.}
\label{fig:r50-gradcam}
\end{figure}
\paragraph{Disagreement across Dimensions.} The information packing in \mrs often results in gradual increase of accuracy with increase in capacity. However, we observed that this trend was not ubiquitous and certain instances and classes were more accurate when evaluated with lower-dimensions (Figure~\ref{fig:r50-perclass} in Appendix~\ref{app:Model_Disagree}). With perfect routing of instances to appropriate dimension, \mrl can gain up to $4.6\%$ classification accuracy. At the same time, the low-dimensional models are less accurate either due to confusion within the same superclass~\citep{robustness} of the ImageNet hierarchy or presence of multiple objects of interest. Figure~\ref{fig:r50-gradcam} showcases 2 such examples for $8$-dimensional representation. These results along with Appendix~\ref{app:Model_Disagree} put forward the potential for \mrl to be a systematic framework for analyzing the utility and efficiency of information bottlenecks.


\begin{figure}[b!]
    \centering
    \begin{minipage}{.48\columnwidth}
        \includegraphics[width=\columnwidth]{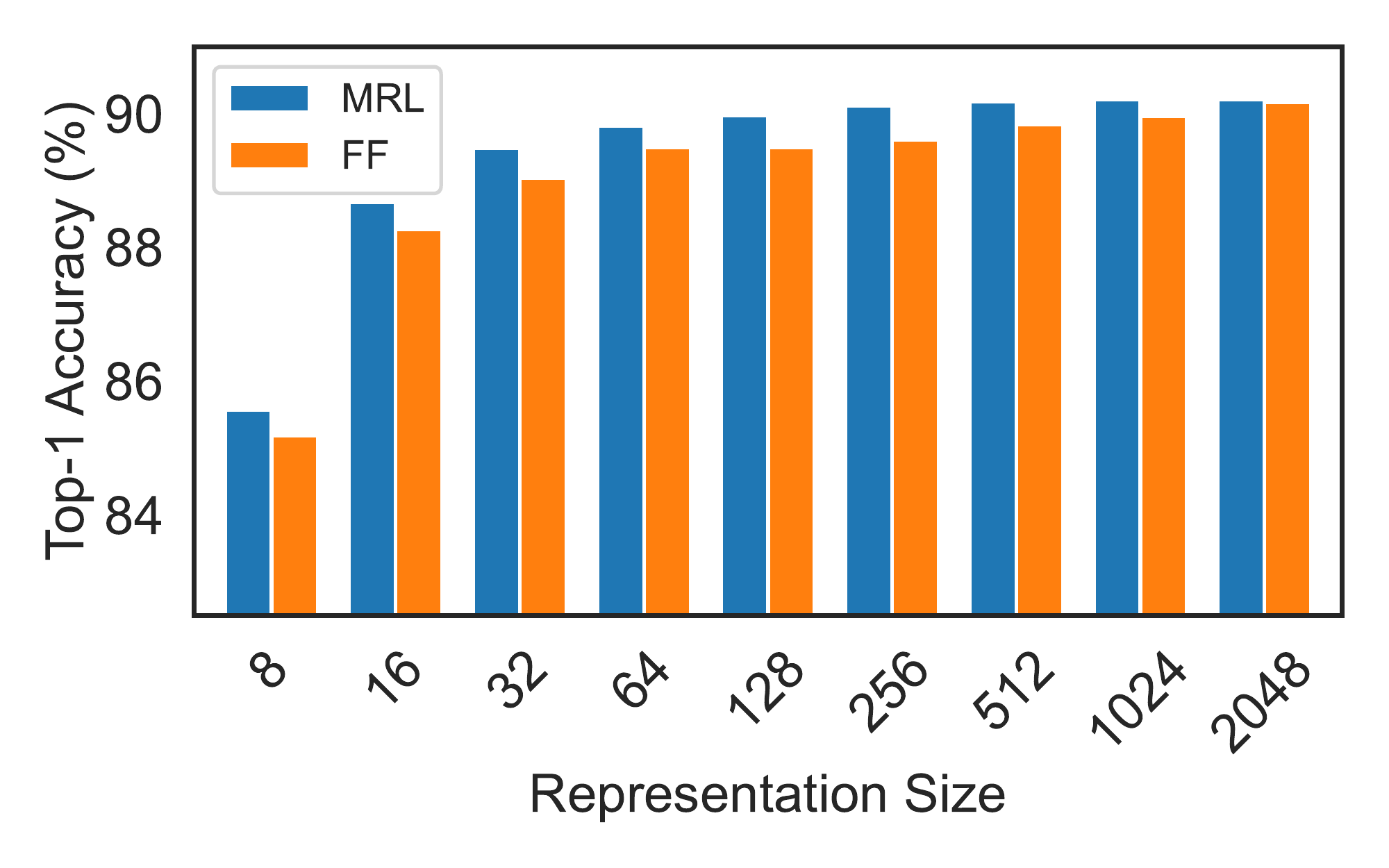}
        \caption{31-way ImageNet-1K superclass classification across representation size for \mrl \& FF models showing the capture of underlying hierarchy through tight information bottlenecks.}
        \label{fig:superclass_barplot_main}
    \end{minipage}
    \hspace{3mm}
    \begin{minipage}{.48\columnwidth}
    \includegraphics[width=\columnwidth]{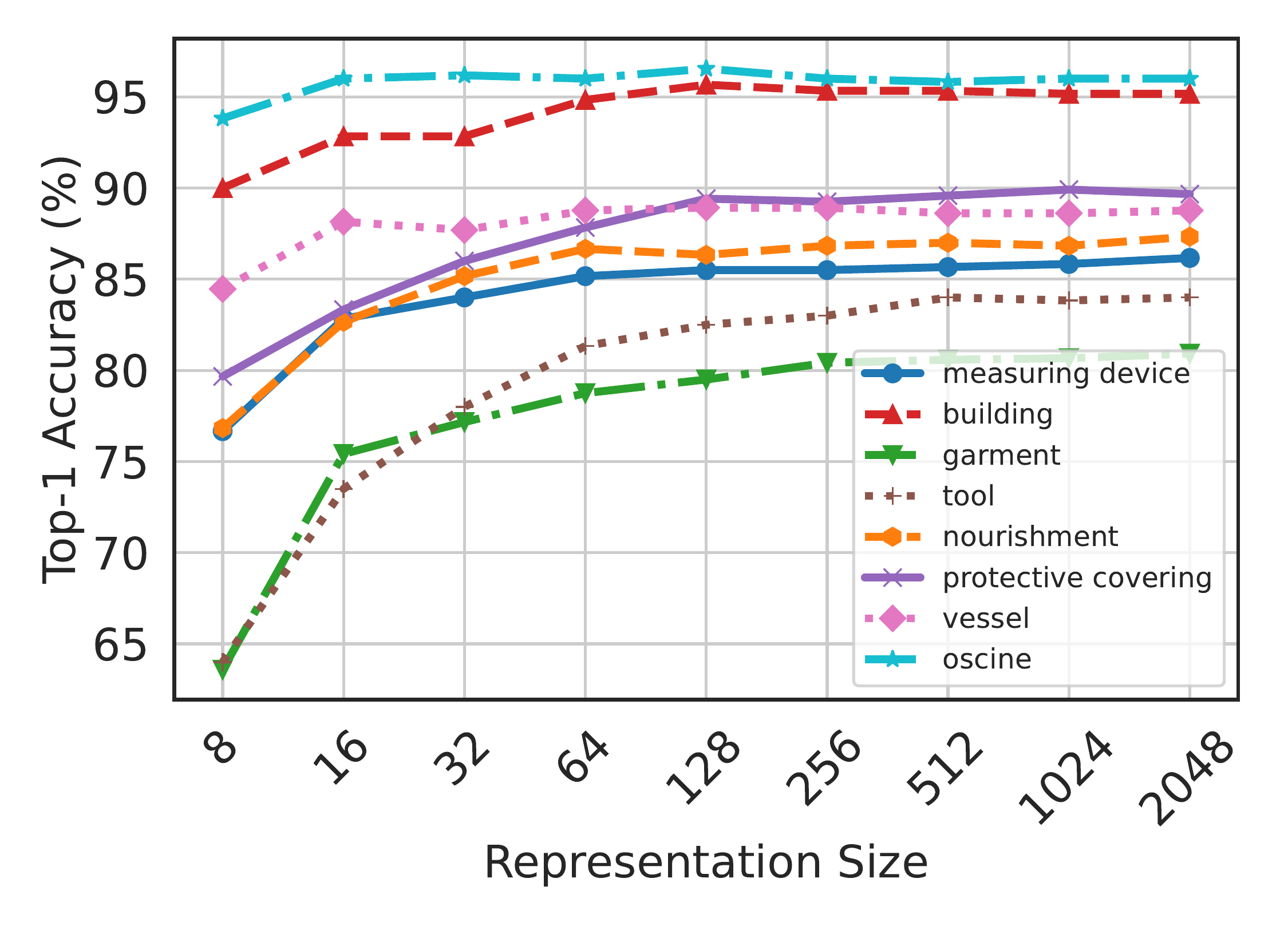}
    \vspace{-6mm}
    \caption{Diverse per-superclass accuracy trends across representation sizes for ResNet50-\mrl on ImageNet-1K.}
    \label{fig:superclass_main}
\end{minipage}

\end{figure}

\paragraph{Superclass Accuracy.} As the information bottleneck becomes smaller, the overall accuracy on fine-grained classes decreases rapidly (Figure~\ref{fig:r50-knn-acc}). However, the drop-off is not as significant when evaluated at a superclass level (Table~\ref{tab:superclass_names} in Appendix~\ref{app:Model_Disagree}). Figure~\ref{fig:superclass_barplot_main} presents that this phenomenon occurs with both \mrl and FF models; \mrl is more accurate across dimensions. This shows that tight information bottlenecks while not highly accurate for fine-grained classification, do capture required semantic information for coarser classification that could be leveraged for adaptive routing for retrieval and classification. Mutifidelity of \mr naturally captures the underlying hierarchy of the class labels with one single model. Lastly, Figure~\ref{fig:superclass_main} showcases the accuracy trends per superclass with \mrl. The utility of additional dimensions in distinguishing a class from others within the same superclass is evident for ``garment'' which has up to 11\% improvement for 8 $\to$ 16 dimensional representation transition. We also observed that superclasses such as ``oscine (songbird)'' had a clear visual distinction between the object and background and thus predictions using 8 dimensions also led to a good inter-class separability within the superclass.

\subsection{Ablations} 
\label{sec:Ablations}
Table~\ref{tab:nesting_as_finetuning} in Appendix~\ref{sec:ablation} presents that \mrs can be enabled within off-the-shelf pretrained models with inexpensive partial finetuning thus paving a way for ubiquitous adoption of \mrl. At the same time, Table~\ref{tab:ablation_trainloss} in Appendix~\ref{sec:appendix-mrl_model_training} indicates that with optimal weighting of the nested losses we could improve accuracy of lower-dimensions representations without accuracy loss. Tables~\ref{tab:ablation_train_smalldim} and~\ref{tab:ablation_uniform_nesting} in Appendix~\ref{sec:appendix-mrl_model_training} ablate over the choice of initial granularity and spacing of the granularites. Table~\ref{tab:ablation_train_smalldim} reaffirms the design choice to shun extremely low dimensions that have poor classification accuracy as initial granularity for \mrl while Table~\ref{tab:ablation_uniform_nesting} confirms the effectiveness of logarthmic granularity spacing inspired from the behaviour of accuracy saturation across dimensions over uniform. Lastly, Tables~\ref{tab:rerank_ablation_shortlist_in1k} and~\ref{tab:rerank_ablation_shortlist_in4k} in Appendix~\ref{sec:appendix_retrieval_ablation} show that the retrieval performance saturates after a certain shortlist dimension and length depending on the complexity of the dataset.

\section{Discussion and Conclusions}
\label{sec:conc}

The results in Section \ref{sec:Ablations} reveal interesting weaknesses of \mrl that would be logical directions for future work. (1) Optimizing the weightings of the nested losses to obtain a Pareto optimal accuracy-vs-efficiency trade-off -- a potential solution could emerge from adaptive loss balancing aspects of anytime neural networks~\citep{hu2019learning}. (2) Using different losses at various fidelities aimed at solving a specific aspect of adaptive deployment -- e.g. high recall for $8$-dimension and robustness for $2048$-dimension. (3) Learning a search data-structure, like differentiable k-d tree, on top of \mr to enable dataset and representation aware retrieval. (4) Finally, the joint optimization of multi-objective \mrl combined with end-to-end learnable search data-structure to have data-driven adaptive large-scale retrieval for web-scale search applications. 

In conclusion, we presented \mdoll~\alg~(\mrl), a flexible representation learning approach that encodes information at multiple granularities in a single embedding vector. This enables the \mrl to adapt to a downstream task's statistical complexity as well as the  available compute resources. We demonstrate that \mrl can be used for large-scale adaptive classification as well as adaptive retrieval. On standard benchmarks, \mrl matches the accuracy of the fixed-feature baseline despite using $14\times$ smaller representation size on average. Furthermore, the \mr based adaptive shortlisting and re-ranking system ensures comparable mAP@$10$ to the baseline while being $128\times$ cheaper in FLOPs and $14\times$ faster in wall-clock time. Finally, most of the efficiency techniques for model inference and vector search are complementary to \mrl~\mdoll~further assisting in deployment at the compute-extreme environments.
\section*{Acknowledgments}

We are grateful to Srinadh Bhojanapalli, Lovish Madaan, Raghav Somani, Ludwig Schmidt, and Venkata Sailesh Sanampudi for helpful discussions and feedback. Aditya Kusupati also thanks Tom Duerig and Rahul Sukthankar for their support. Part of the paper's large-scale experimentation is supported through a research GCP credit award from Google Cloud and Google Research. Gantavya Bhatt is supported in part by the CONIX Research Center, one of six centers in JUMP, a Semiconductor Research Corporation (SRC) program sponsored by DARPA. Sham Kakade acknowledges funding from the NSF award CCF-1703574 and ONR N00014-22-1-2377. Ali Farhadi acknowledges funding from the NSF awards IIS 1652052, IIS 17303166, DARPA N66001-19-2-4031, DARPA W911NF-15-1-0543 and gifts from Allen Institute for Artificial Intelligence.

\bibliography{local}

\begin{thebibliography}{102}
\providecommand{\natexlab}[1]{#1}
\providecommand{\url}[1]{\texttt{#1}}
\expandafter\ifx\csname urlstyle\endcsname\relax
  \providecommand{\doi}[1]{doi: #1}\else
  \providecommand{\doi}{doi: \begingroup \urlstyle{rm}\Url}\fi

\bibitem[Abadi et~al.(2015)Abadi, Agarwal, Barham, Brevdo, Chen, Citro,
  Corrado, Davis, Dean, Devin, Ghemawat, Goodfellow, Harp, Irving, Isard, Jia,
  Jozefowicz, Kaiser, Kudlur, Levenberg, Man\'{e}, Monga, Moore, Murray, Olah,
  Schuster, Shlens, Steiner, Sutskever, Talwar, Tucker, Vanhoucke, Vasudevan,
  Vi\'{e}gas, Vinyals, Warden, Wattenberg, Wicke, Yu, and
  Zheng]{tensorflow2015-whitepaper}
M.~Abadi, A.~Agarwal, P.~Barham, E.~Brevdo, Z.~Chen, C.~Citro, G.~S. Corrado,
  A.~Davis, J.~Dean, M.~Devin, S.~Ghemawat, I.~Goodfellow, A.~Harp, G.~Irving,
  M.~Isard, Y.~Jia, R.~Jozefowicz, L.~Kaiser, M.~Kudlur, J.~Levenberg,
  D.~Man\'{e}, R.~Monga, S.~Moore, D.~Murray, C.~Olah, M.~Schuster, J.~Shlens,
  B.~Steiner, I.~Sutskever, K.~Talwar, P.~Tucker, V.~Vanhoucke, V.~Vasudevan,
  F.~Vi\'{e}gas, O.~Vinyals, P.~Warden, M.~Wattenberg, M.~Wicke, Y.~Yu, and
  X.~Zheng.
\newblock {TensorFlow}: Large-scale machine learning on heterogeneous systems,
  2015.
\newblock URL \url{https://www.tensorflow.org/}.
\newblock Software available from tensorflow.org.

\bibitem[Barbu et~al.(2019)Barbu, Mayo, Alverio, Luo, Wang, Gutfreund,
  Tenenbaum, and Katz]{barbu2019objectnet}
A.~Barbu, D.~Mayo, J.~Alverio, W.~Luo, C.~Wang, D.~Gutfreund, J.~Tenenbaum, and
  B.~Katz.
\newblock Objectnet: A large-scale bias-controlled dataset for pushing the
  limits of object recognition models.
\newblock \emph{Advances in neural information processing systems}, 32, 2019.

\bibitem[Bengio et~al.(2010)Bengio, Weston, and Grangier]{bengio2010label}
S.~Bengio, J.~Weston, and D.~Grangier.
\newblock Label embedding trees for large multi-class tasks.
\newblock \emph{Advances in Neural Information Processing Systems}, 23, 2010.

\bibitem[Bengio(2012)]{bengio2012deep}
Y.~Bengio.
\newblock Deep learning of representations for unsupervised and transfer
  learning.
\newblock In \emph{Proceedings of ICML workshop on unsupervised and transfer
  learning}, pages 17--36. JMLR Workshop and Conference Proceedings, 2012.

\bibitem[Bentley(1990)]{bentley1990k}
J.~L. Bentley.
\newblock K-d trees for semidynamic point sets.
\newblock In \emph{Proceedings of the sixth annual symposium on Computational
  geometry}, pages 187--197, 1990.

\bibitem[Beygelzimer et~al.(2006)Beygelzimer, Kakade, and
  Langford]{beygelzimer2006cover}
A.~Beygelzimer, S.~Kakade, and J.~Langford.
\newblock Cover trees for nearest neighbor.
\newblock In \emph{Proceedings of the 23rd international conference on Machine
  learning}, pages 97--104, 2006.

\bibitem[Brin and Page(1998)]{brin1998anatomy}
S.~Brin and L.~Page.
\newblock The anatomy of a large-scale hypertextual web search engine.
\newblock \emph{Computer networks and ISDN systems}, 30\penalty0
  (1-7):\penalty0 107--117, 1998.

\bibitem[Brown et~al.(2020)Brown, Mann, Ryder, Subbiah, Kaplan, Dhariwal,
  Neelakantan, Shyam, Sastry, Askell, et~al.]{brown2020language}
T.~Brown, B.~Mann, N.~Ryder, M.~Subbiah, J.~D. Kaplan, P.~Dhariwal,
  A.~Neelakantan, P.~Shyam, G.~Sastry, A.~Askell, et~al.
\newblock Language models are few-shot learners.
\newblock \emph{Advances in neural information processing systems},
  33:\penalty0 1877--1901, 2020.

\bibitem[Cai et~al.(2019)Cai, Gan, Wang, Zhang, and Han]{cai2019once}
H.~Cai, C.~Gan, T.~Wang, Z.~Zhang, and S.~Han.
\newblock Once-for-all: Train one network and specialize it for efficient
  deployment.
\newblock \emph{arXiv preprint arXiv:1908.09791}, 2019.

\bibitem[Chang et~al.(2020)Chang, Yu, Chang, Yang, and Kumar]{chang2020pre}
W.-C. Chang, F.~X. Yu, Y.-W. Chang, Y.~Yang, and S.~Kumar.
\newblock Pre-training tasks for embedding-based large-scale retrieval.
\newblock \emph{arXiv preprint arXiv:2002.03932}, 2020.

\bibitem[Chang et~al.(2021)Chang, Jiang, Yu, Teo, Zhang, Zhong, Kolluri, Hu,
  Shandilya, Ievgrafov, et~al.]{chang2021extreme}
W.-C. Chang, D.~Jiang, H.-F. Yu, C.~H. Teo, J.~Zhang, K.~Zhong, K.~Kolluri,
  Q.~Hu, N.~Shandilya, V.~Ievgrafov, et~al.
\newblock Extreme multi-label learning for semantic matching in product search.
\newblock In \emph{Proceedings of the 27th ACM SIGKDD Conference on Knowledge
  Discovery \& Data Mining}, pages 2643--2651, 2021.

\bibitem[Chen et~al.(2020)Chen, Kornblith, Norouzi, and Hinton]{chen2020simple}
T.~Chen, S.~Kornblith, M.~Norouzi, and G.~Hinton.
\newblock A simple framework for contrastive learning of visual
  representations.
\newblock In \emph{International conference on machine learning}, pages
  1597--1607. PMLR, 2020.

\bibitem[Chen et~al.(2021)Chen, Liu, Xu, Darrell, and Wang]{chen2021meta}
Y.~Chen, Z.~Liu, H.~Xu, T.~Darrell, and X.~Wang.
\newblock Meta-baseline: exploring simple meta-learning for few-shot learning.
\newblock In \emph{Proceedings of the IEEE/CVF International Conference on
  Computer Vision}, pages 9062--9071, 2021.

\bibitem[Datar et~al.(2004)Datar, Immorlica, Indyk, and
  Mirrokni]{datar2004locality}
M.~Datar, N.~Immorlica, P.~Indyk, and V.~S. Mirrokni.
\newblock Locality-sensitive hashing scheme based on p-stable distributions.
\newblock In \emph{Proceedings of the twentieth annual symposium on
  Computational geometry}, pages 253--262, 2004.

\bibitem[Dean(2009)]{dean2009challenges}
J.~Dean.
\newblock Challenges in building large-scale information retrieval systems.
\newblock In \emph{Keynote of the 2nd ACM International Conference on Web
  Search and Data Mining (WSDM)}, volume~10, 2009.

\bibitem[Deng et~al.(2009)Deng, Dong, Socher, Li, Li, and
  Fei-Fei]{deng2009imagenet}
J.~Deng, W.~Dong, R.~Socher, L.-J. Li, K.~Li, and L.~Fei-Fei.
\newblock Imagenet: A large-scale hierarchical image database.
\newblock In \emph{2009 IEEE conference on computer vision and pattern
  recognition}, pages 248--255. Ieee, 2009.

\bibitem[Deng et~al.(2011)Deng, Berg, and Fei-Fei]{deng2011hierarchical}
J.~Deng, A.~C. Berg, and L.~Fei-Fei.
\newblock Hierarchical semantic indexing for large scale image retrieval.
\newblock In \emph{CVPR 2011}, pages 785--792. IEEE, 2011.

\bibitem[Desai and Johnson(2021)]{desai2021virtex}
K.~Desai and J.~Johnson.
\newblock Virtex: Learning visual representations from textual annotations.
\newblock In \emph{Proceedings of the IEEE/CVF Conference on Computer Vision
  and Pattern Recognition}, pages 11162--11173, 2021.

\bibitem[Devlin et~al.(2018)Devlin, Chang, Lee, and Toutanova]{devlin2018bert}
J.~Devlin, M.-W. Chang, K.~Lee, and K.~Toutanova.
\newblock Bert: Pre-training of deep bidirectional transformers for language
  understanding.
\newblock \emph{arXiv preprint arXiv:1810.04805}, 2018.

\bibitem[Dietterich and Bakiri(1994)]{dietterich1994solving}
T.~G. Dietterich and G.~Bakiri.
\newblock Solving multiclass learning problems via error-correcting output
  codes.
\newblock \emph{Journal of artificial intelligence research}, 2:\penalty0
  263--286, 1994.

\bibitem[Divvala et~al.(2014)Divvala, Farhadi, and
  Guestrin]{divvala2014learning}
S.~K. Divvala, A.~Farhadi, and C.~Guestrin.
\newblock Learning everything about anything: Webly-supervised visual concept
  learning.
\newblock In \emph{Proceedings of the IEEE Conference on Computer Vision and
  Pattern Recognition}, pages 3270--3277, 2014.

\bibitem[Dosovitskiy et~al.(2020)Dosovitskiy, Beyer, Kolesnikov, Weissenborn,
  Zhai, Unterthiner, Dehghani, Minderer, Heigold, Gelly,
  et~al.]{dosovitskiy2020image}
A.~Dosovitskiy, L.~Beyer, A.~Kolesnikov, D.~Weissenborn, X.~Zhai,
  T.~Unterthiner, M.~Dehghani, M.~Minderer, G.~Heigold, S.~Gelly, et~al.
\newblock An image is worth 16x16 words: Transformers for image recognition at
  scale.
\newblock \emph{arXiv preprint arXiv:2010.11929}, 2020.

\bibitem[Engelsma et~al.(2022)Engelsma, Jain, and Boddeti]{engelsma2022hers}
J.~J. Engelsma, A.~K. Jain, and V.~N. Boddeti.
\newblock Hers: Homomorphically encrypted representation search.
\newblock \emph{IEEE Transactions on Biometrics, Behavior, and Identity
  Science}, 4\penalty0 (3):\penalty0 349--360, 2022.

\bibitem[Engstrom et~al.(2019)Engstrom, Ilyas, Salman, Santurkar, and
  Tsipras]{robustness}
L.~Engstrom, A.~Ilyas, H.~Salman, S.~Santurkar, and D.~Tsipras.
\newblock Robustness (python library), 2019.
\newblock URL \url{https://github.com/MadryLab/robustness}.

\bibitem[Gholami et~al.(2021)Gholami, Kim, Dong, Yao, Mahoney, and
  Keutzer]{gholami2021survey}
A.~Gholami, S.~Kim, Z.~Dong, Z.~Yao, M.~W. Mahoney, and K.~Keutzer.
\newblock A survey of quantization methods for efficient neural network
  inference.
\newblock \emph{arXiv preprint arXiv:2103.13630}, 2021.

\bibitem[Gong et~al.(2019)Gong, Boddeti, and Jain]{gong2019intrinsic}
S.~Gong, V.~N. Boddeti, and A.~K. Jain.
\newblock On the intrinsic dimensionality of image representations.
\newblock In \emph{Proceedings of the IEEE/CVF Conference on Computer Vision
  and Pattern Recognition}, pages 3987--3996, 2019.

\bibitem[Gutmann and Hyv{\"a}rinen(2010)]{gutmann2010noise}
M.~Gutmann and A.~Hyv{\"a}rinen.
\newblock Noise-contrastive estimation: A new estimation principle for
  unnormalized statistical models.
\newblock In \emph{Proceedings of the thirteenth international conference on
  artificial intelligence and statistics}, pages 297--304. JMLR Workshop and
  Conference Proceedings, 2010.

\bibitem[Harris and Giachritsis(2000)]{harris2000coarse}
M.~G. Harris and C.~D. Giachritsis.
\newblock Coarse-grained information dominates fine-grained information in
  judgments of time-to-contact from retinal flow.
\newblock \emph{Vision research}, 40\penalty0 (6):\penalty0 601--611, 2000.

\bibitem[He et~al.(2016)He, Zhang, Ren, and Sun]{he2016deep}
K.~He, X.~Zhang, S.~Ren, and J.~Sun.
\newblock Deep residual learning for image recognition.
\newblock In \emph{Proceedings of the IEEE conference on computer vision and
  pattern recognition}, pages 770--778, 2016.

\bibitem[He et~al.(2020)He, Fan, Wu, Xie, and Girshick]{he2020momentum}
K.~He, H.~Fan, Y.~Wu, S.~Xie, and R.~Girshick.
\newblock Momentum contrast for unsupervised visual representation learning.
\newblock In \emph{Proceedings of the IEEE/CVF conference on computer vision
  and pattern recognition}, pages 9729--9738, 2020.

\bibitem[He et~al.(2021)He, Chen, Xie, Li, Doll{\'a}r, and
  Girshick]{he2021masked}
K.~He, X.~Chen, S.~Xie, Y.~Li, P.~Doll{\'a}r, and R.~Girshick.
\newblock Masked autoencoders are scalable vision learners.
\newblock \emph{arXiv preprint arXiv:2111.06377}, 2021.

\bibitem[Hegd{\'e}(2008)]{hegde2008time}
J.~Hegd{\'e}.
\newblock Time course of visual perception: coarse-to-fine processing and
  beyond.
\newblock \emph{Progress in neurobiology}, 84\penalty0 (4):\penalty0 405--439,
  2008.

\bibitem[Hendrycks and Gimpel(2016)]{hendrycks2016baseline}
D.~Hendrycks and K.~Gimpel.
\newblock A baseline for detecting misclassified and out-of-distribution
  examples in neural networks.
\newblock \emph{arXiv preprint arXiv:1610.02136}, 2016.

\bibitem[Hendrycks et~al.(2021{\natexlab{a}})Hendrycks, Basart, Mu, Kadavath,
  Wang, Dorundo, Desai, Zhu, Parajuli, Guo, et~al.]{hendrycks2021many}
D.~Hendrycks, S.~Basart, N.~Mu, S.~Kadavath, F.~Wang, E.~Dorundo, R.~Desai,
  T.~Zhu, S.~Parajuli, M.~Guo, et~al.
\newblock The many faces of robustness: A critical analysis of
  out-of-distribution generalization.
\newblock In \emph{Proceedings of the IEEE/CVF International Conference on
  Computer Vision}, pages 8340--8349, 2021{\natexlab{a}}.

\bibitem[Hendrycks et~al.(2021{\natexlab{b}})Hendrycks, Zhao, Basart,
  Steinhardt, and Song]{hendrycks2021natural}
D.~Hendrycks, K.~Zhao, S.~Basart, J.~Steinhardt, and D.~Song.
\newblock Natural adversarial examples.
\newblock In \emph{Proceedings of the IEEE/CVF Conference on Computer Vision
  and Pattern Recognition}, pages 15262--15271, 2021{\natexlab{b}}.

\bibitem[Hooker et~al.(2019)Hooker, Courville, Clark, Dauphin, and
  Frome]{hooker2019compressed}
S.~Hooker, A.~Courville, G.~Clark, Y.~Dauphin, and A.~Frome.
\newblock What do compressed deep neural networks forget?
\newblock \emph{arXiv preprint arXiv:1911.05248}, 2019.

\bibitem[Hooker et~al.(2020)Hooker, Moorosi, Clark, Bengio, and
  Denton]{hooker2020characterising}
S.~Hooker, N.~Moorosi, G.~Clark, S.~Bengio, and E.~Denton.
\newblock Characterising bias in compressed models.
\newblock \emph{arXiv preprint arXiv:2010.03058}, 2020.

\bibitem[Hotelling(1933)]{hotelling1933analysis}
H.~Hotelling.
\newblock Analysis of a complex of statistical variables into principal
  components.
\newblock \emph{Journal of educational psychology}, 24\penalty0 (6):\penalty0
  417, 1933.

\bibitem[Howard et~al.(2017)Howard, Zhu, Chen, Kalenichenko, Wang, Weyand,
  Andreetto, and Adam]{howard2017mobilenets}
A.~G. Howard, M.~Zhu, B.~Chen, D.~Kalenichenko, W.~Wang, T.~Weyand,
  M.~Andreetto, and H.~Adam.
\newblock Mobilenets: Efficient convolutional neural networks for mobile vision
  applications.
\newblock \emph{arXiv preprint arXiv:1704.04861}, 2017.

\bibitem[Howard and Ruder(2018)]{howard2018universal}
J.~Howard and S.~Ruder.
\newblock Universal language model fine-tuning for text classification.
\newblock \emph{arXiv preprint arXiv:1801.06146}, 2018.

\bibitem[Hu et~al.(2019)Hu, Dey, Hebert, and Bagnell]{hu2019learning}
H.~Hu, D.~Dey, M.~Hebert, and J.~A. Bagnell.
\newblock Learning anytime predictions in neural networks via adaptive loss
  balancing.
\newblock In \emph{Proceedings of the AAAI Conference on Artificial
  Intelligence}, volume~33, pages 3812--3821, 2019.

\bibitem[Indyk and Motwani(1998)]{indyk1998approximate}
P.~Indyk and R.~Motwani.
\newblock Approximate nearest neighbors: towards removing the curse of
  dimensionality.
\newblock In \emph{Proceedings of the thirtieth annual ACM symposium on Theory
  of computing}, pages 604--613, 1998.

\bibitem[Jain et~al.(2019)Jain, Balasubramanian, Chunduri, and
  Varma]{jain2019slice}
H.~Jain, V.~Balasubramanian, B.~Chunduri, and M.~Varma.
\newblock Slice: Scalable linear extreme classifiers trained on 100 million
  labels for related searches.
\newblock In \emph{Proceedings of the Twelfth ACM International Conference on
  Web Search and Data Mining}, pages 528--536, 2019.

\bibitem[Jayaram~Subramanya et~al.(2019)Jayaram~Subramanya, Devvrit, Simhadri,
  Krishnawamy, and Kadekodi]{jayaram2019diskann}
S.~Jayaram~Subramanya, F.~Devvrit, H.~V. Simhadri, R.~Krishnawamy, and
  R.~Kadekodi.
\newblock Diskann: Fast accurate billion-point nearest neighbor search on a
  single node.
\newblock \emph{Advances in Neural Information Processing Systems}, 32, 2019.

\bibitem[Jegou et~al.(2010)Jegou, Douze, and Schmid]{jegou2010product}
H.~Jegou, M.~Douze, and C.~Schmid.
\newblock Product quantization for nearest neighbor search.
\newblock \emph{IEEE transactions on pattern analysis and machine
  intelligence}, 33\penalty0 (1):\penalty0 117--128, 2010.

\bibitem[Jia et~al.(2021)Jia, Yang, Xia, Chen, Parekh, Pham, Le, Sung, Li, and
  Duerig]{jia2021scaling}
C.~Jia, Y.~Yang, Y.~Xia, Y.-T. Chen, Z.~Parekh, H.~Pham, Q.~Le, Y.-H. Sung,
  Z.~Li, and T.~Duerig.
\newblock Scaling up visual and vision-language representation learning with
  noisy text supervision.
\newblock In \emph{International Conference on Machine Learning}, pages
  4904--4916. PMLR, 2021.

\bibitem[Johnson et~al.(2019)Johnson, Douze, and J{\'e}gou]{johnson2019billion}
J.~Johnson, M.~Douze, and H.~J{\'e}gou.
\newblock Billion-scale similarity search with {GPUs}.
\newblock \emph{IEEE Transactions on Big Data}, 7\penalty0 (3):\penalty0
  535--547, 2019.

\bibitem[Johnson(1984)]{johnson1984extensions}
W.~B. Johnson.
\newblock Extensions of lipschitz mappings into a hilbert space.
\newblock \emph{Contemp. Math.}, 26:\penalty0 189--206, 1984.

\bibitem[Jouppi et~al.(2017)Jouppi, Young, Patil, Patterson, Agrawal, Bajwa,
  Bates, Bhatia, Boden, Borchers, et~al.]{jouppi2017datacenter}
N.~P. Jouppi, C.~Young, N.~Patil, D.~Patterson, G.~Agrawal, R.~Bajwa, S.~Bates,
  S.~Bhatia, N.~Boden, A.~Borchers, et~al.
\newblock In-datacenter performance analysis of a tensor processing unit.
\newblock In \emph{Proceedings of the 44th annual international symposium on
  computer architecture}, pages 1--12, 2017.

\bibitem[Kaz~Sato(2021)]{VertexAIMatchingEngine}
T.~C. Kaz~Sato.
\newblock Vertex ai matching engine.
\newblock \emph{Microsoft AI Blog}, 2021.
\newblock URL
  \url{https://cloud.google.com/blog/topics/developers-practitioners/find-anything-blazingly-fast-googles-vector-search-technology}.

\bibitem[Krizhevsky et~al.(2012)Krizhevsky, Sutskever, and
  Hinton]{krizhevsky2012imagenet}
A.~Krizhevsky, I.~Sutskever, and G.~E. Hinton.
\newblock Imagenet classification with deep convolutional neural networks.
\newblock \emph{Advances in neural information processing systems}, 25, 2012.

\bibitem[Kulis et~al.(2009)Kulis, Jain, and Grauman]{kulis2009fast}
B.~Kulis, P.~Jain, and K.~Grauman.
\newblock Fast similarity search for learned metrics.
\newblock \emph{IEEE Transactions on Pattern Analysis and Machine
  Intelligence}, 31\penalty0 (12):\penalty0 2143--2157, 2009.

\bibitem[Kusupati et~al.(2018)Kusupati, Singh, Bhatia, Kumar, Jain, and
  Varma]{kusupati2018fastgrnn}
A.~Kusupati, M.~Singh, K.~Bhatia, A.~Kumar, P.~Jain, and M.~Varma.
\newblock Fastgrnn: A fast, accurate, stable and tiny kilobyte sized gated
  recurrent neural network.
\newblock \emph{Advances in Neural Information Processing Systems}, 31, 2018.

\bibitem[Kusupati et~al.(2020)Kusupati, Ramanujan, Somani, Wortsman, Jain,
  Kakade, and Farhadi]{kusupati2020soft}
A.~Kusupati, V.~Ramanujan, R.~Somani, M.~Wortsman, P.~Jain, S.~Kakade, and
  A.~Farhadi.
\newblock Soft threshold weight reparameterization for learnable sparsity.
\newblock In \emph{International Conference on Machine Learning}, pages
  5544--5555. PMLR, 2020.

\bibitem[Kusupati et~al.(2021)Kusupati, Wallingford, Ramanujan, Somani, Park,
  Pillutla, Jain, Kakade, and Farhadi]{kusupati2021llc}
A.~Kusupati, M.~Wallingford, V.~Ramanujan, R.~Somani, J.~S. Park, K.~Pillutla,
  P.~Jain, S.~Kakade, and A.~Farhadi.
\newblock Llc: Accurate, multi-purpose learnt low-dimensional binary codes.
\newblock \emph{Advances in Neural Information Processing Systems}, 34, 2021.

\bibitem[Leclerc et~al.(2022)Leclerc, Ilyas, Engstrom, Park, Salman, and
  Madry]{ffcv}
G.~Leclerc, A.~Ilyas, L.~Engstrom, S.~M. Park, H.~Salman, and A.~Madry.
\newblock ffcv.
\newblock \url{https://github.com/libffcv/ffcv/}, 2022.
\newblock commit 607d117.

\bibitem[LeCun et~al.(2015)LeCun, Bengio, and Hinton]{lecun2015deep}
Y.~LeCun, Y.~Bengio, and G.~Hinton.
\newblock Deep learning.
\newblock \emph{nature}, 521\penalty0 (7553):\penalty0 436--444, 2015.

\bibitem[Lee et~al.(2016)Lee, Purushwalkam Shiva~Prakash, Cogswell, Ranjan,
  Crandall, and Batra]{lee2016stochastic}
S.~Lee, S.~Purushwalkam Shiva~Prakash, M.~Cogswell, V.~Ranjan, D.~Crandall, and
  D.~Batra.
\newblock Stochastic multiple choice learning for training diverse deep
  ensembles.
\newblock \emph{Advances in Neural Information Processing Systems}, 29, 2016.

\bibitem[Li et~al.(2018)Li, Farkhoor, Liu, and Yosinski]{li2018measuring}
C.~Li, H.~Farkhoor, R.~Liu, and J.~Yosinski.
\newblock Measuring the intrinsic dimension of objective landscapes.
\newblock \emph{arXiv preprint arXiv:1804.08838}, 2018.

\bibitem[Linde et~al.(1980)Linde, Buzo, and Gray]{linde1980algorithm}
Y.~Linde, A.~Buzo, and R.~Gray.
\newblock An algorithm for vector quantizer design.
\newblock \emph{IEEE Transactions on communications}, 28\penalty0 (1):\penalty0
  84--95, 1980.

\bibitem[Loshchilov and Hutter(2017)]{loshchilov2017decoupled}
I.~Loshchilov and F.~Hutter.
\newblock Decoupled weight decay regularization.
\newblock \emph{arXiv preprint arXiv:1711.05101}, 2017.

\bibitem[Malkov and Yashunin(2018)]{malkov2018efficient}
Y.~A. Malkov and D.~A. Yashunin.
\newblock Efficient and robust approximate nearest neighbor search using
  hierarchical navigable small world graphs.
\newblock \emph{IEEE transactions on pattern analysis and machine
  intelligence}, 42\penalty0 (4):\penalty0 824--836, 2018.

\bibitem[Masci et~al.(2011)Masci, Meier, Cire{\c{s}}an, and
  Schmidhuber]{masci2011stacked}
J.~Masci, U.~Meier, D.~Cire{\c{s}}an, and J.~Schmidhuber.
\newblock Stacked convolutional auto-encoders for hierarchical feature
  extraction.
\newblock In \emph{International conference on artificial neural networks},
  pages 52--59. Springer, 2011.

\bibitem[Mitra et~al.(2002)Mitra, Murthy, and Pal]{mitra2002unsupervised}
P.~Mitra, C.~Murthy, and S.~K. Pal.
\newblock Unsupervised feature selection using feature similarity.
\newblock \emph{IEEE transactions on pattern analysis and machine
  intelligence}, 24\penalty0 (3):\penalty0 301--312, 2002.

\bibitem[Nanda et~al.(2023)Nanda, Speicher, Dickerson, Feizi, Gummadi, and
  Weller]{nanda2023diffused}
V.~Nanda, T.~Speicher, J.~P. Dickerson, S.~Feizi, K.~P. Gummadi, and A.~Weller.
\newblock Diffused redundancy in pre-trained representations.
\newblock \emph{arXiv preprint arXiv:2306.00183}, 2023.

\bibitem[Nayak(2019)]{NayakUnderstanding}
P.~Nayak.
\newblock Understanding searches better than ever before.
\newblock \emph{Google AI Blog}, 2019.
\newblock URL
  \url{https://blog.google/products/search/search-language-understanding-bert/}.

\bibitem[Paszke et~al.(2019)Paszke, Gross, Massa, Lerer, Bradbury, Chanan,
  Killeen, Lin, Gimelshein, Antiga, et~al.]{paszke2019pytorch}
A.~Paszke, S.~Gross, F.~Massa, A.~Lerer, J.~Bradbury, G.~Chanan, T.~Killeen,
  Z.~Lin, N.~Gimelshein, L.~Antiga, et~al.
\newblock Pytorch: An imperative style, high-performance deep learning library.
\newblock \emph{Advances in neural information processing systems}, 32, 2019.

\bibitem[Peters et~al.(2018)Peters, Neumann, Iyyer, Gardner, Clark, Lee, and
  Zettlemoyer]{peters-etal-2018-deep}
M.~E. Peters, M.~Neumann, M.~Iyyer, M.~Gardner, C.~Clark, K.~Lee, and
  L.~Zettlemoyer.
\newblock Deep contextualized word representations.
\newblock In \emph{Proceedings of the 2018 Conference of the North {A}merican
  Chapter of the Association for Computational Linguistics: Human Language
  Technologies, Volume 1 (Long Papers)}, pages 2227--2237, New Orleans,
  Louisiana, June 2018. Association for Computational Linguistics.
\newblock \doi{10.18653/v1/N18-1202}.
\newblock URL \url{https://aclanthology.org/N18-1202}.

\bibitem[Prabhu et~al.(2020)Prabhu, Kusupati, Gupta, and
  Varma]{prabhu2020extreme}
Y.~Prabhu, A.~Kusupati, N.~Gupta, and M.~Varma.
\newblock Extreme regression for dynamic search advertising.
\newblock In \emph{Proceedings of the 13th International Conference on Web
  Search and Data Mining}, pages 456--464, 2020.

\bibitem[Radford et~al.(2018)Radford, Narasimhan, Salimans, and
  Sutskever]{radford2018improving}
A.~Radford, K.~Narasimhan, T.~Salimans, and I.~Sutskever.
\newblock Improving language understanding by generative pre-training.
\newblock \emph{OpenAI Blog}, 2018.
\newblock URL \url{https://openai.com/blog/language-unsupervised/}.

\bibitem[Radford et~al.(2021)Radford, Kim, Hallacy, Ramesh, Goh, Agarwal,
  Sastry, Askell, Mishkin, Clark, et~al.]{radford2021learning}
A.~Radford, J.~W. Kim, C.~Hallacy, A.~Ramesh, G.~Goh, S.~Agarwal, G.~Sastry,
  A.~Askell, P.~Mishkin, J.~Clark, et~al.
\newblock Learning transferable visual models from natural language
  supervision.
\newblock In \emph{International Conference on Machine Learning}, pages
  8748--8763. PMLR, 2021.

\bibitem[Recht et~al.(2019)Recht, Roelofs, Schmidt, and
  Shankar]{recht2019imagenet}
B.~Recht, R.~Roelofs, L.~Schmidt, and V.~Shankar.
\newblock Do imagenet classifiers generalize to imagenet?
\newblock In \emph{International Conference on Machine Learning}, pages
  5389--5400. PMLR, 2019.

\bibitem[Rippel et~al.(2014)Rippel, Gelbart, and Adams]{rippel2014learning}
O.~Rippel, M.~Gelbart, and R.~Adams.
\newblock Learning ordered representations with nested dropout.
\newblock In \emph{International Conference on Machine Learning}, pages
  1746--1754. PMLR, 2014.

\bibitem[Rissanen(1978)]{rissanen1978modeling}
J.~Rissanen.
\newblock Modeling by shortest data description.
\newblock \emph{Automatica}, 14\penalty0 (5):\penalty0 465--471, 1978.

\bibitem[Ruder et~al.(2019)Ruder, Peters, Swayamdipta, and
  Wolf]{ruder2019transfer}
S.~Ruder, M.~E. Peters, S.~Swayamdipta, and T.~Wolf.
\newblock Transfer learning in natural language processing.
\newblock In \emph{Proceedings of the 2019 conference of the North American
  chapter of the association for computational linguistics: Tutorials}, pages
  15--18, 2019.

\bibitem[Russakovsky et~al.(2015)Russakovsky, Deng, Su, Krause, Satheesh, Ma,
  Huang, Karpathy, Khosla, Bernstein, et~al.]{russakovsky2015imagenet}
O.~Russakovsky, J.~Deng, H.~Su, J.~Krause, S.~Satheesh, S.~Ma, Z.~Huang,
  A.~Karpathy, A.~Khosla, M.~Bernstein, et~al.
\newblock Imagenet large scale visual recognition challenge.
\newblock \emph{International journal of computer vision}, 115\penalty0
  (3):\penalty0 211--252, 2015.

\bibitem[Salakhutdinov and Hinton(2007)]{salakhutdinov2007learning}
R.~Salakhutdinov and G.~Hinton.
\newblock Learning a nonlinear embedding by preserving class neighbourhood
  structure.
\newblock In \emph{Artificial Intelligence and Statistics}, pages 412--419.
  PMLR, 2007.

\bibitem[Salakhutdinov and Hinton(2009)]{salakhutdinov2009semantic}
R.~Salakhutdinov and G.~Hinton.
\newblock Semantic hashing.
\newblock \emph{International Journal of Approximate Reasoning}, 50\penalty0
  (7):\penalty0 969--978, 2009.

\bibitem[S{\'a}nchez et~al.(1997)S{\'a}nchez, Pla, and Ferri]{sanchez1997use}
J.~S. S{\'a}nchez, F.~Pla, and F.~J. Ferri.
\newblock On the use of neighbourhood-based non-parametric classifiers.
\newblock \emph{Pattern Recognition Letters}, 18\penalty0 (11-13):\penalty0
  1179--1186, 1997.

\bibitem[Selvaraju et~al.(2017)Selvaraju, Cogswell, Das, Vedantam, Parikh, and
  Batra]{selvaraju2017grad}
R.~R. Selvaraju, M.~Cogswell, A.~Das, R.~Vedantam, D.~Parikh, and D.~Batra.
\newblock Grad-cam: Visual explanations from deep networks via gradient-based
  localization.
\newblock In \emph{Proceedings of the IEEE international conference on computer
  vision}, pages 618--626, 2017.

\bibitem[Shazeer and Stern(2018)]{shazeer2018adafactor}
N.~Shazeer and M.~Stern.
\newblock Adafactor: Adaptive learning rates with sublinear memory cost.
\newblock In \emph{International Conference on Machine Learning}, pages
  4596--4604. PMLR, 2018.

\bibitem[Simonyan and Zisserman(2014)]{simonyan2014very}
K.~Simonyan and A.~Zisserman.
\newblock Very deep convolutional networks for large-scale image recognition.
\newblock \emph{arXiv preprint arXiv:1409.1556}, 2014.

\bibitem[Smith(2017)]{smith2017cyclical}
L.~N. Smith.
\newblock Cyclical learning rates for training neural networks.
\newblock In \emph{2017 IEEE winter conference on applications of computer
  vision (WACV)}, pages 464--472. IEEE, 2017.

\bibitem[Soudry et~al.(2018)Soudry, Hoffer, Nacson, Gunasekar, and
  Srebro]{soudry2018implicit}
D.~Soudry, E.~Hoffer, M.~S. Nacson, S.~Gunasekar, and N.~Srebro.
\newblock The implicit bias of gradient descent on separable data.
\newblock \emph{The Journal of Machine Learning Research}, 19\penalty0
  (1):\penalty0 2822--2878, 2018.

\bibitem[Sun et~al.(2017)Sun, Shrivastava, Singh, and Gupta]{sun2017revisiting}
C.~Sun, A.~Shrivastava, S.~Singh, and A.~Gupta.
\newblock Revisiting unreasonable effectiveness of data in deep learning era.
\newblock In \emph{Proceedings of the IEEE international conference on computer
  vision}, pages 843--852, 2017.

\bibitem[Sutskever et~al.(2013)Sutskever, Martens, Dahl, and
  Hinton]{sutskever2013importance}
I.~Sutskever, J.~Martens, G.~Dahl, and G.~Hinton.
\newblock On the importance of initialization and momentum in deep learning.
\newblock In \emph{International conference on machine learning}, pages
  1139--1147. PMLR, 2013.

\bibitem[Tan and Le(2019)]{tan2019efficientnet}
M.~Tan and Q.~Le.
\newblock Efficientnet: Rethinking model scaling for convolutional neural
  networks.
\newblock In \emph{International conference on machine learning}, pages
  6105--6114. PMLR, 2019.

\bibitem[Van Der~Maaten et~al.(2009)Van Der~Maaten, Postma, Van~den Herik,
  et~al.]{van2009dimensionality}
L.~Van Der~Maaten, E.~Postma, J.~Van~den Herik, et~al.
\newblock Dimensionality reduction: a comparative.
\newblock \emph{J Mach Learn Res}, 10\penalty0 (66-71):\penalty0 13, 2009.

\bibitem[Varma(2019)]{varma2019extreme}
M.~Varma.
\newblock Extreme classification.
\newblock \emph{Communications of the ACM}, 62\penalty0 (11):\penalty0 44--45,
  2019.

\bibitem[Viola and Jones(2001)]{viola2001rapid}
P.~Viola and M.~Jones.
\newblock Rapid object detection using a boosted cascade of simple features.
\newblock In \emph{Proceedings of the 2001 IEEE computer society conference on
  computer vision and pattern recognition. CVPR 2001}, volume~1, pages I--I.
  Ieee, 2001.

\bibitem[Waldburger(2019)]{Waldburger2019Search}
C.~Waldburger.
\newblock As search needs evolve, microsoft makes ai tools for better search
  available to researchers and developers.
\newblock \emph{Microsoft AI Blog}, 2019.
\newblock URL \url{https://blogs.microsoft.com/ai/bing-vector-search/}.

\bibitem[Wallingford et~al.(2020)Wallingford, Kusupati, Alizadeh-Vahid,
  Walsman, Kembhavi, and Farhadi]{wallingford2020overfitting}
M.~Wallingford, A.~Kusupati, K.~Alizadeh-Vahid, A.~Walsman, A.~Kembhavi, and
  A.~Farhadi.
\newblock Are we overfitting to experimental setups in recognition?
\newblock \emph{arXiv preprint arXiv:2007.02519}, 2020.

\bibitem[Wallingford et~al.(2022)Wallingford, Li, Achille, Ravichandran,
  Fowlkes, Bhotika, and Soatto]{wallingford2022task}
M.~Wallingford, H.~Li, A.~Achille, A.~Ravichandran, C.~Fowlkes, R.~Bhotika, and
  S.~Soatto.
\newblock Task adaptive parameter sharing for multi-task learning.
\newblock \emph{arXiv preprint arXiv:2203.16708}, 2022.

\bibitem[Wang et~al.(2019)Wang, Ge, Lipton, and Xing]{wang2019learning}
H.~Wang, S.~Ge, Z.~Lipton, and E.~P. Xing.
\newblock Learning robust global representations by penalizing local predictive
  power.
\newblock In \emph{Advances in Neural Information Processing Systems}, pages
  10506--10518, 2019.

\bibitem[Wang et~al.(2020)Wang, Kondratyuk, Kitani, Movshovitz-Attias, and
  Eban]{wang2020multiple}
X.~Wang, D.~Kondratyuk, K.~M. Kitani, Y.~Movshovitz-Attias, and E.~Eban.
\newblock Multiple networks are more efficient than one: Fast and accurate
  models via ensembles and cascades.
\newblock \emph{arXiv preprint arXiv:2012.01988}, 2020.

\bibitem[Wortsman et~al.(2021)Wortsman, Ilharco, Li, Kim, Hajishirzi, Farhadi,
  Namkoong, and Schmidt]{wortsman2021robust}
M.~Wortsman, G.~Ilharco, M.~Li, J.~W. Kim, H.~Hajishirzi, A.~Farhadi,
  H.~Namkoong, and L.~Schmidt.
\newblock Robust fine-tuning of zero-shot models.
\newblock \emph{arXiv preprint arXiv:2109.01903}, 2021.

\bibitem[Wu et~al.(2018)Wu, Xiong, Yu, and Lin]{wu2018unsupervised}
Z.~Wu, Y.~Xiong, S.~Yu, and D.~Lin.
\newblock Unsupervised feature learning via non-parametric instance-level
  discrimination.
\newblock \emph{arXiv preprint arXiv:1805.01978}, 2018.

\bibitem[Yosinski et~al.(2014)Yosinski, Clune, Bengio, and
  Lipson]{yosinski2014transferable}
J.~Yosinski, J.~Clune, Y.~Bengio, and H.~Lipson.
\newblock How transferable are features in deep neural networks?
\newblock \emph{Advances in neural information processing systems}, 27, 2014.

\bibitem[Yu et~al.(2022)Yu, Zhong, Zhang, Chang, and Dhillon]{yu2022pecos}
H.-F. Yu, K.~Zhong, J.~Zhang, W.-C. Chang, and I.~S. Dhillon.
\newblock Pecos: Prediction for enormous and correlated output spaces.
\newblock \emph{Journal of Machine Learning Research}, 23\penalty0
  (98):\penalty0 1--32, 2022.

\bibitem[Yu et~al.(2018)Yu, Yang, Xu, Yang, and Huang]{yu2018slimmable}
J.~Yu, L.~Yang, N.~Xu, J.~Yang, and T.~Huang.
\newblock Slimmable neural networks.
\newblock \emph{arXiv preprint arXiv:1812.08928}, 2018.

\bibitem[Zellers et~al.(2022)Zellers, Lu, Lu, Yu, Zhao, Salehi, Kusupati,
  Hessel, Farhadi, and Choi]{zellers2022merlot}
R.~Zellers, J.~Lu, X.~Lu, Y.~Yu, Y.~Zhao, M.~Salehi, A.~Kusupati, J.~Hessel,
  A.~Farhadi, and Y.~Choi.
\newblock Merlot reserve: Neural script knowledge through vision and language
  and sound.
\newblock \emph{arXiv preprint arXiv:2201.02639}, 2022.

\bibitem[Zhu et~al.(2015)Zhu, Kiros, Zemel, Salakhutdinov, Urtasun, Torralba,
  and Fidler]{zhu2015aligning}
Y.~Zhu, R.~Kiros, R.~Zemel, R.~Salakhutdinov, R.~Urtasun, A.~Torralba, and
  S.~Fidler.
\newblock Aligning books and movies: Towards story-like visual explanations by
  watching movies and reading books.
\newblock In \emph{Proceedings of the IEEE international conference on computer
  vision}, pages 19--27, 2015.

\end{thebibliography}
\clearpage
\section*{Checklist}


\begin{enumerate}

\item For all authors...
\begin{enumerate}
  \item Do the main claims made in the abstract and introduction accurately reflect the paper's contributions and scope?
    \answerYes{}
  \item Did you describe the limitations of your work?
    \answerYes{See Section~\ref{sec:conc}}
  \item Did you discuss any potential negative societal impacts of your work?
    \answerNA{Our work does not have any additional negative societal impact on top of the existing impact of representation learning. However, a study on the trade-off between representation size and the tendency to encode biases is an interesting future direction along the lines of existing literature~\citep{hooker2019compressed,hooker2020characterising}. A part of this is already presented in Section~\ref{sec:analysis}.}
  \item Have you read the ethics review guidelines and ensured that your paper conforms to them?
    \answerYes{}
\end{enumerate}

\item If you are including theoretical results...
\begin{enumerate}
  \item Did you state the full set of assumptions of all theoretical results?
    \answerNA{}
        \item Did you include complete proofs of all theoretical results?
    \answerNA{}
\end{enumerate}

\item If you ran experiments...
\begin{enumerate}
  \item Did you include the code, data, and instructions needed to reproduce the main experimental results (either in the supplemental material or as a URL)?
    \answerYes{See supplemental material and Appendix~\ref{sec:code}. All the code and public models will be open sourced.}
  \item Did you specify all the training details (e.g., data splits, hyperparameters, how they were chosen)?
    \answerYes{See Section~\ref{sec:apps} and Appendix~\ref{sec:appendix-mrl_model_training}.}
        \item Did you report error bars (e.g., with respect to the random seed after running experiments multiple times)?
    \answerNo{We benchmarked on large-scale datasets like ImageNet-1K, JFT-300M and ALIGN data with models like ResNet and ViT making it extremely expensive to run things multiple times.}
        \item Did you include the total amount of compute and the type of resources used (e.g., type of GPUs, internal cluster, or cloud provider)?
    \answerYes{See Appendix~\ref{sec:appendix-mrl_model_training} and Appendix~\ref{sec:appendix_real-world-perf}.}
\end{enumerate}

\item If you are using existing assets (e.g., code, data, models) or curating/releasing new assets...
\begin{enumerate}
  \item If your work uses existing assets, did you cite the creators?
    \answerYes{}
  \item Did you mention the license of the assets?
    \answerNo{All the non-proprietary datasets and code used are public under MIT, BSD or CC licenses.}
  \item Did you include any new assets either in the supplemental material or as a URL?
    \answerYes{We created a new subset of ImageNet-21K for downstream evaluation of retrieval performance at scale. See Section~\ref{sec:retrieval} and Appendix~\ref{sec:datasets}}
  \item Did you discuss whether and how consent was obtained from people whose data you're using/curating?
    \answerNA{}
  \item Did you discuss whether the data you are using/curating contains personally identifiable information or offensive content?
    \answerNA{}
\end{enumerate}

\item If you used crowdsourcing or conducted research with human subjects...
\begin{enumerate}
  \item Did you include the full text of instructions given to participants and screenshots, if applicable?
    \answerNA{}
  \item Did you describe any potential participant risks, with links to Institutional Review Board (IRB) approvals, if applicable?
    \answerNA{}
  \item Did you include the estimated hourly wage paid to participants and the total amount spent on participant compensation?
    \answerNA{}
\end{enumerate}

\end{enumerate}
\clearpage
\appendix
\tableofcontents

\section{Code for \alg~\mdoll~(\mrl)}
\label{sec:code}
We use Alg~\ref{code:NCE-Loss} and~\ref{code:MRL} provided below to train supervised ResNet50--\mrl models on \InIk. We provide this code as a template to extend \mrl to any domain.

\definecolor{codeblue}{rgb}{0.25,0.5,0.5}
\definecolor{codeblue2}{rgb}{0,0,1}
\lstset{
  backgroundcolor=\color{white},
  basicstyle=\fontsize{10pt}{10pt}\ttfamily\selectfont,
  columns=fullflexible,
  breaklines=true,
  captionpos=b,
  commentstyle=\fontsize{8pt}{8pt}\color{codeblue},
  keywordstyle=\fontsize{8pt}{8pt}\color{codeblue2},
}

\begin{algorithm}[!h]
\caption{\large Pytorch code for \ma Cross-Entropy Loss}
\begin{lstlisting}[language=Python]

class Matryoshka_CE_Loss(nn.Module):
    def __init__(self, relative_importance, **kwargs):
		super(Matryoshka_CE_Loss, self).__init__()
		self.criterion = nn.CrossEntropyLoss(**kwargs)
		self.relative_importance = relative_importance # usually set to all ones

  def forward(self, output, target):
		loss=0
		for i in range(len(output)):
		  loss+= self.relative_importance[i] * self.criterion(output[i], target)
		return loss
\end{lstlisting}
\label{code:NCE-Loss}
\end{algorithm}

\begin{algorithm}[!h]
\caption{\large Pytorch code for \mrl Linear Layer}
\begin{lstlisting}[language=Python]

class MRL_Linear_Layer(nn.Module):
	def __init__(self, nesting_list: List, num_classes=1000, efficient=False, **kwargs):
	    super(MRL_Linear_Layer, self).__init__()
	    self.nesting_list=nesting_list # set of m in M (Eq. 1)
	    self.num_classes=num_classes 
	    self.is_efficient=efficient # flag for MRL-E
	    
            if not is_efficient:
                for i, num_feat in enumerate(self.nesting_list):
                    setattr(self, f"nesting_classifier_{i}", nn.Linear(num_feat, self.num_classes, **kwargs))
            else:
                setattr(self, "nesting_classifier_0", nn.Linear(self.nesting_list[-1], self.num_classes, **kwargs)) # Instantiating one nn.Linear layer for MRL-E

        def forward(self, x):
        	nesting_logits = ()
        	for i, num_feat in enumerate(self.nesting_list):
        		if(self.is_efficient):
        			efficient_logit = torch.matmul(x[:, :num_feat], (self.nesting_classifier_0.weight[:, :num_feat]).t())
        		else:
        			nesting_logits.append(getattr(self, f"nesting_classifier_{i}")(x[:, :num_feat]))
        
        	if(self.is_efficient):
        		nesting_logits.append(efficient_logit)
        
        	return nesting_logits
\end{lstlisting}
\label{code:MRL}
\end{algorithm}
\newpage

\section{Datasets}
\label{sec:datasets}
\textbf{\InIk}~\citep{russakovsky2015imagenet} contains 1,281,167 labeled train images, and 50,000 labelled validation images across 1,000 classes. The images were transformed with standard procedures detailed by FFCV~\citep{ffcv}.

\textbf{\InIVk} dataset was constructed by selecting 4,202 classes, non-overlapping with ImageNet-1K, from ImageNet-21K~\citep{deng2009imagenet} with 1,050 or more examples. The train set contains 1,000 examples and the query/validation set contains 50 examples per class totalling to $\sim$4.2M and $\sim$200K respectively. We will release the list of images curated together to construct ImageNet-4K.

\textbf{JFT-300M}~\citep{sun2017revisiting} is a large-scale multi-label dataset with 300M images labelled across 18,291 categories.

\textbf{ALIGN}~\citep{jia2021scaling} utilizes a large scale noisy image-text dataset containing 1.8B image-text pairs.

\paragraph{ImageNet Robustness Datasets}
We experimented on the following datasets to examine the robustness of \mrl models:

\textbf{\INVTwo}~\citep{recht2019imagenet} is a collection of 10K images sampled a decade after the original construction of ImageNet~\citep{deng2009imagenet}. \INVTwo contains 10 examples each from the 1,000 classes of \InIk.

\textbf{ImageNet-A}~\citep{hendrycks2021natural} contains 7.5K real-world adversarially filtered images from 200 \InIk~classes.

\textbf{ImageNet-R}~\citep{hendrycks2021many} contains 30K artistic image renditions for 200 of the original \InIk~classes.

\textbf{ImageNet-Sketch}~\citep{wang2019learning} contains 50K sketches, evenly distributed over all 1,000 \InIk~classes.

\textbf{ObjectNet}~\citep{barbu2019objectnet} contains 50K images across 313 object classes, each containing $\sim$160 images each.

\section{\alg Model Training}
\label{sec:appendix-mrl_model_training}
We trained all ResNet50--\mrl models using the efficient dataloaders of FFCV~\citep{ffcv}. We utilized the \url{rn50_40_epochs.yaml} configuration file of FFCV to train all \nrl models defined below:

\begin{itemize}[leftmargin=*]\vspace{-1mm}
    \item \textbf{\MH}: ResNet50 model with the fc layer replaced by \lstinline{MRL_Linear_Layer(efficient=False)}
    \item \textbf{\SH}: ResNet50 model with the fc layer replaced by \lstinline{MRL_Linear_Layer(efficient=True)}
    \item \FF--k: ResNet50 model with the fc layer replaced by \lstinline{torch.nn.Linear(k, num_classes)}, where k~$\in [8, 16, 32, 64, 128, 256, 512, 1024, 2048]$. We will henceforth refer to these models as simply \FF, with the k value denoting representation size.
\end{itemize}

We trained all ResNet50 models with a learning rate of $0.475$ with a cyclic learning rate schedule~\citep{smith2017cyclical}. This was after appropriate scaling (0.25$\times$) of the learning rate specified in the configuration file to accommodate for 2xA100 NVIDIA GPUs available for training, compared to the 8xA100 GPUs utilized in the FFCV benchmarks. We trained with a batch size of 256 per GPU, momentum~\citep{sutskever2013importance} of 0.9, and an SGD optimizer with a weight decay of 1e-4.

Our code (Appendix~\ref{sec:code}) makes minimal modifications to the training pipeline provided by FFCV to learn \mrs.

We trained ViT-B/16 models for JFT-300M on a 8x8 cloud TPU pod~\citep{jouppi2017datacenter} using Tensorflow~\citep{tensorflow2015-whitepaper} with a batchsize of 128 and trained for 300K steps. Similarly, ALIGN models were trained using Tensorflow on 8x8 cloud TPU pod for 1M steps with a batchsize of 64 per TPU. Both these models were trained with adafactor optimizer~\citep{shazeer2018adafactor}  with a linear learning rate decay starting at 1e-3.

Lastly, we trained a BERT-Base model on English Wikipedia and BookCorpus. We trained our models in Tensorflow using a 4x4 cloud TPU pod  with a total batchsize of 1024. We used AdamW~\citep{loshchilov2017decoupled} optimizer with a linear learning rate decay starting at 1e-4 and trained for 450K steps.

In each configuration/case, if the final representation was normalized in the FF implementation, \mrl models adopted the same for each nested dimension for a fair comparison.

\section{Classification Results}
\label{sec:appendix_classification_results}

\begin{table}[ht]
\centering
 \caption{Top-1 classification accuracy (\%) for ResNet50 \mrl and baseline models on \InIk.}
 \vspace{1mm}
 \resizebox{0.7\columnwidth}{!}{
\begin{tabular}{@{}c|cccccc@{}}
\toprule
\Dims & Rand. LP & SVD            & \FF            & Slim. Net & \MH            & \SH   \\ \midrule
8     & ~4.56   & ~2.34          & 65.29          & ~0.42     & \textbf{66.63} & 56.66 \\
16    & 11.29   & ~7.17          & 72.85          & ~0.96     & \textbf{73.53} & 71.94 \\
32    & 27.21   & 20.46          & 74.60          & ~2.27     & \textbf{75.03} & 74.48 \\
64    & 49.47   & 48.10          & 75.27          & ~5.59     & \textbf{75.82} & 75.35 \\
128   & 65.70   & 67.24          & 75.29          & 14.15     & \textbf{76.30} & 75.80 \\
256   & 72.43   & 74.59          & 75.71          & 38.42     & \textbf{76.47} & 76.22 \\
512   & 74.94   & \textbf{76.78} & 76.18          & 69.80     & 76.65          & 76.36 \\
1024  & 76.10   & \textbf{76.87} & 76.63          & 74.61     & 76.76          & 76.48 \\
2048  & 76.87   & --             & \textbf{76.87} & 76.26     & 76.80          & 76.51 \\ \bottomrule
\end{tabular}
\label{tab:r50_accuracy_main}
}
\end{table}
We show the top-1 classification accuracy of ResNet50--\mrl models on \InIk in Table~\ref{tab:r50_accuracy_main} and Figure~\ref{fig:r50-acc}. We compare the performance of \nrl models (\MH, \SH) to several baselines: 

\begin{itemize}[leftmargin=*]\vspace{-1mm}
    \item \FF: We utilize the \FF-k models described in Appendix~\ref{sec:appendix-mrl_model_training} for $k\in\{8, ... 2048\}$.
    \item \textbf{SVD}: We performed a low rank approximation of the 1000-way classification layer of \FF-2048, with rank = 1000.
    \item \textbf{Rand. LP}: We compared against a linear classifier fit on randomly selected features~\citep{he2020momentum}.
    \item \textbf{Slim. Net}: We take pretrained slimmable neural networks~\citep{yu2018slimmable} which are trained with a flexible width backbone (25\%, 50\%, 75\% and full width). For each representation size, we consider the first $k$ dimensions for classification. Note that training of slimmable neural networks becomes unstable when trained below 25\% width due to the hardness in optimization and low complexity of the model.
\end{itemize}
At lower dimensions ( $d \leq 128$), \nrl outperforms all baselines significantly, which indicates that pretrained models lack the multifidelity of \mrs and are incapable of fitting an accurate linear classifier at low representation sizes.

We compared the performance of \nrl models at various representation sizes via 1-nearest neighbors (1-NN) image classification accuracy on \InIk in Table~\ref{tab:r50-knn-acc} and Figure~\ref{fig:r50-knn-acc}. We provide detailed information regarding the k-NN search pipeline in Appendix~\ref{sec:appendix-retrieval}. We compared against a baseline of attempting to enforce nesting to a \FF-2048 model by 1) Random Feature Selection (Rand. FS): considering the first $m$ dimensions of \FF-2048 for NN lookup, and 2) \FF+SVD: performing SVD on the \FF-2048 representations at the specified representation size, 3) \FF+JL: performing random projection according to the Johnson-Lindenstrauss lemma~\citep{johnson1984extensions} on the \FF-2048 representations at the specified representation size. We also compared against the 1-NN accuracy of slimmable neural nets~\citep{yu2018slimmable} as an additional baseline. We observed these baseline models to perform very poorly at lower dimensions, as they were not explicitly trained to learn \mrs.
\begin{table}[h]
\centering
 \caption{1-NN accuracy (\%) on \InIk~for various ResNet50 models.}
 \vspace{1mm}
 \resizebox{0.7\columnwidth}{!}{
\begin{tabular}{@{}c|ccccccc@{}}
\toprule
\Dims & Rand. FS   & SVD   & JL   & \FF & Slimmable  & \MH & \SH \\ \midrule
8                    & ~2.36 & 19.14 & 0.11 & 58.93              & ~1.00 & 62.19              & 57.45              \\
16                   & 12.06      & 46.02 & 0.09 & 66.77              & ~5.12 & 67.91              & 67.05              \\
32                   & 32.91      & 60.78 & 0.06 & 68.84              & 16.95      & 69.46              & 68.6               \\
64                   & 49.91      & 67.04 & 0.05 & 69.41              & 35.60      & 70.17              & 69.61              \\
128                  & 60.91      & 69.63 & 0.06 & 69.35              & 51.16      & 70.52              & 70.12              \\
256                  & 65.75      & 70.67 & 0.04 & 69.72              & 60.61      & 70.62              & 70.36              \\
512                  & 68.77      & 71.06 & 0.03 & 70.18              & 65.82      & 70.82              & 70.74              \\
1024                 & 70.41      & 71.22 & -    & 70.34              & 67.19      & 70.89              & 71.07              \\
2048                 & 71.19      & 71.21 & -    & 71.19              & 66.10      & 70.97              & 71.21              \\ \bottomrule
\end{tabular}
\label{tab:r50-knn-acc}
}
\end{table}

\subsection{Adaptive Classification (\nrl--AC)}
\label{sec:appendix_adaptive_classification}
\begin{table}[ht!]
\centering
 \caption{Threshold-based adaptive classification performance of ResNet50 \MH~on a 40K sized held-out subset of the \InIk~validation set. Results are averaged over 30 random held-out subsets.}
 \resizebox{0.3\columnwidth}{!}{%
\begin{tabular}{@{}cc@{}}
\toprule
Expected \Dims    & Accuracy      \\ \midrule
~13.43 $\pm$ ~0.81 & 73.79 $\pm$ 0.10 \\
~18.32 $\pm$ ~1.36 & 75.25 $\pm$ 0.11 \\
~25.87 $\pm$ ~2.41 & 76.05 $\pm$ 0.15 \\
~36.26 $\pm$ ~4.78 & 76.28 $\pm$ 0.16 \\
~48.00 $\pm$ ~8.24 & 76.43 $\pm$ 0.18 \\
~64.39 $\pm$ ~12.55 & 76.53 $\pm$ 0.19 \\
~90.22 $\pm$ 20.88 & 76.55 $\pm$ 0.20 \\
~118.85 $\pm$ 33.37 & 76.56 $\pm$ 0.20 \\ \bottomrule
\end{tabular}
}
\label{tab:r50_cascades}
\end{table}
In an attempt to use the smallest representation that works well for classification for every image in the \InIk~validation set, we learned a policy to increase the representation size from $m_i$ to $m_{i+1}$ using a 10K sized subset of the \InIk~validation set. This policy is based on whether the prediction confidence $p_i$ using representation size $m_i$ exceeds a learned threshold $t_{i}^{\ast}$. If $p_i \geq t_{i}^{\ast}$, we used predictions from representation size $m_i$ otherwise, we increased to representation size $m_{i+1}$. To learn the optimal threshold $t_{i}^{\ast}$, we performed a grid search between 0 and 1 (100 samples). For each threshold $t_k$, we computed the classification accuracy over our 10K image subset. We set $t_{i}^{\ast}$ equal to the smallest threshold $t_k$ that gave the best accuracy. We use this procedure to obtain thresholds for successive models, i.e., $\{t_{j}^{\ast} \mid j \in \{8, 16, 32, 64, \ldots, 2048\}\}$. To improve reliability of threshold based greedy policy, we use test time augmentation which has been used successfully in the past~\citep{simonyan2014very}. 

For inference, we used the remaining held-out 40K samples from the \InIk validation set. We began with smallest sized representation ($m = 8$) and compared the computed prediction confidence $p_8$ to learned optimal threshold $t_8^{\ast}$. If $p_8 \leq t_8^{\ast}$, then we increased $m = 16$, and repeated this procedure until $m = d = 2048$. To compute the expected dimensions, we performed early stopping at $m = \{16, 32, 64, \ldots 2048\}$ and computed the expectation using the distribution of representation sizes. As shown in Table~\ref{tab:r50_cascades} and Figure~\ref{fig:r50-mrl-cascade-acc}, we observed that in expectation, we only needed a $\sim37$ sized representation to achieve $76.3\%$ classification accuracy on \InIk, which was roughly $14\times$ smaller than the \FF--512 baseline. Even if we computed the expectation as a weighted average over the cumulative sum of representation sizes $\{8, 24, 56, \ldots\}$, due to the nature of multiple linear heads for \mrl, we ended up with an expected size of $62$ that still provided a roughly $8.2\times$ efficient representation than the \FF--512 baseline. However, \mrle alleviates this extra compute with a minimal drop in accuracy.

\subsection{JFT, ALIGN and BERT}
\label{sec:appendix_jft_align_bert}
We examine the k-NN classification accuracy of learned \mrs via ALIGN--\nrl and JFT-ViT--\nrl in Table~\ref{tab:r50-align_jft_knn}. For ALIGN~\citep{jia2021scaling}, we observed that learning \mrs via ALIGN--\MH improved classification accuracy at nearly all dimensions when compared to ALIGN. We observed a similar trend when training ViT-B/16~\citep{dosovitskiy2020image} for JFT-300M~\citep{sun2017revisiting} classification, where learning \mrs via \MH~ and \SH~ on top of JFT-ViT improved classification accuracy for nearly all dimensions, and significantly for lower ones. This demonstrates that training to learn \mrs is feasible and extendable even for extremely large scale datasets. We also demonstrate that \mrs are learned at interpolated dimensions for both ALIGN and JFT-ViT, as shown in Table~\ref{tab:align_jft_interpolated}, despite not being trained explicitly at these dimensions. Lastly, Table~\ref{tab:align-cosine-sim} shows that \mrl training leads to a increase in the cosine similarity span between positive and random image-text pairs.

\begin{table}[ht]
\centering
\caption{ViT-B/16 and ViT-B/16-\nrl top-1 and top-5 k-NN accuracy (\%) for ALIGN and JFT. Top-1 entries where \SH~and \MH~outperform baselines are bolded for both ALIGN and JFT-ViT.}
\vspace{1mm}
 \resizebox{0.9\columnwidth}{!}{%
\begin{tabular}{@{}c|cccc|cccccc@{}}
\toprule
\multirow{2}{*}{\begin{tabular}[c]{@{}c@{}}\Dims\end{tabular}} & \multicolumn{2}{c}{ALIGN} & \multicolumn{2}{c|}{ALIGN-\MH} & \multicolumn{2}{c}{JFT-ViT} & \multicolumn{2}{c}{JFT-ViT-\MH} & \multicolumn{2}{c}{JFT-ViT-\SH} \\ \cmidrule(l){2-11} 
                                                                             & Top-1       & Top-5       & Top-1              & Top-5    & Top-1        & Top-5        & Top-1               & Top-5     & Top-1               & Top-5     \\ \midrule
12                                                                           & 11.90       & 28.05       & \textbf{43.57}     & 67.36    & 27.07        & 48.57        & \textbf{53.61}      & 75.30     & \textbf{51.54}      & 73.94     \\
24                                                                           & 33.35       & 55.58       & \textbf{56.44}     & 78.19    & 48.64        & 70.20        & \textbf{62.80}      & 81.51     & \textbf{62.40}      & 81.36     \\
48                                                                           & 51.32       & 73.15       & \textbf{62.33}     & 82.30    & 63.58        & 81.80        & \textbf{67.24}      & 84.37     & \textbf{66.89}      & 83.80     \\
96                                                                           & 61.82       & 81.97       & \textbf{65.72}     & 84.61    & 68.56        & 85.13        & \textbf{69.74}      & 85.86     & \textbf{68.80}      & 85.13     \\
192                                                                          & 66.71       & 85.27       & \textbf{67.00}     & 85.36    & 71.32        & 86.21        & \textbf{71.34}      & 86.62     & \textbf{70.41}      & 86.01     \\
384                                                                          & 67.65       & 85.70       & \textbf{67.70}     & 85.73    & 71.67        & 86.98        & \textbf{71.73}      & 87.08     & 71.18               & 86.46     \\
768                                                                          & 68.00       & 86.10       & 67.85              & 85.85    & 72.10        & 87.20        & 71.85               & 86.92     & 71.31               & 86.62     \\ \bottomrule
\end{tabular}
}
\label{tab:r50-align_jft_knn}
\end{table}
\begin{table}[ht]
\centering
 \caption{Examining top-1 and top-5 k-NN accuracy (\%) at interpolated hidden dimensions for ALIGN and JFT. This indicates that \mrl is able to scale classification accuracy as hidden dimensions increase even at dimensions that were not explicitly considered during training.}
 \vspace{1mm}
 \resizebox{0.5\columnwidth}{!}{%
\begin{tabular}{@{}c|cc|cc@{}}
\toprule
\multirow{2}{*}{\begin{tabular}[c]{@{}c@{}}Interpolated \\ \Dims\end{tabular}} & \multicolumn{2}{c}{ALIGN-\MH} & \multicolumn{2}{c}{JFT-ViT-\MH} \\ \cmidrule(l){2-5} 
                                                                                     & Top-1         & Top-5         & Top-1          & Top-5          \\ \midrule
16                                                                                   & 49.06         & 72.26         & 58.35          & 78.55          \\
32                                                                                   & 58.64         & 79.96         & 64.98          & 82.89          \\
64                                                                                   & 63.90         & 83.39         & 68.19          & 84.85          \\
128                                                                                  & 66.63         & 85.00         & 70.35          & 86.24          \\
256                                                                                  & 67.10         & 85.30         & 71.57          & 86.77          \\
512                                                                                  & 67.64         & 85.72         & 71.55          & 86.67          \\ \bottomrule
\end{tabular}
\label{tab:align_jft_interpolated}
}
\end{table}
\begin{table}[ht]
\centering
 \caption{Cosine similarity between embeddings}
 \resizebox{0.5\columnwidth}{!}{%
\begin{tabular}{@{}lcc@{}}
\toprule
Avg. Cosine Similarity & ALIGN    & ALIGN-\nrl \\ \midrule
Positive Text to Image    & 0.27     & 0.49       \\
Random Text to Image & 8e-3 & -4e-03  \\
Random Image to Image & 0.10     & 0.08       \\
Random Text to Text & 0.22     & 0.07       \\ \bottomrule
\end{tabular}
\label{tab:align-cosine-sim}
}
\end{table}

We also evaluated the capability of \mrs to extend to other natural language processing via masked language modeling (MLM) with BERT~\citep{devlin2018bert}, whose results are tabulated in Table~\ref{tab:bert_mlm}. Without any hyper-parameter tuning, we observed \mrs to be within $0.5\%$ of \FF~representations for BERT MLM validation accuracy. This is a promising initial result that could help with large-scale adaptive document retrieval using BERT--\mrl.
\begin{table}[h]
\centering
\caption{Masked Language Modelling (MLM) accuracy(\%) of \FF\space and \nrl models on the validation set.}
\resizebox{0.4\columnwidth}{!}{
\begin{tabular}{@{}c|cc@{}}
\toprule
\Dims & BERT-\FF & BERT-\nrl \\ \midrule
12    & 60.12   & 59.92    \\
24    & 62.49   & 62.05    \\
48    & 63.85   & 63.40    \\
96    & 64.32   & 64.15    \\
192   & 64.70   & 64.58    \\
384   & 65.03   & 64.81    \\
768   & 65.54   & 65.00    \\ \bottomrule
\end{tabular}
\label{tab:bert_mlm}
}
\end{table}

\section{Image Retrieval}
\label{sec:appendix-retrieval}

We evaluated the strength of \mrs via image retrieval on \InIk~(the training distribution), as well as on out-of-domain datasets \INVTwo~and \InIVk~for all \nrl ResNet50 models. We generated the database and query sets, containing $N$ and $Q$ samples respectively, with a standard PyTorch~\citep{paszke2019pytorch} forward pass on each dataset. We specify the representation size at which we retrieve a shortlist of k-nearest neighbors (k-NN) by $D_s$. The database is a thus a [$N$, \retdim] array, the query set is a [$Q$, \retdim] array, and the neighbors set is a [$Q$, k] array. For metrics, we utilized corrected mean average precision (mAP@k)~\citep{kusupati2021llc} and precision (P@k):
$P@k = \dfrac{correct\_pred}{k}$
where $correct\_pred$ is the average number of retrieved NN with the correct label over the entire query set using a shortlist of length $k$.

We performed retrieval with FAISS~\citep{johnson2019billion}, a library for efficient similarity search. To obtain a shortlist of k-NN, we built an index to search the database. We performed an exhaustive NN search with the L2 distance metric with \lstinline{faiss.IndexFlatL2}, as well as an approximate NN search (ANNS) via HNSW~\citep{johnson2019billion} with \lstinline{faiss.IndexHNSWFlat}. We used HNSW with $M = 32$ unless otherwise mentioned, and henceforth referred to as HNSW32. The exact search index was moved to the GPU for fast k-NN search computation, whereas the HNSW index was kept on the CPU as it currently lacks GPU support. We show the wall clock times for building the index as well as the index size in Table~\ref{tab:ret_buildtime_indexsize}. We observed exact search to have a smaller index size which was faster to build when compared to HNSW, which trades off a larger index footprint for fast NN search (discussed in more detail in Appendix~\ref{sec:ablation}). The database and query vectors are normalized with \lstinline{faiss.normalize_L2} before building the index and performing search.

\begin{table}[h]
\centering
 \caption{Retrieve a shortlist of 200-NN with \retdim\space sized representations on \InIk\space via exact search with L2 distance metric. Top-1 and mAP@10 entries (\%) where \SH~and \MH~outperform \FF~ at their respective representation sizes are bolded.}
 \vspace{1mm}
 \resizebox{\columnwidth}{!}{%
\begin{tabular}{@{}c|c|c|ccc|cccc|cccc@{}}
\toprule
Model                                                                                       & \retdim & MFlops & Top-1          & Top-5 & Top-10 & mAP@10         & mAP@25 & mAP@50 & mAP@100 & P@10  & P@25  & P@50  & P@100 \\ \midrule
\multirow{9}{*}{\FF}                                                         & 8                      & 10     & 58.93          & 75.76 & 80.25  & 53.42          & 52.29  & 51.84  & 51.57   & 59.32 & 59.28 & 59.25 & 59.21 \\
                                                                                            & 16                     & 20     & 66.77          & 80.88 & 84.40  & 61.63          & 60.51  & 59.98  & 59.62   & 66.76 & 66.58 & 66.43 & 66.27 \\
                                                                                            & 32                     & 41     & 68.84          & 82.58 & 86.14  & 63.35          & 62.08  & 61.36  & 60.76   & 68.43 & 68.13 & 67.83 & 67.48 \\
                                                                                            & 64                     & 82     & 69.41          & 83.56 & 87.33  & 63.26          & 61.64  & 60.63  & 59.67   & 68.49 & 67.91 & 67.38 & 66.74 \\
                                                                                            & 128                    & 164    & 69.35          & 84.23 & 88.24  & 62.30          & 60.16  & 58.73  & 57.29   & 67.84 & 66.83 & 65.96 & 64.92 \\
                                                                                            & 256                    & 328    & 69.72          & 84.71 & 88.54  & 61.47          & 58.85  & 57.02  & 55.13   & 67.19 & 65.82 & 64.64 & 63.24 \\
                                                                                            & 512                    & 656    & 70.18          & 85.04 & 88.91  & 61.37          & 58.41  & 56.26  & 53.98   & 67.12 & 65.49 & 64.07 & 62.35 \\
                                                                                            & 1024                   & 1312   & 70.34          & 85.38 & 89.19  & 61.13          & 57.87  & 55.47  & 52.90   & 66.93 & 65.08 & 63.43 & 61.45 \\
                                                                                            & 2048                   & 2624   & 71.19          & 85.66 & 89.17  & 62.90          & 60.06  & 57.99  & 55.76   & 68.46 & 66.9  & 65.52 & 63.83 \\ \midrule\midrule
\multirow{9}{*}{\SH}                                                         & 8                      & 10     & 57.39          & 74.18 & 79.16  & 51.80          & 50.41  & 49.60  & 48.86   & 57.50 & 57.16 & 56.81 & 56.36 \\
                                                                                            & 16                     & 20     & \textbf{67.08} & 81.38 & 85.15  & 61.60          & 60.36  & 59.66  & 59.04   & 66.79 & 66.53 & 66.24 & 65.87 \\
                                                                                            & 32                     & 41     & 68.62          & 82.92 & 86.44  & 63.34          & 61.97  & 61.14  & 60.39   & 68.49 & 68.06 & 67.65 & 67.17 \\
                                                                                            & 64                     & 82     & \textbf{69.56} & 83.49 & 86.85  & \textbf{63.84} & 62.33  & 61.43  & 60.57   & 68.93 & 68.4  & 67.96 & 67.38 \\
                                                                                            & 128                    & 164    & \textbf{70.13} & 83.63 & 87.07  & \textbf{64.15} & 62.58  & 61.61  & 60.70   & 69.19 & 68.62 & 68.11 & 67.50 \\
                                                                                            & 256                    & 328    & \textbf{70.39} & 83.8  & 87.28  & \textbf{64.35} & 62.76  & 61.76  & 60.82   & 69.36 & 68.79 & 68.26 & 67.63 \\
                                                                                            & 512                    & 656    & \textbf{70.74} & 83.91 & 87.33  & \textbf{64.69} & 63.05  & 62.06  & 61.14   & 69.63 & 69.00 & 68.50 & 67.88 \\
                                                                                            & 1024                   & 1312   & \textbf{71.05} & 84.13 & 87.46  & \textbf{64.85} & 63.22  & 62.19  & 61.26   & 69.78 & 69.16 & 68.60 & 67.99 \\
                                                                                            & 2048                   & 2624   & 71.17          & 84.27 & 87.67  & \textbf{64.99} & 63.33  & 62.29  & 61.33   & 69.90 & 69.24 & 68.68 & 68.05 \\ \midrule
\multirow{8}{*}{\begin{tabular}[c]{@{}c@{}}\SH \\ Interpolated\end{tabular}} & 12                     & 15     & 64.25          & 79.21 & 83.29  & 58.83          & 57.50  & 56.71  & 56.02   & 64.10 & 63.78 & 63.42 & 63.02 \\
                                                                                            & 24                     & 31     & 68.28          & 82.31 & 85.89  & 62.75          & 61.41  & 60.62  & 59.92   & 67.89 & 67.49 & 67.11 & 66.69 \\
                                                                                            & 48                     & 61     & 69.20          & 83.15 & 86.67  & 63.58          & 62.12  & 61.23  & 60.42   & 68.71 & 68.19 & 67.75 & 67.22 \\
                                                                                            & 96                     & 123    & 70.05          & 83.63 & 87.11  & 64.04          & 62.46  & 61.52  & 60.63   & 69.10 & 68.51 & 68.04 & 67.45 \\
                                                                                            & 192                    & 246    & 70.36          & 83.72 & 87.21  & 64.26          & 62.65  & 61.65  & 60.72   & 69.26 & 68.67 & 68.15 & 67.53 \\
                                                                                            & 384                    & 492    & 70.54          & 83.88 & 87.28  & 64.55          & 62.94  & 61.93  & 61.01   & 69.51 & 68.92 & 68.40 & 67.78 \\
                                                                                            & 768                    & 984    & 70.96          & 84.05 & 87.44  & 64.79          & 63.15  & 62.15  & 61.22   & 69.72 & 69.10 & 68.56 & 67.95 \\
                                                                                            & 1536                   & 1968   & 71.19          & 84.17 & 87.57  & 64.94          & 63.29  & 62.26  & 61.32   & 69.85 & 69.21 & 68.66 & 68.04 \\ \midrule\midrule
\multirow{9}{*}{\MH}                                                         & 8                      & 10     & \textbf{62.19} & 77.05 & 81.34  & \textbf{56.74} & 55.47  & 54.76  & 54.12   & 62.06 & 61.81 & 61.54 & 61.17 \\
                                                                                            & 16                     & 20     & \textbf{67.91} & 81.44 & 85.00  & \textbf{62.94} & 61.79  & 61.16  & 60.64   & 67.93 & 67.71 & 67.48 & 67.20 \\
                                                                                            & 32                     & 41     & \textbf{69.46} & 83.01 & 86.30  & \textbf{64.21} & 62.96  & 62.22  & 61.58   & 69.18 & 68.87 & 68.54 & 68.17 \\
                                                                                            & 64                     & 82     & \textbf{70.17} & 83.53 & 86.95  & \textbf{64.69} & 63.33  & 62.53  & 61.80   & 69.67 & 69.25 & 68.89 & 68.42 \\
                                                                                            & 128                    & 164    & \textbf{70.52} & 83.98 & 87.25  & \textbf{64.94} & 63.50  & 62.63  & 61.83   & 69.93 & 69.44 & 69.02 & 68.50 \\
                                                                                            & 256                    & 328    & \textbf{70.62} & 84.17 & 87.38  & \textbf{65.04} & 63.56  & 62.66  & 61.81   & 70.02 & 69.52 & 69.07 & 68.50 \\
                                                                                            & 512                    & 656    & \textbf{70.82} & 84.31 & 87.55  & \textbf{65.14} & 63.57  & 62.62  & 61.73   & 70.12 & 69.53 & 69.04 & 68.45 \\
                                                                                            & 1024                   & 1312   & \textbf{70.89} & 84.44 & 87.68  & \textbf{65.16} & 63.58  & 62.60  & 61.68   & 70.14 & 69.54 & 69.01 & 68.41 \\
                                                                                            & 2048                   & 2624   & 70.97          & 84.41 & 87.74  & \textbf{65.20} & 63.57  & 62.56  & 61.60   & 70.18 & 69.52 & 68.98 & 68.35 \\ \midrule
\multirow{8}{*}{\begin{tabular}[c]{@{}c@{}}\MH \\ Interpolated\end{tabular}} & 12                     & 15     & 65.89          & 80.04 & 83.68  & 60.84          & 59.66  & 58.98  & 58.37   & 65.94 & 65.72 & 65.45 & 65.08 \\
                                                                                            & 24                     & 31     & 68.76          & 82.48 & 85.87  & 63.64          & 62.42  & 61.74  & 61.13   & 68.64 & 68.35 & 68.07 & 67.71 \\
                                                                                            & 48                     & 61     & 69.96          & 83.40 & 86.65  & 64.58          & 63.2   & 62.42  & 61.72   & 69.53 & 69.10 & 68.75 & 68.32 \\
                                                                                            & 96                     & 123    & 70.40          & 83.83 & 87.04  & 64.86          & 63.46  & 62.62  & 61.84   & 69.82 & 69.38 & 68.98 & 68.48 \\
                                                                                            & 192                    & 246    & 70.64          & 84.09 & 87.37  & 65.00          & 63.53  & 62.66  & 61.83   & 69.98 & 69.49 & 69.05 & 68.50 \\
                                                                                            & 384                    & 492    & 70.69          & 84.25 & 87.41  & 65.09          & 63.56  & 62.64  & 61.76   & 70.05 & 69.51 & 69.04 & 68.46 \\
                                                                                            & 768                    & 984    & 70.84          & 84.40 & 87.63  & 65.16          & 63.59  & 62.62  & 61.71   & 70.14 & 69.55 & 69.03 & 68.44 \\
                                                                                            & 1536                   & 1968   & 70.88          & 84.39 & 87.71  & 65.18          & 63.59  & 62.58  & 61.64   & 70.16 & 69.54 & 68.99 & 68.38 \\ \midrule 
\end{tabular}
\label{tab:retrieval_IN1k}
}
\end{table}
\begin{table}[ht]
\centering
 \caption{Retrieve a shortlist of 200-NN with \retdim\space sized representations on \INVTwo~via exact search with L2 distance metric. Top-1 and mAP@10 entries (\%) where \SH~outperforms \FF~ are bolded. \MH~outperforms \FF~at all \retdim~and is thus not bolded.}
 \vspace{1mm}
 \resizebox{\columnwidth}{!}{%
\begin{tabular}{@{}c|c|c|ccc|cccc|cccc@{}}
\toprule
Config               & \retdim & MFLOPs & Top-1          & Top-5 & Top-10 & mAP@10         & mAP@25 & mAP@50 & mAP@100 & P@10  & P@25  & P@50  & P@100 \\ \midrule
\multirow{9}{*}{\FF} & 8     & ~~10   & 48.79          & 64.70 & 69.72  & 43.04          & 41.89  & 41.42  & 41.17   & 48.43 & 48.27 & 48.25 & 48.19 \\
                     & 16    & ~~20   & 55.08          & 69.50 & 74.08  & 49.63          & 48.53  & 48.06  & 47.75   & 54.76 & 54.64 & 54.53 & 54.39 \\
                     & 32    & ~~41   & 56.69          & 71.10 & 76.47  & 51.11          & 49.85  & 49.17  & 48.65   & 56.23 & 55.96 & 55.71 & 55.42 \\
                     & 64    & ~~82   & 57.37          & 72.71 & 77.48  & 51.28          & 49.75  & 48.85  & 47.99   & 56.65 & 56.14 & 55.71 & 55.15 \\
                     & 128   & ~164   & 57.17          & 73.31 & 78.64  & 50.07          & 48.09  & 46.79  & 45.58   & 55.75 & 54.89 & 54.12 & 53.28 \\
                     & 256   & ~328   & 57.09          & 74.04 & 79.24  & 49.11          & 46.66  & 44.99  & 43.35   & 55.02 & 53.77 & 52.74 & 51.53 \\
                     & 512   & ~656   & 57.12          & 73.91 & 79.32  & 48.95          & 46.25  & 44.37  & 42.42   & 54.88 & 53.49 & 52.29 & 50.83 \\
                     & 1024  & 1312   & 57.53          & 74.17 & 79.55  & 48.27          & 45.41  & 43.36  & 41.26   & 54.31 & 52.84 & 51.49 & 49.87 \\
                     & 2048  & 2624   & 57.84          & 74.59 & 79.45  & 49.99          & 47.47  & 45.66  & 43.87   & 55.89 & 54.63 & 53.45 & 52.12 \\ \midrule
\multirow{9}{*}{\SH} & 8     & ~~10   & 47.05          & 62.53 & 67.60  & 40.79          & 39.47  & 38.78  & 38.16   & 46.03 & 45.77 & 45.54 & 45.17 \\
                     & 16    & ~~20   & \textbf{55.73} & 70.54 & 74.86  & \textbf{49.86} & 48.57  & 47.84  & 47.26   & 54.97 & 54.71 & 54.44 & 54.10 \\
                     & 32    & ~~41   & \textbf{57.33} & 71.61 & 76.64  & \textbf{51.26} & 49.92  & 49.09  & 48.42   & 56.46 & 56.11 & 55.70 & 55.30 \\
                     & 64    & ~~82   & \textbf{57.90} & 72.55 & 77.44  & \textbf{51.89} & 50.29  & 49.34  & 48.53   & 57.06 & 56.45 & 55.97 & 55.43 \\
                     & 128   & ~164   & \textbf{57.73} & 72.79 & 77.28  & \textbf{52.02} & 50.38  & 49.49  & 48.62   & 57.13 & 56.58 & 56.15 & 55.58 \\
                     & 256   & ~328   & \textbf{58.22} & 72.77 & 77.67  & \textbf{52.16} & 50.61  & 49.67  & 48.81   & 57.30 & 56.79 & 56.33 & 55.77 \\
                     & 512   & ~656   & \textbf{58.46} & 73.00 & 77.88  & \textbf{52.52} & 50.97  & 50.02  & 49.16   & 57.65 & 57.10 & 56.64 & 56.08 \\
                     & 1024  & 1312   & \textbf{58.71} & 73.29 & 78.00  & \textbf{52.70} & 51.13  & 50.17  & 49.30   & 57.83 & 57.26 & 56.77 & 56.20 \\
                     & 2048  & 2624   & \textbf{58.86} & 73.17 & 78.00  & \textbf{52.88} & 51.25  & 50.26  & 49.36   & 57.95 & 57.35 & 56.85 & 56.25 \\ \midrule
\multirow{9}{*}{\MH} & 8     & ~~10   & \textbf{50.41}          & 65.56 & 70.27  & \textbf{45.51}          & 44.38  & 43.71  & 43.17   & 50.55 & 50.44 & 50.17 & 49.91 \\
                     & 16    & ~~20   & \textbf{56.64}          & 70.19 & 74.61  & \textbf{50.98}          & 49.76  & 49.16  & 48.69   & 55.90 & 55.66 & 55.52 & 55.29 \\
                     & 32    & ~~41   & \textbf{57.96}          & 71.88 & 76.41  & \textbf{52.06}          & 50.78  & 50.09  & 49.54   & 57.18 & 56.83 & 56.57 & 56.27 \\
                     & 64    & ~~82   & \textbf{58.94}          & 72.74 & 77.17  & \textbf{52.65}          & 51.24  & 50.44  & 49.76   & 57.72 & 57.29 & 56.94 & 56.52 \\
                     & 128   & ~164   & \textbf{59.13}          & 73.07 & 77.49  & \textbf{52.94}          & 51.42  & 50.53  & 49.74   & 58.00 & 57.47 & 57.05 & 56.55 \\
                     & 256   & ~328   & \textbf{59.18}          & 73.64 & 77.75  & \textbf{52.96}          & 51.45  & 50.52  & 49.70   & 58.01 & 57.53 & 57.06 & 56.54 \\
                     & 512   & ~656   & \textbf{59.40}          & 73.85 & 77.97  & \textbf{53.01}          & 51.39  & 50.46  & 49.61   & 58.11 & 57.49 & 57.04 & 56.48 \\
                     & 1024  & 1312   & \textbf{59.11}          & 73.77 & 77.92  & \textbf{52.98}          & 51.37  & 50.40  & 49.54   & 58.13 & 57.51 & 57.00 & 56.45 \\
                     & 2048  & 2624   & \textbf{59.63}          & 73.84 & 77.97  & \textbf{52.96}          & 51.34  & 50.34  & 49.44   & 58.07 & 57.48 & 56.95 & 56.36 \\ \bottomrule
\end{tabular}
\label{tab:retrieval_INV2}
}
\end{table}
\begin{table}[t]
\centering
 \caption{Retrieve a shortlist of 200-NN with \retdim~ sized representations on \InIVk~ via exact search with L2 distance metric. \SH~ and \FF~ models are omitted for clarity and compute/inference time costs. All entries are in \%.}
 \vspace{1mm}
 \resizebox{\columnwidth}{!}{%
\begin{tabular}{@{}c|c|c|ccc|cccc|cccc@{}}
\toprule
Config                                                                      & \retdim & MFLOPs & Top-1 & Top-5 & Top-10 & mAP@10 & mAP@25 & mAP@50 & mAP@100 & P@10  & P@25  & P@50  & P@100 \\ \midrule
\multirow{9}{*}{\MH}                                                        & 8       & ~~34   & 10.60 & 26.23 & 35.57  & ~5.32  & ~4.29  & ~3.76  & 3.36    & ~9.13 & ~8.77 & ~8.46 & ~8.13 \\
                                                                            & 16      & ~~67   & 16.74 & 36.91 & 47.28  & ~8.64  & ~6.83  & ~5.84  & 5.05    & 13.82 & 12.79 & 12.04 & 13.27 \\
                                                                            & 32      & ~134   & 21.54 & 43.75 & 54.11  & 11.36  & ~8.88  & ~7.47  & 6.31    & 17.25 & 15.67 & 14.47 & 13.27 \\
                                                                            & 64      & ~269   & 25.00 & 47.97 & 58.25  & 13.38  & 10.40  & ~8.67  & 7.23    & 19.68 & 17.64 & 16.14 & 14.65 \\
                                                                            & 128     & ~538   & 27.27 & 50.35 & 60.47  & 14.77  & 11.47  & ~9.53  & 7.91    & 21.25 & 18.95 & 17.26 & 15.59 \\
                                                                            & 256     & 1076   & 28.53 & 51.95 & 61.90  & 15.66  & 12.19  & 10.12  & 8.38    & 22.28 & 19.81 & 18.01 & 16.22 \\
                                                                            & 512     & 2151   & 29.46 & 53.03 & 62.81  & 16.29  & 12.70  & 10.55  & 8.72    & 22.96 & 20.42 & 18.54 & 16.68 \\
                                                                            & 1024    & 4303   & 30.23 & 53.72 & 63.45  & 16.76  & 13.08  & 10.86  & 8.97    & 23.48 & 20.88 & 18.93 & 17.00 \\
                                                                            & 2048    & 8606   & 30.87 & 54.32 & 64.02  & 17.20  & 13.43  & 11.14  & 9.19    & 23.97 & 21.28 & 19.28 & 17.30 \\\midrule
\multirow{8}{*}{\begin{tabular}[c]{@{}c@{}}\MH-\\ Interpolated\end{tabular}} & 12      & ~~50   & 14.04 & 32.56 & 42.71  & ~7.16  & ~5.70  & ~4.92  & 4.32    & 11.81 & 11.08 & 10.52 & ~9.94 \\
                                                                            & 24      & ~101   & 19.49 & 40.82 & 51.26  & 10.17  & ~7.98  & ~6.75  & 5.75    & 15.76 & 14.43 & 13.42 & 12.40 \\
                                                                            & 48      & ~202   & 23.51 & 46.23 & 56.56  & 12.49  & ~9.72  & ~8.13  & 6.81    & 18.62 & 16.75 & 15.39 & 14.04 \\
                                                                            & 96      & ~403   & 26.25 & 49.32 & 59.48  & 14.15  & 11.00  & ~9.15  & 7.61    & 20.55 & 18.36 & 16.78 & 15.17 \\
                                                                            & 192     & ~807   & 27.94 & 51.32 & 61.32  & 15.29  & 11.89  & ~9.88  & 8.18    & 21.86 & 19.46 & 17.71 & 15.96 \\
                                                                            & 384     & 1614   & 29.03 & 52.53 & 62.45  & 15.99  & 12.46  & 10.35  & 8.56    & 22.64 & 20.14 & 18.29 & 16.47 \\
                                                                            & 768     & 3227   & 29.87 & 53.36 & 63.13  & 16.54  & 12.90  & 10.71  & 8.85    & 23.23 & 20.67 & 18.75 & 16.85 \\
                                                                            & 1536    & 6454   & 30.52 & 54.02 & 63.79  & 16.99  & 13.27  & 11.01  & 9.08    & 23.73 & 21.09 & 19.12 & 17.16 \\ \bottomrule
\end{tabular}
\label{tab:retrieval_IN4k}
}
\end{table}

Retrieval performance on \InIk, \textit{i.e.} the training distribution, is shown in Table~ \ref{tab:retrieval_IN1k}. \MH~outperforms \FF~models for nearly all representation size for both top-1 and mAP@10, and especially at low representation size (\retdim~$\leq 32$). \SH~loses out to \FF~significantly only at \retdim~$ = 8$. This indicates that training ResNet50 models via the \mrl training paradigm improves retrieval at low representation size over models explicitly trained at those representation size (\FF-$8...2048$). 

We carried out all retrieval experiments at \retdim~$\in \{8, 16, 32, 64, 128, 256, 512, 1024, 2048\}$, as these were the representation sizes which were a part of the \lstinline{nesting_list} at which losses were added during training, as seen in Algorithm~\ref{code:NCE-Loss}, Appendix~\ref{sec:code}. To examine whether \nrl is able to learn \mrs at dimensions in between the representation size for which it was trained, we also tabulate the performance of \MH at interpolated \retdim~$\in \{12, 24, 48, 96, 192, 384, 768, 1536\}$ as \MH--Interpolated and \SH--Interpolated (see Table~\ref{tab:retrieval_IN1k}). We observed that performance scaled nearly monotonically between the original representation size and the interpolated representation size as we increase \retdim, which demonstrates that \nrl is able to learn \mrs at nearly all representation size $m\in[8, 2048]$ despite optimizing only for $|\mathcal{M}|$ nested representation sizes. 

We examined the robustness of \nrl for retrieval on out-of-domain datasets \INVTwo and \InIVk, as shown in Table~\ref{tab:retrieval_INV2} and Table~\ref{tab:retrieval_IN4k} respectively. On \INVTwo, we observed that \MH outperformed \FF~at all \retdim~on top-1 Accuracy and mAP@10, and \SH outperformed \FF~ at all \retdim~ except \retdim~$ = 8$. This demonstrates the robustness of the learned \mrs for out-of-domain image retrieval.

\section{Adaptive Retrieval}
\label{sec:adaptive_retrieval}
The time complexity of retrieving a shortlist of k-NN often scales as $O(d)$, where $d = $\retdim, for a fixed k and $N$. We thus will have a theoretical $256\times$ higher cost for \retdim~$ = 2048$ over \retdim~$ = 8$. We discuss search complexity in more detail in Appendix~\ref{sec:appendix_real-world-perf}. In an attempt to replicate performance at higher \retdim~ while using less FLOPs, we perform adaptive retrieval via retrieving a k-NN shortlist with representation size \retdim, and then re-ranking the shortlist with representations of size \rerankdim. Adaptive retrieval for a shortlist length $k = 200$ is shown in Table~\ref{tab:rerank-in1k} for \InIk, and in Table~\ref{tab:rerank-in4k} for \InIVk. On \InIk, we are able to achieve comparable performance to retrieval with \retdim~$ = 2048$ (from Table \ref{tab:retrieval_IN1k}) with \retdim~$= 16$ at $128\times$ less MFLOPs/Query (used interchangeably with MFLOPs). Similarly, on \InIVk, we are able to achieve comparable performance to retrieval with \retdim~$ = 2048$ (from Table \ref{tab:retrieval_IN4k}) with \retdim~$ = 64$ on \InIk~and \InIVk, at $32\times$ less MFLOPs. This demonstrates the value of intelligent routing techniques which utilize appropriately sized \mrs for retrieval. 
\newpage
\begin{table}[!ht]
\centering
 \caption{Retrieve a shortlist of k-NN with \retdim\space sized representations on \InIk~with \MH~representations, and then re-order the neighbors shortlist with L2 distances using \rerankdim\space sized representations. Top-1 and mAP@10 entries (\%) that are within $0.1\%$ of the maximum value achievable without reranking on \MH~representations, as seen in Table~\ref{tab:retrieval_IN1k}, are bolded.}
 \vspace{1mm}
 \resizebox{.9\columnwidth}{!}{
\begin{tabular}{@{}c|c|c|c|c|cccc|cccc@{}}
\toprule
\multirow{37}{*}{\begin{sideways}
Shortlist Length = 200 \end{sideways}} & \retdim       & \rerankdim & MFLOPs                & Top-1            & mAP@10         & mAP@25 & mAP@50 & mAP@100 & P@10  & P@25  & P@50  & P@100 \\ \cmidrule(l){2-13} 
                                         & \multirow{8}{*}{8}   & ~~16        & \multirow{8}{*}{~~10} & 68.21          & 63.35          & 62.25  & 61.70  & 61.19   & 68.32 & 68.14 & 67.96 & 67.65 \\
                                         &                      & ~32         &                       & 69.42          & 64.12          & 62.81  & 62.03  & 61.32   & 69.04 & 68.63 & 68.22 & 67.71 \\
                                         &                      & ~64         &                       & 70.05          & 64.46          & 63.03  & 62.14  & 61.29   & 69.37 & 68.83 & 68.32 & 67.66 \\
                                         &                      & ~128        &                       & 70.34          & 64.68          & 63.16  & 62.21  & 61.27   & 69.59 & 68.96 & 68.38 & 67.65 \\
                                         &                      & ~256        &                       & 70.40          & 64.77          & 63.21  & 62.23  & 61.26   & 69.66 & 69.02 & 68.41 & 67.65 \\
                                         &                      & ~512        &                       & 70.60          & 64.86          & 63.22  & 62.21  & 61.22   & 69.74 & 69.02 & 68.39 & 67.62 \\
                                         &                      & 1024        &                       & 70.71          & 64.88          & 63.23  & 62.20  & 61.20   & 69.76 & 69.01 & 68.39 & 67.60 \\
                                         &                      & 2048        &                       & 70.81          & 64.90          & 63.22  & 62.17  & 61.16   & 69.77 & 68.99 & 68.36 & 67.57 \\\cmidrule(l){2-13} 
                                         & \multirow{7}{*}{16}  & ~32         & \multirow{7}{*}{~~21} & 69.47          & 64.27          & 63.04  & 62.36  & 61.75   & 69.21 & 68.90 & 68.58 & 68.12 \\
                                         &                      & ~64         &                       & 70.16          & 64.74          & 63.42  & 62.66  & 61.94   & 69.66 & 69.22 & 68.81 & 68.22 \\
                                         &                      & ~128        &                       & 70.52          & 65.00          & 63.60  & 62.77  & 61.98   & 69.91 & 69.36 & 68.89 & 68.24 \\
                                         &                      & ~256        &                       & 70.55          & \textbf{65.10} & 63.67  & 62.82  & 62.01   & 69.98 & 69.43 & 68.92 & 68.25 \\
                                         &                      & ~512        &                       & 70.74          & \textbf{65.21} & 63.70  & 62.83  & 62.00   & 70.08 & 69.43 & 68.92 & 68.24 \\
                                         &                      & 1024        &                       & 70.83          & \textbf{65.23} & 63.72  & 62.83  & 61.99   & 70.08 & 69.45 & 68.92 & 68.23 \\
                                         &                      & 2048        &                       & \textbf{70.90} & \textbf{65.27} & 63.73  & 62.82  & 61.97   & 70.10 & 69.44 & 68.90 & 68.21 \\\cmidrule(l){2-13} 
                                         & \multirow{6}{*}{32}  & ~64         & \multirow{6}{*}{~~41} & 70.16          & 64.69          & 63.35  & 62.57  & 61.93   & 69.68 & 69.26 & 68.92 & 68.51 \\
                                         &                      & ~128        &                       & 70.52          & 64.97          & 63.54  & 62.73  & 62.04   & 69.95 & 69.47 & 69.06 & 68.59 \\
                                         &                      & ~256        &                       & 70.63          & 65.07          & 63.63  & 62.79  & 62.07   & 70.04 & 69.55 & 69.12 & 68.61 \\
                                         &                      & ~512        &                       & 70.82          & \textbf{65.17} & 63.66  & 62.80  & 62.06   & 70.11 & 69.57 & 69.12 & 68.60 \\
                                         &                      & 1024        &                       & \textbf{70.89} & \textbf{65.20} & 63.68  & 62.80  & 62.04   & 70.15 & 69.59 & 69.12 & 68.59 \\
                                         &                      & 2048        &                       & \textbf{70.97} & \textbf{65.24} & 63.70  & 62.79  & 62.02   & 70.19 & 69.59 & 69.10 & 68.56 \\\cmidrule(l){2-13} 
                                         & \multirow{5}{*}{64}  & ~128        & \multirow{5}{*}{~~82} & 70.51          & 64.94          & 63.50  & 62.64  & 61.88   & 69.94 & 69.44 & 69.02 & 68.54 \\
                                         &                      & ~256        &                       & 70.63          & 65.04          & 63.57  & 62.69  & 61.91   & 70.02 & 69.52 & 69.08 & 68.57 \\
                                         &                      & ~512        &                       & 70.83          & \textbf{65.14} & 63.59  & 62.67  & 61.87   & 70.12 & 69.54 & 69.06 & 68.54 \\
                                         &                      & 1024        &                       & \textbf{70.89} & \textbf{65.16} & 63.59  & 62.65  & 61.85   & 70.15 & 69.54 & 69.05 & 68.52 \\
                                         &                      & 2048        &                       & \textbf{70.97} & \textbf{65.20} & 63.59  & 62.63  & 61.82   & 70.18 & 69.53 & 69.03 & 68.49 \\\cmidrule(l){2-13} 
                                         & \multirow{4}{*}{128} & ~256        & \multirow{4}{*}{~164} & 70.63          & 65.04          & 63.56  & 62.66  & 61.82   & 70.02 & 69.52 & 69.07 & 68.51 \\
                                         &                      & ~512        &                       & 70.82          & \textbf{65.14} & 63.58  & 62.63  & 61.77   & 70.11 & 69.54 & 69.04 & 68.47 \\
                                         &                      & 1024        &                       & \textbf{70.89} & \textbf{65.16} & 63.58  & 62.60  & 61.73   & 70.14 & 69.54 & 69.02 & 68.45 \\
                                         &                      & 2048        &                       & \textbf{70.97} & \textbf{65.20} & 63.57  & 62.57  & 61.68   & 70.18 & 69.52 & 68.99 & 68.41 \\\cmidrule(l){2-13} 
                                         & \multirow{3}{*}{256} & ~512        & \multirow{3}{*}{~328} & 70.82          & \textbf{65.14} & 63.57  & 62.62  & 61.74   & 70.12 & 69.53 & 69.04 & 68.45 \\
                                         &                      & 1024        &                       & \textbf{70.88} & \textbf{65.16} & 63.58  & 62.60  & 61.69   & 70.14 & 69.54 & 69.01 & 68.41 \\
                                         &                      & 2048        &                       & \textbf{70.97} & \textbf{65.20} & 63.56  & 62.56  & 61.62   & 70.18 & 69.52 & 68.98 & 68.37 \\\cmidrule(l){2-13} 
                                         & \multirow{2}{*}{512} & 1024        & \multirow{2}{*}{~656} & \textbf{70.90}  & \textbf{65.16} & 63.58  & 62.60  & 61.68   & 70.14 & 69.54 & 69.01 & 68.41 \\
                                         &                      & 2048        &                       & \textbf{70.98} & \textbf{65.20} & 63.57  & 62.56  & 61.60   & 70.18 & 69.52 & 68.98 & 68.35 \\\cmidrule(l){2-13} 
                                         & 1024                 & 2048        & 1312                  & \textbf{70.97} & \textbf{65.20} & 63.57  & 62.56  & 61.60   & 70.18 & 69.52 & 68.98 & 68.35 \\ \bottomrule
\end{tabular}
\label{tab:rerank-in1k}
}
\end{table}
\begin{table}[!ht]
\centering
 \caption{\small Retrieve a shortlist of k-NN with \retdim\space sized representations on \InIVk~with \MH~representations, and then re-order the neighbors shortlist with L2 distances using \rerankdim\space sized representations. Top-1 and mAP@10 entries (\%)  that are within $0.1\%$ of the maximum value achievable without reranking on \MH~representations, as seen in Table~\ref{tab:retrieval_IN4k}, are bolded.}
 \vspace{1mm}
 \resizebox{.9\columnwidth}{!}{
\begin{tabular}{@{}c|c|c|c|c|cccc|cccc@{}}
\toprule
\multirow{37}{*}{\begin{sideways}
Shortlist Length = 200 \end{sideways}} & \retdim              & \rerankdim & MFLOPs                & Top-1          & mAP@10         & mAP@25 & mAP@50 & mAP@100 & P@10  & P@25  & P@50  & P@100 \\ \cmidrule(l){2-13} 
                                         & \multirow{8}{*}{8}   & ~~16       & \multirow{8}{*}{~~34} & 16.84          & ~8.70          & ~6.88  & 5.88   & 5.08    & 13.86 & 12.80 & 11.98 & 11.10 \\
                                         &                      & ~32        &                       & 20.73          & 10.66          & ~8.19  & 6.77   & 5.61    & 16.18 & 14.39 & 13.02 & 11.61 \\
                                         &                      & ~64        &                       & 23.11          & 11.91          & ~9.03  & 7.36   & 6.00    & 17.56 & 15.34 & 13.67 & 11.99 \\
                                         &                      & ~128       &                       & 24.63          & 12.71          & ~9.59  & 7.76   & 6.25    & 18.42 & 15.94 & 14.08 & 12.22 \\
                                         &                      & ~256       &                       & 25.5           & 13.24          & ~9.96  & 8.03   & 6.42    & 19.00 & 16.35 & 14.36 & 12.37 \\
                                         &                      & ~512       &                       & 26.07          & 13.59          & 10.21  & 8.20   & 6.53    & 19.37 & 16.62 & 14.54 & 12.46 \\
                                         &                      & 1024       &                       & 26.52          & 13.85          & 10.40  & 8.34   & 6.61    & 19.65 & 16.80 & 14.68 & 12.53 \\
                                         &                      & 2048       &                       & 26.94          & 14.11          & 10.57  & 8.45   & 6.68    & 19.92 & 16.98 & 14.79 & 12.58 \\\cmidrule(l){2-13} 
                                         & \multirow{7}{*}{16}  & ~32        & \multirow{7}{*}{~~67} & 21.44          & 11.24          & ~8.72  & 7.26   & 6.02    & 17.02 & 15.30 & 13.92 & 12.41 \\
                                         &                      & ~64        &                       & 24.36          & 12.78          & ~9.75  & 7.96   & 6.43    & 18.72 & 16.41 & 14.63 & 12.74 \\
                                         &                      & ~128       &                       & 26.08          & 13.70          & 10.39  & 8.39   & 6.69    & 19.68 & 17.07 & 15.05 & 12.94 \\
                                         &                      & ~256       &                       & 26.99          & 14.27          & 10.79  & 8.67   & 6.85    & 20.27 & 17.48 & 15.31 & 13.07 \\
                                         &                      & ~512       &                       & 27.60          & 14.66          & 11.06  & 8.86   & 6.97    & 20.67 & 17.75 & 15.50 & 13.16 \\
                                         &                      & 1024       &                       & 28.12          & 14.94          & 11.26  & 8.99   & 7.05    & 20.96 & 17.95 & 15.62 & 13.22 \\
                                         &                      & 2048       &                       & 28.56          & 15.21          & 11.43  & 9.11   & 7.12    & 21.23 & 18.13 & 15.73 & 13.27 \\\cmidrule(l){2-13} 
                                         & \multirow{6}{*}{32}  & ~64        & \multirow{6}{*}{~134} & 24.99          & 13.35          & 10.35  & ~8.59  & 7.09    & 19.61 & 17.52 & 15.92 & 14.21 \\
                                         &                      & ~128       &                       & 27.17          & 14.61          & 11.27  & ~9.26  & 7.51    & 20.99 & 18.52 & 16.62 & 14.59 \\
                                         &                      & ~256       &                       & 28.33          & 15.37          & 11.83  & ~9.67  & 7.77    & 21.80 & 19.12 & 17.05 & 14.81 \\
                                         &                      & ~512       &                       & 29.12          & 15.88          & 12.20  & ~9.94  & 7.93    & 22.33 & 19.51 & 17.32 & 14.94 \\
                                         &                      & 1024       &                       & 29.78          & 16.25          & 12.47  & 10.13  & 8.05    & 22.71 & 19.79 & 17.5  & 15.03 \\
                                         &                      & 2048       &                       & 30.33          & 16.59          & 12.72  & 10.30  & 8.16    & 23.07 & 20.05 & 17.66 & 15.11 \\\cmidrule(l){2-13} 
                                         & \multirow{5}{*}{64}  & ~128       & \multirow{5}{*}{~269} & 27.27          & 14.76          & 11.47  & ~9.51  & 7.85    & 21.25 & 18.92 & 17.20 & 15.40 \\
                                         &                      & ~256       &                       & 28.54          & 15.64          & 12.15  & 10.05  & 8.21    & 22.24 & 19.71 & 17.81 & 15.76 \\
                                         &                      & ~512       &                       & 29.45          & 16.25          & 12.62  & 10.40  & 8.44    & 22.88 & 20.24 & 18.20 & 15.97 \\
                                         &                      & 1024       &                       & 30.19          & 16.69          & 12.96  & 10.66  & 8.60    & 23.35 & 20.61 & 18.46 & 16.10 \\
                                         &                      & 2048       &                       & \textbf{30.81} & \textbf{17.10} & 13.27  & 10.88  & 8.74    & 23.79 & 20.93 & 18.69 & 16.21 \\\cmidrule(l){2-13} 
                                         & \multirow{4}{*}{128} & ~256       & \multirow{4}{*}{~538} & 28.54          & 15.66          & 12.19  & 10.12  & 8.36    & 22.28 & 19.81 & 18.00 & 16.16 \\
                                         &                      & ~512       &                       & 29.45          & 16.29          & 12.69  & 10.53  & 8.66    & 22.96 & 20.41 & 18.50 & 16.48 \\
                                         &                      & 1024       &                       & 30.22          & 16.76          & 13.07  & 10.83  & 8.86    & 23.47 & 20.84 & 18.83 & 16.68 \\
                                         &                      & 2048       &                       & \textbf{30.86} & \textbf{17.19} & 13.41  & 11.09  & 9.03    & 23.95 & 21.22 & 19.12 & 16.84 \\\cmidrule(l){2-13} 
                                         & \multirow{3}{*}{256} & ~512       & \multirow{3}{*}{1076} & 29.45          & 16.29          & 12.70  & 10.55  & 8.71    & 22.97 & 20.42 & 18.54 & 16.66 \\
                                         &                      & 1024       &                       & 30.21          & 16.76          & 13.08  & 10.86  & 8.95    & 23.48 & 20.87 & 18.92 & 16.94 \\
                                         &                      & 2048       &                       & \textbf{30.85} & \textbf{17.20} & 13.43  & 11.14  & 9.15    & 23.97 & 21.27 & 19.26 & 17.16 \\\cmidrule(l){2-13} 
                                         & \multirow{2}{*}{512} & 1024       & \multirow{2}{*}{2152} & 30.22          & 16.76          & 13.08  & 10.86  & 8.97    & 23.48 & 20.88 & 18.93 & 17.00 \\
                                         &                      & 2048       &                       & \textbf{30.87} & \textbf{17.20} & 13.43  & 11.14  & 9.19    & 23.97 & 21.28 & 19.28 & 17.28 \\\cmidrule(l){2-13} 
                                         & 1024                 & 2048       & 4303                  & \textbf{30.87} & \textbf{17.20} & 13.43  & 11.15  & 9.19    & 23.97 & 21.28 & 19.28 & 17.29 \\ \bottomrule
\end{tabular}
\label{tab:rerank-in4k}
}
\end{table}

\paragraph{Funnel Retrieval.}We also designed a simple cascade policy which we call funnel retrieval to successively improve and refine the k-NN shortlist at increasing \retdim. This was an attempt to remove the dependence on manual choice of \retdim~\&~\rerankdim. We retrieved a shortlist at \retdim~and then re-ranked the shortlist five times while simultaneously increasing \rerankdim~(rerank cascade) and decreasing the shortlist length (shortlist cascade), which resembles a funnel structure. We tabulate the performance of funnel retrieval in various configurations in Table~\ref{tab:rerankcascade_in1k} on \InIk, and in Table~\ref{tab:rerankcascade_in4k} on \InIVk. With funnel retrieval on \InIk, we were able to achieve top-1 accuracy within $0.1\%$ of retrieval with \retdim~$= 2048$ (as in Table~\ref{tab:retrieval_IN1k}) with a funnel with \retdim~$ = 16$, with $128\times$ less MFLOPs. Similarly, we are able to achieve equivalent top-1 accuracy within $0.15\%$ of retrieval at \retdim~$ = 2048$ (as in Table~\ref{tab:retrieval_IN4k}) with funnel retrieval at \retdim~$ = 32$ on \InIVk, with $64\times$ less MFLOPs. This demonstrates that with funnel retrieval, we can emulate the performance of retrieval with \retdim~$ = 2048$ with a fraction of the MFLOPs. 

\begin{table}[ht!]
\centering
 \caption{Retrieve a shortlist of k-NN with \retdim\space sized representations on \InIk\space with \MH. This shortlist is then reranked with funnel retrieval, which uses a rerank cascade with a one-to-one mapping with a monotonically decreasing shortlist length as shown in the shortlist cascade. Top-1 and mAP@10 entries (\%) within $0.1\%$ of the maximum achievable without reranking on \MH~representations, as seen in Table~\ref{tab:retrieval_IN1k}, are bolded.}
 \vspace{1mm}
 \resizebox{\columnwidth}{!}{
\begin{tabular}{@{}c|c|c|c|ccccc@{}}
\toprule
\retdim             & Rerank Cascade                           & Shortlist Cascade           & MFLOPs & Top-1          & Top-5 & Top-10 & mAP@10         & P@10  \\ \midrule
\multirow{3}{*}{8}  & \multirow{3}{*}{16$\to$32$\to$64$\to$128$\to$2048}   & ~~200$\to$100$\to$50$\to$25$\to$10 & 10.28  & 70.22          & 82.63 & 85.49  & 64.06          & 68.65 \\
                    &                                          & ~~400$\to$200$\to$50$\to$25$\to$10 & 10.29  & 70.46          & 83.13 & 86.08  & 64.43          & 69.10 \\
                    &                                          & 800$\to$400$\to$200$\to$50$\to$10  & 10.31  & 70.58          & 83.54 & 86.53  & 64.62          & 69.37 \\\midrule
\multirow{3}{*}{16} & \multirow{3}{*}{32$\to$64$\to$128$\to$256$\to$2048}  & ~~200$\to$100$\to$50$\to$25$\to$10 & 20.54  & \textbf{70.90} & 83.96 & 86.85  & \textbf{65.19} & 69.97 \\
                    &                                          & ~~400$\to$200$\to$50$\to$25$\to$10 & 20.56  & \textbf{70.95} & 84.05 & 87.04  & \textbf{65.18} & 70.00 \\
                    &                                          & 800$\to$400$\to$200$\to$50$\to$10  & 20.61  & \textbf{70.96} & 84.18 & 87.22  & \textbf{65.14} & 70.01 \\\midrule
\multirow{3}{*}{32} & \multirow{3}{*}{64$\to$128$\to$256$\to$512$\to$2048} & ~~200$\to$100$\to$50$\to$25$\to$10 & 41.07  & \textbf{70.96} & 84.32 & 87.47  & \textbf{65.21} & 70.11 \\
                    &                                          & ~~400$\to$200$\to$50$\to$25$\to$10 & 41.09  & \textbf{70.97} & 84.32 & 87.47  & \textbf{65.19} & 70.11 \\
                    &                                          & 800$\to$400$\to$200$\to$50$\to$10  & 41.20  & \textbf{70.97} & 84.36 & 87.53  & \textbf{65.18} & 70.11 \\ \bottomrule
\end{tabular}
\label{tab:rerankcascade_in1k}
}
\end{table}
\begin{table}[h]
\centering
 \caption{Retrieve a shortlist of k-NN with \retdim\space sized representations on \InIVk with \MH. This shortlist is then reranked with funnel retrieval, which uses a rerank cascade with a one-to-one mapping with a monotonically decreasing shortlist length as shown in the shortlist cascade. Top-1 and mAP@10 entries (\%) within $0.15\%$ of the maximum achievable without reranking on \MH~representations, as seen in Table~\ref{tab:retrieval_IN4k}, are bolded.}
 \vspace{1mm}
 \resizebox{\columnwidth}{!}{
\begin{tabular}{@{}c|c|c|c|ccccc@{}}
\toprule
\retdim             & Rerank Cascade                             & Shortlist Cascade           & MFLOPs & Top-1 & Top-5 & Top-10 & mAP@10 & P@10 \\ \midrule
\multirow{3}{*}{8}  & \multirow{3}{*}{16$\to$32$\to$64$\to$128$\to$2048}     & ~~200$\to$100$\to$50$\to$25$\to$10 & ~33.65       & 26.20 & 46.45 & 54.12  & 12.79  & 17.85   \\
                    &                                            & ~~400$\to$200$\to$50$\to$25$\to$10 & ~33.66       & 26.55 & 47.02 & 54.72  & 13.02  & 18.15   \\
                    &                                            & 800$\to$400$\to$200$\to$50$\to$10  & ~33.68       & 26.83 & 47.54 & 55.35  & 13.24  & 18.44   \\\midrule
\multirow{3}{*}{16} & \multirow{3}{*}{32$\to$64$\to$128$\to$256$\to$2048}    & ~~200$\to$100$\to$50$\to$25$\to$10 & ~67.28       & 29.51 & 51.44 & 59.56  & 15.27  & 21.03   \\
                    &                                            & ~~400$\to$200$\to$50$\to$25$\to$10 & ~67.29       & 29.66 & 51.71 & 59.88  & 15.42  & 21.22   \\
                    &                                            & 800$\to$400$\to$200$\to$50$\to$10  & ~67.34       & 29.79 & 52.00 & 60.25  & 15.55  & 21.41   \\\midrule
\multirow{3}{*}{32} & \multirow{3}{*}{64$\to$128$\to$256$\to$512$\to$2048}   & ~~200$\to$100$\to$50$\to$25$\to$10 & 134.54       & 30.64 & 53.52 & 62.16  & 16.45  & 22.64   \\
                    &                                            & ~~400$\to$200$\to$50$\to$25$\to$10 & 134.56       & 30.69 & 53.65 & 62.31  & 16.51  & 22.73   \\
                    &                                            & 800$\to$400$\to$200$\to$50$\to$10  & 134.66       & \textbf{30.72} & 53.78 & 62.43  & 16.55  & 22.79   \\\midrule
\multirow{3}{*}{64} & \multirow{3}{*}{128$\to$256$\to$512$\to$1024$\to$2048} & ~~200$\to$100$\to$50$\to$25$\to$10 & 269.05       & \textbf{30.81} & 54.06 & 63.15  & 16.87  & 23.34   \\
                    &                                            & ~~400$\to$200$\to$50$\to$25$\to$10 & 269.10       & \textbf{30.84} & 54.20 & 63.31  & 16.92  & 23.42   \\
                    &                                            & 800$\to$400$\to$200$\to$50$\to$10  & 269.31       & \textbf{30.87} & 54.27 & 63.42  & 16.95  & 23.46   \\ \midrule
\end{tabular}
\label{tab:rerankcascade_in4k}
}
\end{table}

\section{Few-shot and Sample Efficiency}
\label{sec:appendix_few_shot}
We compared \MH, \SH, and \FF~on various benchmarks to observe the effect of representation size on sample efficiency. We used Nearest Class Means \cite{sanchez1997use} for classification which has been shown to be effective in the few-shot regime \cite{chen2021meta}.

\paragraph{\INVTwo.}
Representations are evaluated on \INVTwo~with the n-shot k-way setup. \INVTwo~is a dataset traditionally used to evaluate the robustness of models to natural distribution shifts. For our experiments we evaluate accuracy of the model given $n$ examples from the \INVTwo~distribution. We benchmark representations in the traditional small-scale (10-way) and large-scale (1000-way) setting. We evaluate for $n \in {1,3,5,7,9}$ with 9 being the maximum value for $n$ because there are 10 images per class.

We observed that \MH~had equal performance to \FF~across all representation sizes and shot numbers. We also found that for both \MH~and \FF, as the shot number decreased, the required representation size to reach optimal accuracy decreased (Table~\ref{Tab:FSL-INV2}). For example, we observed that 1-shot performance at $32$ representation size had equal accuracy to $2048$ representation size.
\begin{table}[ht]
\centering
\caption{Few-shot accuracy (\%) on ImageNetV2 for 1000-way classification. \MH performs equally to \FF~across all shots and representation sizes. We also observed that accuracy saturated at a lower dimension for lower shot numbers. E.g. for 1-shot, 32-dim performed comparably to 2048-dim.}
\vspace{1mm}
\begin{tabular}{@{}
c |
c |
c
c 
c 
c 
c @{}}
\toprule
\Dims& Method &
  1-Shot &
  3-Shot &
  5-Shot &
  7-Shot &
  9-Shot \\ \midrule
  \multirow{2}{*}{8}
& \FF & 35.41     & 45.73     & 49.23     & 50.89     & 51.72     \\
& \MH & 35.37     & 45.69     & 49.25     & 50.85     & 51.73     \\ \midrule
\multirow{2}{*}{16}
& \FF & 40.88     & 53.96     & 57.36     & 58.72     & 59.39     \\
& \MH & 40.90     & 53.94     & 57.37     & 58.65     & 59.29     \\ \midrule
\multirow{2}{*}{32}
& \FF & 41.41     & 54.88     & 58.28     & 59.63     & 60.40     \\
& \MH & 41.40     & 54.91     & 58.30     & 59.65     & 60.45     \\ \midrule
\multirow{2}{*}{64}
& \FF & 41.25     & 54.83     & 58.29     & 59.82     & 60.61     \\
& \MH & 41.28     & 54.80     & 58.32     & 59.77     & 60.69 \\ \midrule
\multirow{2}{*}{128}
& \FF & 41.36     & 54.90     & 58.50     & 60.05     & 60.90     \\
& \MH & 41.38     & 54.95     & 58.50     & 60.06     & 60.83     \\ \midrule
\multirow{2}{*}{256}
& \FF & 41.36 & 54.90 & 58.50 & 60.05 & 60.90 \\
& \MH & 41.38 & 54.95 & 58.50 & 60.06 & 60.83 \\ \midrule
\multirow{2}{*}{512}
& \FF & 41.36 & 55.05 & 58.70 & 60.19 & 61.02 \\
& \MH & 41.34 & 55.14 & 58.78 & 60.40 & 61.18 \\ \midrule
\multirow{2}{*}{1024} & \FF & 41.32 & 55.20 & 58.85 & 60.46 & 61.38 \\
& \MH & 41.31 & 55.24 & 58.86 & 60.42 & 61.34 \\ \midrule
\multirow{2}{*}{2048} & \FF & 41.18 & 55.09 & 58.77 & 60.38 & 61.34 \\
& \MH & 41.16 & 55.10 & 58.77 & 60.40 & 61.28 \\
\bottomrule
\end{tabular}
\label{Tab:FSL-INV2}
\end{table}

\paragraph{FLUID.}

For the long-tailed setting we evaluated \mrl on the FLUID benchmark~\citep{wallingford2020overfitting} which contains a mixture of pretrain and new classes. Table~\ref{tab:fluid} shows the evaluation of the learned representation on FLUID. We observed that \mrl provided up to 2\% higher accuracy on novel classes in the tail of the distribution, without sacrificing accuracy on other classes. Additionally we found the accuracy between low-dimensional and high-dimensional representations was marginal for pretrain classes. For example, the 64-dimensional \mrl performed $\sim1\%$ lower in accuracy compared to the 2048-dimensional counterpart on pretrain-head classes (84.46\% vs 85.60\%). However for novel-tail classes the gap was far larger (6.22\% vs 12.88\%). We hypothesize that the higher-dimensional representations are required to differentiate the classes when few training examples of each are known. These results provide further evidence that different tasks require varying capacity based on their difficulty.

\begin{table}[h]
\centering
\caption{Accuracy (\%) categories indicates whether classes were present during ImageNet pretraining and head/tail indicates classes that have greater/less than 50 examples in the streaming test set. We observed that \mrl performed better than the baseline on novel tail classes by $\sim2\%$ on average.}
\vspace{1mm}
\resizebox{\columnwidth}{!}{
\begin{tabular}{c|ccccccc}
\toprule
  Rep. Size & {Method} &
  {\begin{tabular}[c]{@{}c@{}}Pretrain \\ - Head (\textgreater 50)\end{tabular}} &
  {\begin{tabular}[c]{@{}c@{}}Novel \\ - Head (\textgreater 50)\end{tabular}} &
  {\begin{tabular}[c]{@{}c@{}}Pretrain \\ - Tail (\textless 50)\end{tabular}} &
  {\begin{tabular}[c]{@{}c@{}}Novel \\ - Tail (\textless 50)\end{tabular}} &
  {\begin{tabular}[c]{@{}c@{}}Mean Per Class\\  Acc.\end{tabular}} &
  {Acc.} \\ \midrule
\multirow{3}{*}{8}
& \FF                         & 68.04 & \textbf{11.30} & 33.18 & \textbf{0.36}  & 16.29 & 28.47          \\
& \MH                         & \textbf{71.75} & 10.70 & \textbf{38.29} & 0.19  & \textbf{17.15} & \textbf{29.34} \\
& \SH                         & 57.40 & 6.25  & 23.14 & 0.04  & 11.78 & 22.81          \\ \midrule

\multirow{3}{*}{16}
& \FF                         & 80.74 & \textbf{19.12} & \textbf{63.29} & \textbf{2.78}  & \textbf{25.65} & \textbf{37.61} \\
& \MH                         & \textbf{81.79} & 17.90 & 61.39 & 1.95  & 24.73 & 37.59          \\
& \SH                         & 79.08 & 9.15  & 60.33 & 0.08  & 20.45 & 30.24          \\ \midrule

\multirow{3}{*}{32}
& \FF                         & \textbf{83.67} & \textbf{24.30} & \textbf{66.66} & \textbf{4.23}  & \textbf{28.86} & \textbf{42.40} \\
& \MH                         & 83.46 & 23.26 & 65.82 & 3.75  & 28.16 & 41.90          \\
& \SH                         & 81.42 & 10.47 & 68.01 & 0.23  & 22.31 & 32.17          \\ \midrule

\multirow{3}{*}{64}
& \FF                         & 84.12 & 27.49 & 68.20 & 5.17  & 30.64 & 45.18          \\
& \MH                         & \textbf{84.46} & \textbf{27.61} & 67.59 & \textbf{6.22}  & \textbf{31.03} & \textbf{45.35} \\
& \SH                         & 82.57 & 13.23 & \textbf{70.18} & 0.52  & 23.83 & 34.74          \\ \midrule

\multirow{3}{*}{128}
& \FF                         & 84.87 & 29.96 & \textbf{68.79} & 5.54  & 31.84 & 47.06          \\
& \MH                         & \textbf{84.88} & \textbf{30.86} & 68.58 & \textbf{8.41}  & \textbf{33.23} & \textbf{47.79} \\
& \SH                         & 82.76 & 18.93 & 64.46 & 2.22  & 25.75 & 39.19          \\ \midrule

\multirow{3}{*}{256}
& \FF                         & 84.77 & 32.78 & \textbf{69.96} & 7.21  & 33.65 & 49.15          \\
& \MH                         & \textbf{85.10} & \textbf{32.91} & 69.39 & \textbf{9.99}  & \textbf{34.74} & \textbf{49.39} \\
& \SH                         & 82.96 & 22.63 & 64.55 & 3.59  & 27.64 & 41.96          \\ \midrule

\multirow{3}{*}{512}
& \FF                         & \textbf{85.62} & \textbf{35.27} & \textbf{70.27} & 9.05  & 35.42 & \textbf{51.14} \\
& \MH                         & \textbf{85.62} & 34.67 & 70.24 & \textbf{11.43} & \textbf{36.11} & {50.79} \\
& \SH                         & 82.86 & 25.62 & 64.34 & 4.99  & 29.22 & 44.20          \\ \midrule

\multirow{3}{*}{1024}
& \FF                         & \textbf{86.30} & 37.49 & \textbf{71.12} & 10.92 & \textbf{37.14} & \textbf{52.88} \\
& \MH                         & 85.64 & \textbf{35.88} & 70.02 & \textbf{12.19} & 36.80 & {51.58} \\
& \SH                         & 83.03 & 27.78 & 64.58 & 6.32  & 30.57 & 45.71          \\ \midrule

\multirow{3}{*}{2048}
& \FF & \textbf{86.40} & \textbf{37.09} & \textbf{71.74} & 10.77 & 37.04 & \textbf{52.67} \\
& \MH & 85.60 & 36.83 & 70.34 & \textbf{12.88} & \textbf{37.46} & 52.18          \\
& \SH & 83.01 & 29.99 & 65.37 & 7.60  & 31.97 & 47.16          \\ \bottomrule
\end{tabular}}
\label{tab:fluid}
\end{table}

\section{Robustness Experiments}
\label{sec:appendix_robustness}
\begin{table}[ht!]
\centering
 \caption{Top-1 classification accuracy (\%) on out-of-domain datasets (ImageNet-V2/R/A/Sketch) to examine robustness of \alg. Note that these results are without any fine tuning on these datasets.}
 \resizebox{1.0\linewidth}{!}{%
\begin{tabular}{@{}c|ccc|ccc|ccc|ccc|ccc@{}}
\toprule
                     & \multicolumn{3}{c|}{ImageNet-V1}                              & \multicolumn{3}{c|}{ImageNet-V2}                              & \multicolumn{3}{c|}{ImageNet-R}                               & \multicolumn{3}{c|}{ImageNet-A}                               & \multicolumn{3}{c}{ImageNet-Sketch}                          \\ \cmidrule{2-16}
\Dims & \FF & \SH & \MH & \FF & \SH & \MH & \FF & \SH & \MH & \FF & \SH & \MH & \FF & \SH & \MH \\ \midrule
8                    & 65.86            & 56.92            & 67.46            & 54.05              & 47.40            & 55.59            & 24.60            & 22.98            & 23.57            & 2.92             & 3.63             & 3.39             & 17.73            & 15.07            & 17.98            \\
16                   & 73.10            & 72.38            & 73.80            & 60.52              & 60.48            & 61.71            & 28.51            & 28.45            & 28.85            & 3.00             & 3.55             & 3.59             & 21.70            & 20.38            & 21.77            \\
32                   & 74.68            & 74.80            & 75.26            & 62.24              & 62.23            & 63.05            & 31.28            & 30.79            & 31.47            & 2.60             & 3.65             & 3.57             & 22.03            & 21.87            & 22.48            \\
64                   & 75.45            & 75.48            & 76.17            & 63.51              & 63.15            & 63.99            & 32.96            & 32.13            & 33.39            & 2.87             & 3.99             & 3.76             & 22.13            & 22.56            & 23.43            \\
128                  & 75.47            & 76.05            & 76.46            & 63.67              & 63.52            & 64.69            & 33.93            & 33.48            & 34.54            & 2.81             & 3.71             & 3.73             & 22.73            & 22.73            & 23.70            \\
256                  & 75.78            & 76.31            & 76.66            & 64.13              & 63.80            & 64.71            & 34.80            & 33.91            & 34.85            & 2.77             & 3.65             & 3.60             & 22.63            & 22.88            & 23.59            \\
512                  & 76.30            & 76.48            & 76.82            & 64.11              & 64.09            & 64.78            & 35.53            & 34.20            & 34.97            & 2.37             & 3.57             & 3.59             & 23.41            & 22.89            & 23.67            \\
1024                 & 76.74            & 76.60            & 76.93            & 64.43              & 64.20            & 64.95            & 36.06            & 34.22            & 34.99            & 2.53             & 3.56             & 3.68             & 23.44            & 22.98            & 23.72            \\
2048                 & 77.10            & 76.65            & 76.95            & 64.69              & 64.17            & 64.93            & 37.10            & 34.29            & 35.07            & 2.93             & 3.49             & 3.59             & 24.05            & 23.01            & 23.70            \\ \bottomrule
\end{tabular}
\label{tab:robustness}
}
\end{table}
\begin{table}[ht]
\centering
 \caption{Zero-shot top-1 image classification accuracy (\%) of a ALIGN-\mrl model on ImageNet-V1/V2/R/A and ObjectNet.}
 \resizebox{0.5\columnwidth}{!}{%
\begin{tabular}{@{}c|ccccc@{}}
\toprule
\Dims    & V1    & V2    & A     & R     & ObjectNet \\ \midrule
12       & 30.57 & 23.98 & 14.59 & 24.24 & 25.52     \\
24       & 45.64 & 37.71 & 22.75 & 46.40 & 35.89     \\
48       & 53.84 & 46.16 & 28.88 & 60.71 & 42.76     \\
96       & 58.31 & 51.34 & 33.21 & 70.12 & 45.20     \\
192      & 60.95 & 53.56 & 36.10 & 74.41 & 48.24     \\
384      & 62.06 & 54.77 & 37.95 & 76.51 & 49.10     \\
768      & 62.26 & 55.15 & 37.84 & 76.73 & 49.26     \\ \midrule
Baseline & 66.39 & 59.57 & 39.97 & 80.49 & 51.60      \\ \bottomrule
\end{tabular}
}
\label{tab:r50-align_zeroshot}
\end{table}
We evaluated the robustness of \nrl models on out-of-domain datasets (ImageNetV2/R/A/Sketch) and compared them to the \FF~baseline. Each of these datasets is described in Appendix~\ref{sec:datasets}. The results in Table~\ref{tab:robustness} demonstrate that learning \mrs does not hurt out-of-domain generalization relative to \FF~models, and \mrs in fact improve the performance on ImageNet-A. For a ALIGN--\mrl model, we examine the the robustness via zero-shot retrieval on out-of-domain datasets, including ObjectNet, in Table~\ref{tab:r50-align_zeroshot}.

\section{In Practice Costs}
\label{sec:appendix_real-world-perf}
All approximate NN search experiments via HNSW32 were run on an Intel Xeon 2.20GHz CPU with 24 cores. All exact search experiments were run with CUDA 11.0 on 2xA100-SXM4 NVIDIA GPUs with 40G RAM each. 

\paragraph{\mrl models.}
As \MH makes minimal modifications to the ResNet50 model in the final fc layer via multiple heads for representations at various scales, it has only an 8MB storage overhead when compared to a standard ResNet50 model. \SH~has no storage overhead as it has a shared head for logits at the final fc layer.

\paragraph{Retrieval}
Exact search has a search time complexity of $O(dkN)$, and HNSW has a search time complexity of $O(dk\log(N))$, where $N$ is the database size, $d$ is the representation size, and $k$ is the shortlist length. To examine real-world performance, we tabulated wall clock search time for every query in the \InIk and \InIVk validation sets over all representation sizes $d$ in Table~\ref{tab:ret_searchtime} for both Exact Search and HNSW32, and ablated wall clock query time over shortlist length $k$ on the \InIk validation set in Table~\ref{tab:retrieval_searchtime_shortlist}. The wall clock time to build the index and the index size is also shown in Table~\ref{tab:ret_buildtime_indexsize}.

\begin{table}[!ht]
\centering
 \caption{Retrieval k-NN wall clock search times (s) over the entire validation (query) set of \InIk\space and \InIVk, containing 50K and 200K samples respectively.}
 \vspace{1mm}
 \resizebox{0.5\columnwidth}{!}{%
\begin{tabular}{@{}c|cc|cc@{}}
\toprule
\multirow{2}{*}{\Dims} & \multicolumn{2}{c|}{\InIk} & \multicolumn{2}{c}{\InIVk} \\ \cmidrule(l){2-5} 
                       & ExactL2    & HNSW32    & ExactL2    & HNSW32    \\ \midrule
8                      & 0.60       & 0.14      & ~35.70     & ~1.17     \\
16                     & 0.57       & 0.18      & ~36.16     & ~1.65     \\
32                     & 0.60       & 0.20      & ~36.77     & ~1.75     \\
64                     & 0.66       & 0.24      & ~27.88     & ~2.21     \\
128                    & 0.86       & 0.32      & ~30.10     & ~4.15     \\
256                    & 1.29       & 0.46      & ~34.97     & ~3.39     \\
512                    & 2.17       & 0.68      & ~46.97     & ~4.83     \\
1024                   & 3.89       & 1.05      & ~70.59     & ~7.14     \\
2048                   & 7.31       & 2.05      & 117.78     & 13.43     \\ \bottomrule
\end{tabular}
\label{tab:ret_searchtime}
}
\end{table}
\begin{table}[!ht]
\centering
 \caption{FAISS~\citep{johnson2019billion} index size and build times for exact k-NN search with L2 Distance metric and approximate k-NN search with HNSW32~\citep{malkov2018efficient}. }
 \vspace{1mm}
 \resizebox{\columnwidth}{!}{%
\begin{tabular}{@{}c|cc|cc|cc|cc@{}}
\toprule
\multirow{4}{*}{\Dims} & \multicolumn{4}{c|}{Exact Search}                                                                                                                                                                                                                          & \multicolumn{4}{c}{HNSW32}                                                                                                                                                                                                                                \\ \cmidrule(l){2-9} 
                       & \multicolumn{2}{c|}{\InIk}                                                                                              & \multicolumn{2}{c|}{\InIVk}                                                                                                      & \multicolumn{2}{c|}{\InIk}                                                                                                      & \multicolumn{2}{c}{\InIVk}                                                                                                      \\ \cmidrule{2-9}
                       & \begin{tabular}[c]{@{}c@{}}Index Size\\ (MB)\end{tabular} & \begin{tabular}[c]{@{}c@{}}Index Build \\ Time (s)\end{tabular} & \begin{tabular}[c]{@{}c@{}}Index Size\\ (MB)\end{tabular} & \begin{tabular}[c]{@{}c@{}}Index Build\\  Time (s)\end{tabular} & \begin{tabular}[c]{@{}c@{}}Index Size\\ (MB)\end{tabular} & \begin{tabular}[c]{@{}c@{}}Index Build \\ Time (s)\end{tabular} & \begin{tabular}[c]{@{}c@{}}Index Size\\ (MB)\end{tabular} & \begin{tabular}[c]{@{}c@{}}Index Build\\  Time (s)\end{tabular} \\ \midrule
8                      & ~~~40                                                     & ~0.04                                                           & ~~131                                                     & ~0.33                                                           & ~~381                                                     & ~4.87                                                           & ~1248                                                     & ~24.04                                                          \\
16                     & ~~~80                                                     & ~0.08                                                           & ~~263                                                     & ~0.27                                                           & ~~421                                                     & ~6.15                                                           & ~1379                                                     & ~33.31                                                          \\
32                     & ~~160                                                     & ~0.16                                                           & ~~525                                                     & ~0.52                                                           & ~~501                                                     & ~6.80                                                           & ~1642                                                     & ~37.41                                                          \\
64                     & ~~320                                                     & ~0.38                                                           & ~1051                                                     & ~1.05                                                           & ~~661                                                     & ~8.31                                                           & ~2167                                                     & ~47.23                                                          \\
128                    & ~~641                                                     & ~0.64                                                           & ~2101                                                     & ~2.10                                                           & ~~981                                                     & 11.73                                                           & ~3218                                                     & ~89.87                                                          \\
256                    & ~1281                                                     & ~1.27                                                           & ~4202                                                     & ~4.20                                                           & ~1622                                                     & 17.70                                                           & ~5319                                                     & 102.84                                                          \\
512                    & ~2562                                                     & ~2.52                                                           & ~8404                                                     & ~8.39                                                           & ~2903                                                     & 27.95                                                           & ~9521                                                     & 158.47                                                          \\
1024                   & ~5125                                                     & ~5.10                                                           & 16808                                                     & 17.20                                                           & ~5465                                                     & 44.02                                                           & 17925                                                     & 236.30                                                          \\
2048                   & 10249                                                     & 10.36                                                           & 33616                                                     & 41.05                                                           & 10590                                                     & 86.15                                                           & 34733                                                     & 468.18                                                          \\ \bottomrule
\end{tabular}
\label{tab:ret_buildtime_indexsize}
}
\end{table}
\begin{table}[!ht]
\centering
 \caption{Retrieval k-NN wall clock search times (s) over entire validation (query) set of \InIk~over various shortlist lengths $k$.}
 \vspace{1mm}
 \resizebox{0.7\columnwidth}{!}{%
\begin{tabular}{@{}ccccccc@{}}
\toprule
Index    & k = 50 & k = 100 & k = 200 & k = 500 & k = 1000 & k = 2048 \\ \midrule
Exact L2 & 0.4406 & 0.4605  & 0.5736  & 0.6060  & 1.2781   & 2.7047   \\
HNSW32   & 0.1193 & 0.1455  & 0.1833  & 0.2145  & 0.2333   & 0.2670   \\ \bottomrule
\end{tabular}
\label{tab:retrieval_searchtime_shortlist}
}
\end{table}

\section{Analysis of Model Disagreement}
\label{app:Model_Disagree}

\begin{figure}[ht!]
\centering
\resizebox{\columnwidth}{!}{%
    \includegraphics[width=1\columnwidth]{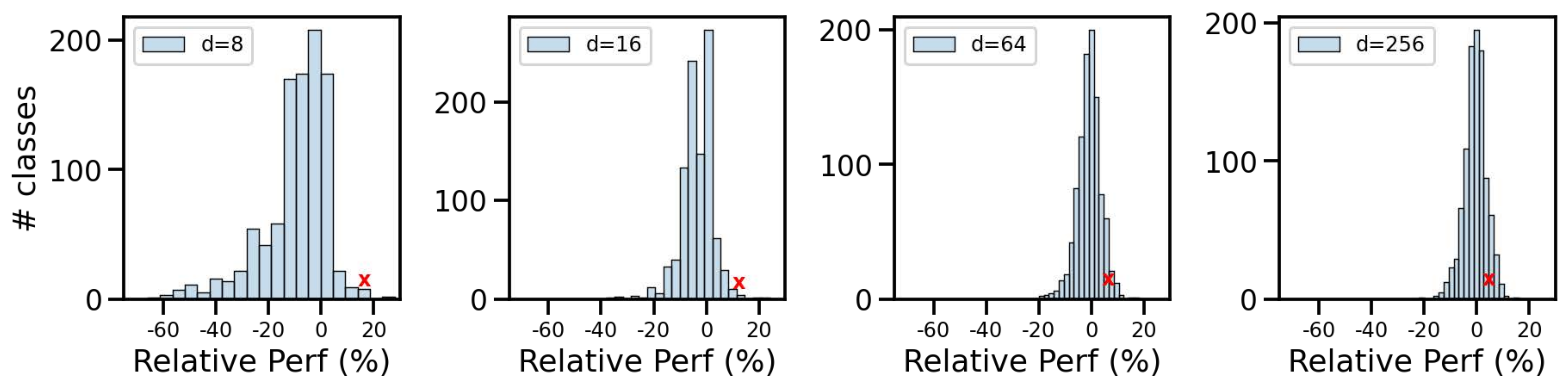}
}
\caption{Progression of relative per-class accuracy vs \mrl-2048. As the dimensionality increases, the spread shrinks while the class marked (\textbf{{\textcolor{red}x}}) (Madagascar cat) loses accuracy.}
\label{fig:r50-perclass}
\end{figure}

\paragraph{Class Trends} \textit{Does increasing representation size necessarily help improve classification performance across all classes in \InIk?} We studied this question by examining trends in performance with increasing representation size from $d = {8, ... 2048}$. For \nrl models, we observed that $244$ classes showed a monotonic improvement in performance with increasing $d$, $177$ classes first improved but then observed a slight dip (one or two misclassifications per class), $49$ classes showed a decline first and then an improvement, and the remaining classes did not show a clear trend. When we repeated this experiment with independently trained \FF~models, we noticed that $950$ classes did not show a clear trend. This motivated us to leverage the disagreement as well as gradual improvement of accuracy at different representation sizes by training \mrs. Figure~\ref{fig:r50-perclass} showcases the progression of relative per-class accuracy distribution compared to the \MRL-2048 dimensional model. This also showed that some instances and classes could benefit from lower-dimensional representations.

\paragraph{Discussion of Oracle Accuracy}
Based on our observed model disagreements for different representation sizes $d$, we defined an optimal \textit{oracle} accuracy~\citep{lee2016stochastic} for \nrl. We labeled an image as correctly predicted if classification using any representation size was correct. The percentage of total samples of \InIk that were firstly correctly predicted using each representation size $d$ is shown in Table~\ref{tab:oracle-acc}. This defined an upper bound on the performance of \nrl models, as $18.46\%$ of the \InIk validation set were incorrectly predicted $\forall d\in\{8, 16, \ldots, 2048\}$. We show the oracle performance on \nrl models for ImageNet-1K/V2/A/R/Sketch datasets in Table~\ref{tab:oracle_robustness}. 

In an attempt to derive an optimal routing policy to emulate oracle accuracy, we designed the adaptive classification via cascading method as discussed in Appendix~\ref{sec:appendix_adaptive_classification}. This led to an interesting observation on the expected dimensionality for $76.30\%$ top-1 classification accuracy being just $d \sim37$. We leave the design and learning of a more optimal policy for future work.
\begin{table}[h]
\centering
 \caption{Percentage of \InIk validation set that is first correctly predicted using each representation size $d$. We note that $18.46\%$ of the samples cannot be correctly predicted by any representation size. The remaining $81.54\%$ constitutes the oracle accuracy.}
 \vspace{1mm}
 \resizebox{.9\columnwidth}{!}{
\begin{tabular}{@{}c|ccccccccc|c@{}}
\toprule
\Dims                                                          & 8     & 16   & 32   & 64   & 128  & 256  & 512  & 1024 & 2048 & \begin{tabular}[c]{@{}c@{}}Always\\ Wrong\end{tabular} \\ \midrule
\begin{tabular}[c]{@{}c@{}}Correctly\\ Predicted\end{tabular} & 67.46 & 8.78 & 2.58 & 1.35 & 0.64 & 0.31 & 0.20 & 0.12 & 0.06 & 18.46                                                  \\ \bottomrule
\end{tabular}
\label{tab:oracle-acc}
}
\end{table}
\begin{table}[h]
\centering
 \caption{Oracle classification accuracy of various evaluation datasets for ResNet50--\mrl model trained on ImageNet-1K.}
 \vspace{1mm}
 \resizebox{0.8\columnwidth}{!}{
\begin{tabular}{@{}c|ccccc@{}}
\toprule
Top-1   & ImageNetV1 & \INVTwo & ImageNet-A & ImageNet-R & ImageNet-Sketch \\ \midrule
\FF--2048 & 76.9  & 64.9    & 3.6        & 35.1       & 23.7            \\
\mrl--Oracle  & 81.5  & 70.6    & 8.7        & 39.8       & 28.9            \\ \bottomrule
\end{tabular}
\label{tab:oracle_robustness}
}
\end{table}

\paragraph{Grad-CAM Examples}
We analyzed the nature of model disagreement across representation sizes with \nrl models with the help of Grad-CAM visualization~\citep{selvaraju2017grad}. We observed there were certain classes in \InIk~such as "tools", "vegetables" and "meat cutting knife" which were occasionally located around multiple objects and a cluttered environment. In such scenarios, we observed that smaller representation size models would often get confused due to other objects and fail to extract the object of interest which generated the correct label. We also observed a different nature of disagreement arising when the models got confused within the same superclass. For example, \InIk has multiple "snake" classes, and models often confuse a snake image for an incorrect species of snake. 

\paragraph{Superclass Performance}
We created a 30 superclass subset of the validation set based on wordnet hierarchy (Table~\ref{tab:superclass_names}) to quantify the performance of \mrl model on \InIk superclasses. Table~\ref{tab:superclass} quantifies the performance with different representation size. 

\begin{table}[ht]
\caption{30 Superclasses in ImageNet-1K corresponding to the performance in Table~\ref{tab:superclass}.}
\begin{tabular}{
c 
c 
c 
c 
c}
\toprule
insect              & motor vehicle & artiodactyl        & vegetable            & game equipment        \\
terrier             & serpent       & machine            & measuring device     & sheepdog              \\
protective covering & sporting dog  & vessel, watercraft & building             & lizard                \\
garment             & hound         & monkey             & home appliance       & wind instrument \\
vessel              & fish          & nourishment        & electronic equipment & oscine                \\
furniture           & wading bird   & tool               & canine               & mechanism            

\end{tabular}
\label{tab:superclass_names}
\end{table}
\begin{table}[!ht]
\centering
 \caption{Performance of \mrl model on 31-way classification (1 extra class is for reject token) on \InIk superclasses.}
 \resizebox{1\columnwidth}{!}{%
\begin{tabular}{@{}c|ccccccccc}
\toprule
\Dims  & 8 & 16 & 32 & 64  &128 & 256 & 512 & 1024 & 2048 \\\midrule
\mrl & 85.57 & 88.67 & 89.48 & 89.82 & 89.97 & 90.11 & 90.18 & 90.22 & 90.21   \\\bottomrule
\end{tabular}
}

\label{tab:superclass}
\end{table}

\section{Ablation Studies}
\label{sec:ablation}

\subsection{\MH~Training Paradigm}
\label{sec:ablation-mrl-training}
\begin{table}[h]
\centering
 \caption{Top-1 classification accuracy (\%) on \InIk~of various ResNet50 models which are finetuned on pretrained \FF-2048 model. We observed that adding more non-linearities is able to induce nesting to a reasonable extent even if the model was not pretrained with nesting in mind.}
 \vspace{1mm}
 \resizebox{0.7\columnwidth}{!}{
\begin{tabular}{@{}c|cccc|c@{}}
\toprule
\Dims & fc         & \begin{tabular}[c]{@{}c@{}}4.2 conv3, \\ fc\end{tabular} & \begin{tabular}[c]{@{}c@{}}4.2 conv2, \\ conv3, fc\end{tabular} & \begin{tabular}[c]{@{}c@{}}4.2 full, \\ fc\end{tabular} & All (\MH) \\ \midrule
8                    & ~5.15 & 36.11                                                    & 54.78                                                           & 60.02                                                   & 66.63                    \\
16                   & 13.79      & 58.42                                                    & 67.26                                                           & 70.10                                                   & 73.53                    \\
32                   & 32.52      & 67.81                                                    & 71.62                                                           & 72.84                                                   & 75.03                    \\
64                   & 52.66      & 72.42                                                    & 73.61                                                           & 74.29                                                   & 75.82                    \\
128                  & 64.60      & 74.41                                                    & 74.67                                                           & 75.03                                                   & 76.30                    \\
256                  & 69.29      & 75.30                                                    & 75.23                                                           & 75.38                                                   & 76.47                    \\
512                  & 70.51      & 75.96                                                    & 75.47                                                           & 75.64                                                   & 76.65                    \\
1024                 & 70.19      & 76.18                                                    & 75.70                                                           & 75.75                                                   & 76.76                    \\
2048                 & 69.72      & 76.44                                                    & 75.96                                                           & 75.97                                                   & 76.80                    \\ \bottomrule
\end{tabular}
\label{tab:nesting_as_finetuning}
}
\end{table}
\paragraph{\mrs via Finetuning.} To observe if nesting can be induced in models that were not explicitly trained with nesting from scratch, we loaded a pretrained \FF-2048 ResNet50 model and initialized a new \MH~ layer, as defined in Algorithm~\ref{code:MRL}, Appendix~\ref{sec:appendix-mrl_model_training}. We then unfroze different layers of the backbone to observe
how much non-linearity in the form of unfrozen conv layers needed to be present to enforce nesting into a pretrained \FF~model. A description of these layers can be found in the ResNet50 architecture~\cite{he2016deep}. All models were finetuned with the FFCV pipeline, with same training configuration as in the end-to-end training aside from changing lr $ = 0.1$ and epochs $ = 10$. We observed that finetuning the linear layer alone was insufficient to learn \mrs at lower dimensionalities. Adding more and more non-linear conv+ReLU layers steadily improved classification accuracy of $d = 8$ from $5\%$ to $60\%$ after finetuning, which was only $6\%$ less than training \MH~end-to-end for 40 epochs. This difference was successively less pronounced as we increased dimensionality past $d = 64$, to within $1.5\%$ for all larger dimensionalities. The full results of this ablation can be seen in Table~\ref{tab:nesting_as_finetuning}.

\begin{table}[t]
\centering
 \caption{An ablation over boosting training loss at lower nesting dimensions, with top-1 and top-5 accuracy (\%). The models are described in Appendix~\ref{sec:ablation-mrl-training}.}
 \vspace{1mm}
 \resizebox{0.7\columnwidth}{!}{
\begin{tabular}{@{}c|cc|cc|cc@{}}
\toprule
Model       & \multicolumn{2}{c}{\MH} & \multicolumn{2}{c}{\MH-8boost} & \multicolumn{2}{c}{\MH-8+16boost} \\ \midrule
\Dims & Top-1      & Top-5      & Top-1          & Top-5         & Top-1           & Top-5           \\ \midrule
8           & 66.63      & 84.66      & \textbf{69.53}          & 86.19         & 69.24           & 85.96           \\
16          & 73.53      & 89.52      & 73.86          & 89.44         & \textbf{73.91}           & 89.55           \\
32          & 75.03      & 91.31      & \textbf{75.28}          & 91.21         & 75.10           & 91.14           \\
64          & 75.82      & 92.27      & \textbf{75.84}          & 92.22         & 75.67           & 92.06           \\
128         & \textbf{76.30}      & 92.82      & 76.28          & 92.74         & 76.07           & 92.52           \\
256         & 76.47      & 93.02      & \textbf{76.48}          & 92.97         & 76.22           & 92.72           \\
512         & \textbf{76.65}      & 93.13      & 76.56          & 93.09         & 76.35           & 92.85           \\
1024        & \textbf{76.76}     & 93.22      & 76.71          & 93.21         & 76.39           & 92.98           \\
2048        & \textbf{76.80}      & 93.32      & 76.76          & 93.28         & 76.52           & 93.05           \\ \bottomrule
\end{tabular}
\label{tab:ablation_trainloss}
}
\end{table}

\paragraph{Relative Importance.} We performed an ablation of \mrl over the relative importance, $c_m$, of different nesting dimensions $m\in\cal{M}$, as defined in Sec.~\ref{sec:method}. In an attempt to improve performance at lower dimensionalities, we boosted the relative importance $c_m$ of training loss at lower dimensions as in Eq.~\ref{eq:phase1} with two models, \MH-8boost and \MH-8+16boost. The \MH-8boost model had $c_{m\in\cal M} = [2, 1, 1, 1, 1, 1, 1, 1, 1]$ and the \MH-8+16boost model had $c_{m\in\cal M} = [2, 1.5, 1, 1, 1, 1, 1, 1, 1]$. The relative importance list $c_{m\in\cal M}$ had a 1-to-1 correspondence with nesting dimension set $\mathcal{M}$. In Table~\ref{tab:ablation_trainloss}, we observed that \MH-8boost improves top-1 accuracy by $3\%$ at $d = 8$, and also improves top-1 accuracy of all representation scales from 16 to 256 over \MH, while only hurting the performance at 512 to 2048 representation scales by a maximum of 0.1\%. This suggests that the relative importance $c_m$ can be tuned/set for optimal accuracy for all $m\in\mathcal{M}$, but we leave this extension for future work.

\paragraph{\mrs at Arbitrary Granularities.} To train \mrl, we used nested dimensions at logarithmic granularities $\mathcal{M} = \{8,16,\ldots,1024,2048\}$ as detailed in Section~\ref{sec:method}. We made this choice for two empirically-driven reasons: a) The accuracy improvement with increasing representation size was more logarithmic than linear (as shown by \FF~models in Figure~\ref{fig:r50-acc}). This indicated that optimizing for granularities increasing in a non-logarithmic fashion would be sub-optimal both for maximum performance and expected efficiency; b) If we have $m$ arbitrary granularities, the expected cost of the linear classifier to train \mrl scales as $O(L*(m^2))$ while logarithmic granularities result in $O(L*2log(d))$ space and compute costs.

To demonstrate this effect, we learned \mrs with uniform (\textit{\mrl-Uniform}) nesting dimensions $m\in\mathcal{M} = \{8, 212, 416, 620, 824, 1028, 1232, 1436, 1640, 1844, 2048\}$. We evaluated this model at the standard (\textit{\mrl-log}) dimensions $m\in\mathcal{M} = \{8, 16, 32, 64, 128, 256, 512, 1024, 2048\}$ for ease of comparison to reported numbers using 1-NN accuracy (\%). As shown in Table~\ref{tab:ablation_uniform_nesting}, we observed that while performance interpolated, \textit{\mrl-Uniform} suffered at low dimensions as the logarithmic spacing of \textit{\mrl-log} resulted in tighter packing of information in these initial dimensions. The higher nesting dimensions of \textit{\mrl-Uniform} did not help in significant accuracy improvement due to accuracy saturation, which is often logarithmic in representation size as shown by \FF~models. Note that the slight improvement at dimensions higher than 512 for \textit{\mrl-Uniform} is due to multiple granularities around them compared to just three for \textit{\mrl-log}, which are not useful in practice for efficiency.

\paragraph{Lower Dimensionality.} We experimented with training \mrl with smaller nesting dimension than $m = 8$, as shown in Table~\ref{tab:ablation_train_smalldim}, with two models: MRL-4 and MRL-6. We found that using lower than 8-dimensions to train \mrl, i.e. $m_0 \in \{4, 6\}$ for MRL-4 and MRL-6 respectively, did not affect the top-1 accuracy of other granularities significantly. However, granularities smaller than 8-dimensions had very low accuracy and were often unusable for deployment along with additional training difficulty. We also observed a small dip in accuracy at higher dimensions which we attribute to the joint loss that now also included the harder optimization of the smallest dimension. Lastly, we hypothesize the dimensionality of 8 is an empirically validated design choice due to the considerable accuracy it provided along with the ease of training.
\begin{table}[t!]
\begin{minipage}[b]{.49\columnwidth}
\centering
\caption{An ablation over training with smaller nesting dimensionalities in terms of Top-1 accuracy (\%). MRL-4 and MRL-6 are variations of the original model (MRL-8) with $m_0 \in \{4,6\}$, where $m\in\mathcal{M}$ is part of the nesting\_list as seen in Alg~\ref{code:MRL}.}
\begin{tabular}{@{}c|ccc@{}}
\toprule
\Dims & MRL-4 & MRL-6 & MRL-8 \\ \midrule
4                    & 27.25 & -              & -              \\
6                    & -     & 58.71          & -              \\
8                    & 66.86 & \textbf{67.55} & 66.63          \\
16                   & 73.36 & 73.10          & \textbf{73.53} \\
32                   & 74.82 & 74.49          & \textbf{75.03} \\
64                   & 75.51 & 75.32          & \textbf{75.82} \\
128                  & 75.93 & 75.61          & \textbf{76.30} \\
256                  & 76.08 & 75.82          & \textbf{76.47} \\
512                  & 76.31 & 75.93          & \textbf{76.65} \\
1024                 & 76.38 & 76.04          & \textbf{76.76} \\
2048                 & 76.43 & 76.12          & \textbf{76.80} \\ \bottomrule
\end{tabular}
\label{tab:ablation_train_smalldim}
\end{minipage}
\hfill
\begin{minipage}[b]{.48\columnwidth}
\centering
\caption{An ablation over training \mrl with nesting list at uniformly distributed granularities. Entries in the \mrl-Uniform column are evaluated at logarithmic dimensions for a fair comparison to \mrl-Log (standard \mrl) with 1-NN accuracy (\%).}
\begin{tabular}{@{}c|cc@{}}
\toprule
\Dims & \mrl-Log & \mrl-Uniform \\ \midrule
8                    & \textbf{62.19}      & 58.44                       \\
16                   & \textbf{67.91}      & 61.11                       \\
32                   & \textbf{69.46}      & 63.82                       \\
64                   & \textbf{70.17}      & 66.44                       \\
128                  & \textbf{70.52}      & 68.71                       \\
256                  & \textbf{70.62}      & 70.06                       \\
512                  & 70.82               & \textbf{70.98}              \\
1024                 & 70.89               & \textbf{71.37}              \\
2048                 & 70.97               & \textbf{71.44}              \\ \bottomrule
\end{tabular}
\label{tab:ablation_uniform_nesting}
\end{minipage}
\end{table}

\subsection{Retrieval}
\label{sec:appendix_retrieval_ablation}
\paragraph{Adaptive Retrieval.}To examine the effect of increasing shortlist lengths on search time, we performed a reranking ablation over shortlist lengths for \retdim = 16 and \rerankdim = 2048 over \InIk~ in Table~\ref{tab:rerank_ablation_shortlist_in1k}, and over \InIVk~in Table~\ref{tab:rerank_ablation_shortlist_in4k}. We observed that using a larger shortlist $k$ saturated \InIk performance at $k$=200. But using larger shortlists until $k=2048$, the maximum value supported by the FAISS framework, steadily improved performance on \InIVk. This is likely due to the increased database size, but could also indicate a correlation with \InIVk being slightly out-of-distribution making the task at hand harder. 
\begin{table}[t!]
\centering
 \caption{Adaptive retrieval ablation over shortlist length $k$ for $D_s = 16$, $D_r = 2048$ on \InIk~ with exact search. Entries with the highest P@1 and mAP@10 across all $k$ are in bold.}
 \vspace{1mm}
 \resizebox{\columnwidth}{!}{%
\begin{tabular}{@{}c|c|cccc|cccc@{}}
\toprule
\begin{tabular}[c]{@{}c@{}}Shortlist\\ Length\end{tabular} & P@1          & mAP@10         & mAP@25 & mAP@50 & mAP@100 & P@10  & P@25  & P@50  & P@100 \\ \midrule
~100                                                       & 70.88          & 65.19          & 63.62  & 62.59  & 61.24   & 69.96 & 69.24 & 68.53 & 67.20 \\
~200                                                       & 70.90          & \textbf{65.27} & 63.73  & 62.82  & 61.97   & 70.10 & 69.44 & 68.90 & 68.21 \\
~400                                                       & 70.94          & 65.26          & 63.71  & 62.81  & 62.03   & 70.15 & 69.51 & 69.02 & 68.47 \\
~800                                                       & 70.96          & 65.23          & 63.64  & 62.69  & 61.85   & 70.16 & 69.52 & 69.02 & 68.45 \\
1600                                                       & 70.96          & 65.20          & 63.58  & 62.58  & 61.66   & 70.16 & 69.5  & 68.97 & 68.36 \\
2048                                                       & \textbf{70.97} & 65.20          & 63.57  & 62.58  & 61.64   & 70.16 & 69.5  & 68.97 & 68.35 \\ \bottomrule
\end{tabular}
\label{tab:rerank_ablation_shortlist_in1k}
}
\end{table}
\begin{table}[t!]
\centering
 \caption{Adaptive retrieval ablation over shortlist length $k$ for $D_s = 16$, $D_r = 2048$ on \InIVk~with exact search.}
 \vspace{1mm}
 \resizebox{\columnwidth}{!}{%
\begin{tabular}{@{}c|c|cccc|cccc@{}}
\toprule
\begin{tabular}[c]{@{}c@{}}Shortlist\\ Length\end{tabular} & P@1   & mAP@10 & mAP@25 & mAP@50 & mAP@100 & P@10  & P@25  & P@50  & P@100 \\ \midrule
~100                                                       & 27.70 & 14.38  & 10.62  & ~8.26  & 6.07    & 20.12 & 16.87 & 14.29 & 11.26 \\
~200                                                       & 28.56 & 15.21  & 11.43  & ~9.11  & 7.12    & 21.23 & 18.13 & 15.73 & 13.27 \\
~400                                                       & 29.34 & 15.83  & 12.06  & ~9.76  & 7.79    & 22.08 & 19.09 & 16.83 & 14.54 \\
~800                                                       & 29.86 & 16.30  & 12.53  & 10.23  & 8.26    & 22.72 & 19.83 & 17.65 & 15.45 \\
1600                                                       & 30.24 & 16.63  & 12.86  & 10.56  & 8.60    & 23.18 & 20.36 & 18.23 & 16.11 \\
2048                                                       & \textbf{30.35} & \textbf{16.73}  & 12.96  & 10.65  & 8.69    & 23.31 & 20.50 & 18.40 & 16.30 \\ \bottomrule
\end{tabular}
\label{tab:rerank_ablation_shortlist_in4k}
}
\end{table}

\end{document}